\documentclass[lettersize,journal]{IEEEtran}

\ifCLASSINFOpdf
\else
\fi
\usepackage{graphicx}
\usepackage{animate}
\usepackage{amsfonts}
\usepackage{bbm}
\usepackage{adjustbox}
\usepackage{bm}
\usepackage[dvipsnames]{xcolor}
\usepackage{tabularx}
\usepackage{xspace}
\usepackage{color}
\usepackage{multirow}
\usepackage{graphics}
\usepackage{adjustbox}
\usepackage{subfigure}
\usepackage{amsmath}
\usepackage{paralist} % compactitem, compactenum
\usepackage{enumitem} % adjust enum margin
\usepackage{booktabs} % Publication quality tables
\usepackage{array}
\usepackage{grffile}
\usepackage{makecell}
\usepackage{diagbox}

\graphicspath{{figures/}}
\usepackage[pagebackref=false,breaklinks=true,letterpaper=true,colorlinks,bookmarks=false,citecolor=black,linkcolor=black]{hyperref}

\long\def\ignorethis#1{}

\usepackage{tabularx}

\definecolor{B1}{RGB}{237,219,201}
\definecolor{MyBlueb}{RGB}{95,169,236}
\definecolor{MyOrange}{RGB}{255,177,98}
\definecolor{MyReda}{RGB}{193,39,45}
\definecolor{gray}{rgb}{0.5,0.5,0.5}
\definecolor{MyBlue}{rgb}{0,0,1.0}
\definecolor{MyYellow}{rgb}{0.9,0.9,0}
\definecolor{MyRed}{rgb}{0.8,0.2,0}
\definecolor{MyGreen}{rgb}{0,0.5,0.0}
\definecolor{MyGray}{rgb}{0.4,0.4,0.4}

\newlength\paramargin
\newlength\figmargin
\newlength\secmargin

\newcolumntype{L}[1]{>{\raggedright\let\newline\\\arraybackslash\hspace{0pt}}m{#1}}
\newcolumntype{C}[1]{>{\centering\let\newline\\\arraybackslash\hspace{0pt}}m{#1}}
\newcolumntype{R}[1]{>{\raggedleft\let\newline\\\arraybackslash\hspace{0pt}}m{#1}}

\def\etal{et~al.\xspace}

\makeatletter

\newcommand{\scriptveryshortarrow}[1][5pt]{{%
		\hbox{\rule[\scriptratio\dimexpr\fontdimen22\textfont2-.2pt\relax]
			{\scriptratio\dimexpr#1\relax}{\scriptratio\dimexpr.4pt\relax}}%
		\mkern-5mu\hbox{\let\f@size\sf@size\usefont{U}{lasy}{m}{n}\symbol{41}}}}
\usepackage{ulem}

\makeatother

\usepackage{bm}
\usepackage{mleftright}
\usepackage{nicefrac}
\usepackage{textcomp}
\usepackage[shortcuts]{extdash}

\begin{document}
\title{H$^2$-Stereo: High-Speed, High-Resolution Stereoscopic Video System}

\author{Ming~Cheng,
        Yiling~Xu,
        Wang~Shen,~%~\IEEEmembership{Student Member,~IEEE,}
        M.~Salman~Asif,~\IEEEmembership{Senior Member,~IEEE},
        Chao~Ma,~%~\IEEEmembership{Member,~IEEE,}
        Jun Sun, \\
        and~Zhan~Ma,~\IEEEmembership{Senior Member,~IEEE}

\thanks{M. Cheng, Y. Xu, J. Sun are with the Cooperative Medianet Innovation Center, Shanghai Jiao Tong University, Shanghai 200240, China (e-mail: ming\_cheng, yl.xu, junsun@sjtu.edu.cn). W. Shen is with the Institute of Image Communication and Network Engineering, Shanghai Jiao Tong University, Shanghai 200240, China (e-mail: shenwang@sjtu.edu.cn). M.S. Asif is with the University of California at Riverside, Riverside, CA 92521. (e-mail: sasif@ece.ucr.edu). C. Ma is with the MoE Key Lab of Artificial Intelligence, AI Institute, Shanghai Jiao Tong University, Shanghai 200240, China. (e-mail: chaoma@sjtu.edu.cn). Z. Ma is with the School of Electronic Science and Engineering, Nanjing University, Jiangsu 210093, China (e-mail: mazhan@nju.edu.cn).
}
\thanks{
This paper is supported in part by National Key R\&D Program of China (2018YFE0206700), National Natural Science Foundation of China (61971282, U20A20185). The corresponding author is Yiling Xu (e-mail: yl.xu@sjtu.edu.cn).
}
\thanks{
This paper has supplementary material available at~\url{https://www.dropbox.com/sh/pb43loenjcvlegy/AADoAcFfDnQMrX8ioh-R-wRXa?dl=0}. 
}}

\IEEEtitleabstractindextext{%
\begin{abstract}
    High-speed, high-resolution stereoscopic (H$^2$-Stereo) video allows us to perceive dynamic 3D content at fine granularity. 
    The acquisition of H$^2$-Stereo video, however, remains challenging with commodity cameras. 
    Existing spatial super-resolution or temporal frame interpolation methods provide compromised solutions that lack temporal or spatial details, respectively.
    To alleviate this problem, we propose a dual camera system, in which one camera captures high-spatial-resolution low-frame-rate (HSR-LFR) videos with rich spatial details, and the other captures low-spatial-resolution high-frame-rate (LSR-HFR) videos with smooth temporal details.
    We then devise a \textbf{L}earned \textbf{I}nformation \textbf{F}usion network (LIFnet) that exploits the cross-camera redundancies to enhance both camera views to high spatiotemporal resolution (HSTR) for reconstructing the H$^2$-Stereo video effectively.
    We utilize a disparity network to transfer spatiotemporal information across views even in large disparity scenes, based on which, we propose disparity-guided flow-based warping for LSR-HFR view and complementary warping for HSR-LFR view.
    A multi-scale fusion method in feature domain is proposed to minimize occlusion-induced warping ghosts and holes in HSR-LFR view.
    The LIFnet is trained in an end-to-end manner using our collected high-quality Stereo Video dataset from YouTube.
    Extensive experiments demonstrate that our model outperforms existing state-of-the-art methods for both views on synthetic data and camera-captured real data with large disparity.
    Ablation studies explore various aspects, including spatiotemporal resolution, camera baseline, camera desynchronization, long/short exposures and applications, of our system to fully understand its capability for potential applications.
\end{abstract}

% % Note that keywords are not normally used for peerreview papers.
\begin{IEEEkeywords}
Dual camera, Stereoscopic video, High-speed high-resolution video, optical flow, disparity
\end{IEEEkeywords}
}

\markboth{IEEE Transactions on Broadcasting, 2022}%
{Cheng \MakeLowercase{\textit{et al.}}: H$^2$-Stereo: High-Speed, High-Resolution Stereoscopic Video System}

% make the title area
\maketitle

% To allow for easy dual compilation without having to reenter the
% abstract/keywords data, the \IEEEtitleabstractindextext text will
% not be used in maketitle, but will appear (i.e., to be "transported")
% here as \IEEEdisplaynontitleabstractindextext when the compsoc 
% or transmag modes are not selected <OR> if conference mode is selected 
% - because all conference papers position the abstract like regular
% papers do.
\IEEEdisplaynontitleabstractindextext
% \IEEEdisplaynontitleabstractindextext has no effect when using
% compsoc or transmag under a non-conference mode.

% For peer review papers, you can put extra information on the cover
% page as needed:
% \ifCLASSOPTIONpeerreview
% \begin{center} \bfseries EDICS Category: 3-BBND \end{center}
% \fi
%
% For peerreview papers, this IEEEtran command inserts a page break and
% creates the second title. It will be ignored for other modes.
\IEEEpeerreviewmaketitle

\begin{figure*}
\vspace{-5pt}
% \begin{center}
\centering
    \includegraphics[width=0.95\textwidth]{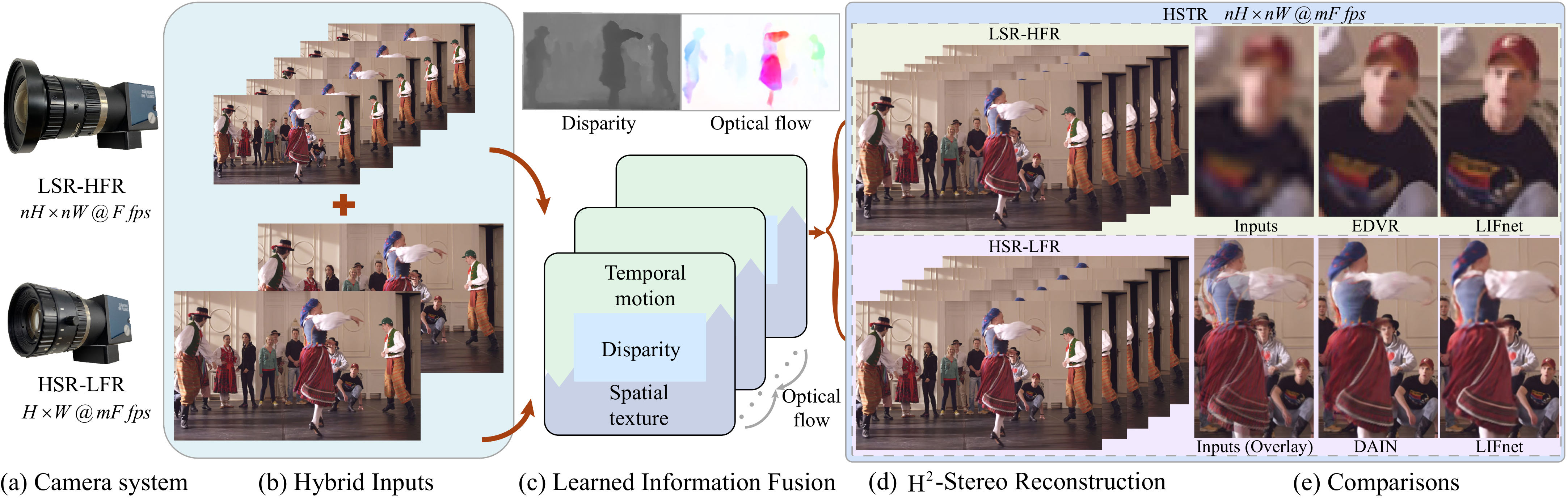}
% \end{center}
\vspace{-10pt}
    \caption{\textbf{H$^2$-Stereo system.} (a) Camera system. One is low-spatial-resolution high-frame-rate (LSR-HFR) camera and the other is high-spatial-resolution low-frame-rate (HSR-LFR) camera. (b) Hybrid inputs. One is HSR-LFR video ($nH\times nW@F$fps) and the other is LSR-HFR video ($H\times W@mF$fps). (c) Learned Information Fusion network (LIFnet). Our proposed LIFnet characterize spatial texture, temporal motion, and disparity from the hybrid inputs for the reconstruction of the H$^2$-Stereo video. (d) Reconstructed H$^2$-Stereo video frames. (e) Comparisons. We compare our method with the state-of-the-art video super-resolution method (EDVR~\cite{wang2019edvr}) and frame interpolation method (DAIN~\cite{bao2019depth}). The results show that our system preserves fine-grained spatial and temporal details better.
    \textit{Zoom in to see the details.}
    }
    \vspace{-15pt}
    \label{fig:system}
\end{figure*}

\section{Introduction}
    \IEEEPARstart{M}ultiview video in the stereoscopic~\cite{yang2018no,li2020bit,yang2017novel} or light-field format~\cite{sabater2017dataset} offers realistic display of 3D objects and scenes.
    High-resolution cameras such as high-definition (HD) or Ultra HD can capture spatial texture with rich details.
    % % %
    High-speed cameras can capture natural and scientific phenomena with thousands of frames per second (fps)~\cite{cheng2020dual}.

    High-speed, high-resolution multiview video acquisition thus allows us to see, analyze, and understand the motion of fast-moving 3D objects for potential applications in science, engineering, and consumer applications (e.g., virtual reality (VR) headsets and stereoscopic displays)~\cite{hu2017dynamic,xu2018mixed,ma2018buffer,hsc}.
    However, processing high-speed and high-resolution multiview videos in real-time for either acquisition or transmission is challenging.
    Using multiple professional high-speed, high-resolution cameras to capture multiview videos is extremely expensive.
    A single professional camera that can capture 4K at 1000 fps can cost upward of 100,000 USD, limiting the applications in practice~\cite{ixcameras,Optronis}.
    In contrast, consumer-level cameras cost less but can not shoot videos at both ultra-high temporal and spatial resolutions. 
    For example, a GoPro 7 camera that costs a few hundred US dollars can capture $960\times 540$ video at 240 fps, but 4K Ultra HD video with $3840\times 2160$ resolution can only be captured at the maximum frame rate of 30 fps. 
    To alleviate this dilemma, video enhancement methods are used to enhance the videos captured by commodity cameras.
    One approach is to capture high-speed multiview videos at low-resolution, and then super-resolve all the low-resolution views using super-resolution algorithms, such as single image super-resolution~\cite{lim2017enhanced,wan2020lightweight,esmaeilzehi2021updresnn,esmaeilzehi2021srnmsm}, video super-resolution~\cite{wang2019edvr,haris2019recurrent} or multi-view super-resolution~\cite{jeon2018enhancing,wang2019learning}.
    An alternative is to shoot high-resolution multiview videos at a low frame rate, and then perform frame interpolation~\cite{bao2019memc,bao2019depth,yan2020fine}.
    However, due to the critical loss of spatiotemporal information, it is impossible to perfectly restore the fine-grained details for high spatiotemporal resolution (HSTR) multiview videos.
    
    Recently, hybrid camera systems have emerged to alleviate the difficulty of high spatiotemporal resolution video acquisition by fusing captured frames from multiple cameras~\cite{cheng2020dual,paliwal2020deep,wang2017light,Sawhney01hybridstereo}.
    Cheng~\textit{\etal}~\cite{cheng2020dual} and Paliwa~\textit{\etal}~\cite{paliwal2020deep} use optical flow to transfer the spatial information from high-resolution view to enhance the spatial resolution of low-resolution view for high-speed, high-resolution video reconstruction.
    However, they can only spatially enhance the low-resolution view and are limited to small disparity scenes.
    None of the above discusses both spatial and temporal enhancements of multiple output views.

    In this paper, we propose a hybrid camera system for high-speed, high-resolution multiview video reconstruction.
    We first focus on the acquisition of stereoscopic video with two cameras.
    Then we demonstrate that our system can be easily extended to multiview videos, such as {light-field} video.
    There are two cameras in our system: one captures high-spatial-resolution low-frame-rate (HSR-LFR) videos, and the other one captures low-spatial-resolution high-frame-rate (LSR-HFR) videos, as shown in~Figure~\ref{fig:system}(a)(b). 
    Our method can not only reduce the burden on the cameras but can also reduce bandwidth requirements for transmission~\cite{chen2020learned,xu2019smart,cheng2017ndmp,xu2022media}. It is also inspiring for multiview video compression~\cite{lu2019learned}.

\begin{figure}[t]
% \vspace{-10pt}
	\footnotesize
% 	\tiny 
% 	\scriptsize        
	\centering
% 	\vspace{-10pt}
	\renewcommand{\tabcolsep}{1.0pt} % adjust horizontal space
	\renewcommand{\arraystretch}{0.8} % adjust vertical space
    
	\begin{tabular}{cccc}
	\includegraphics[width=0.22\linewidth]{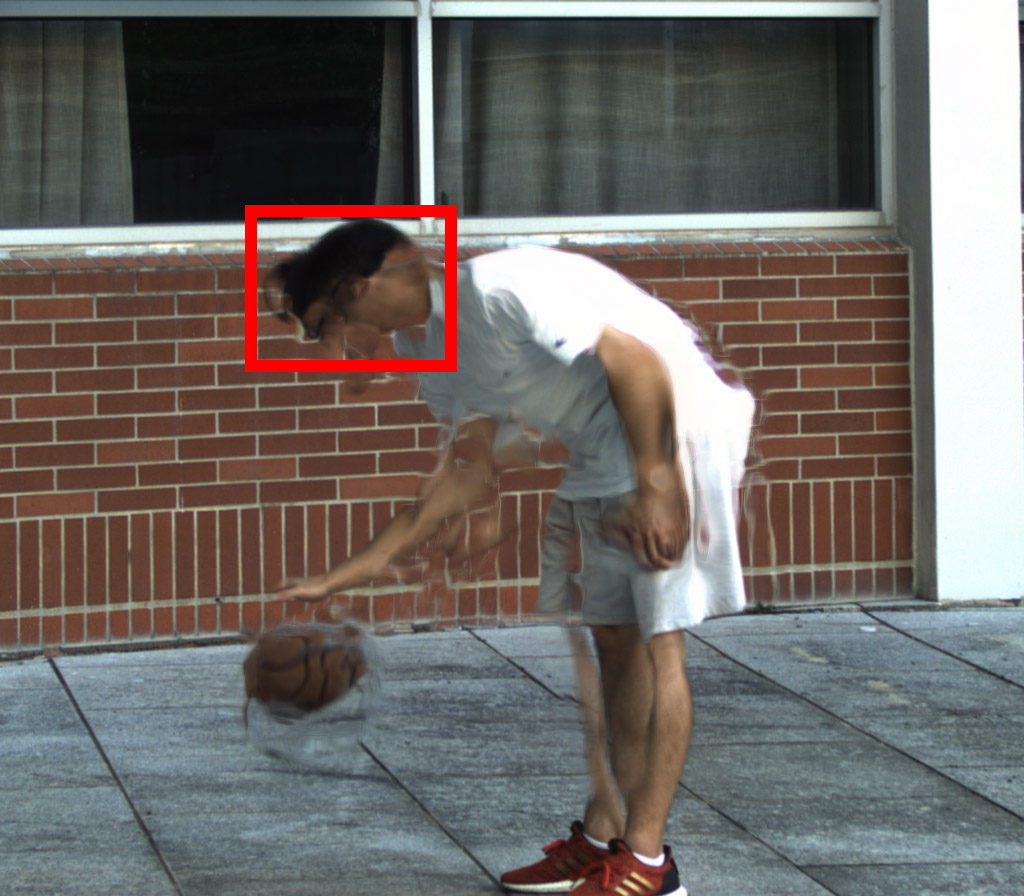} &
	\includegraphics[width=0.22\linewidth]{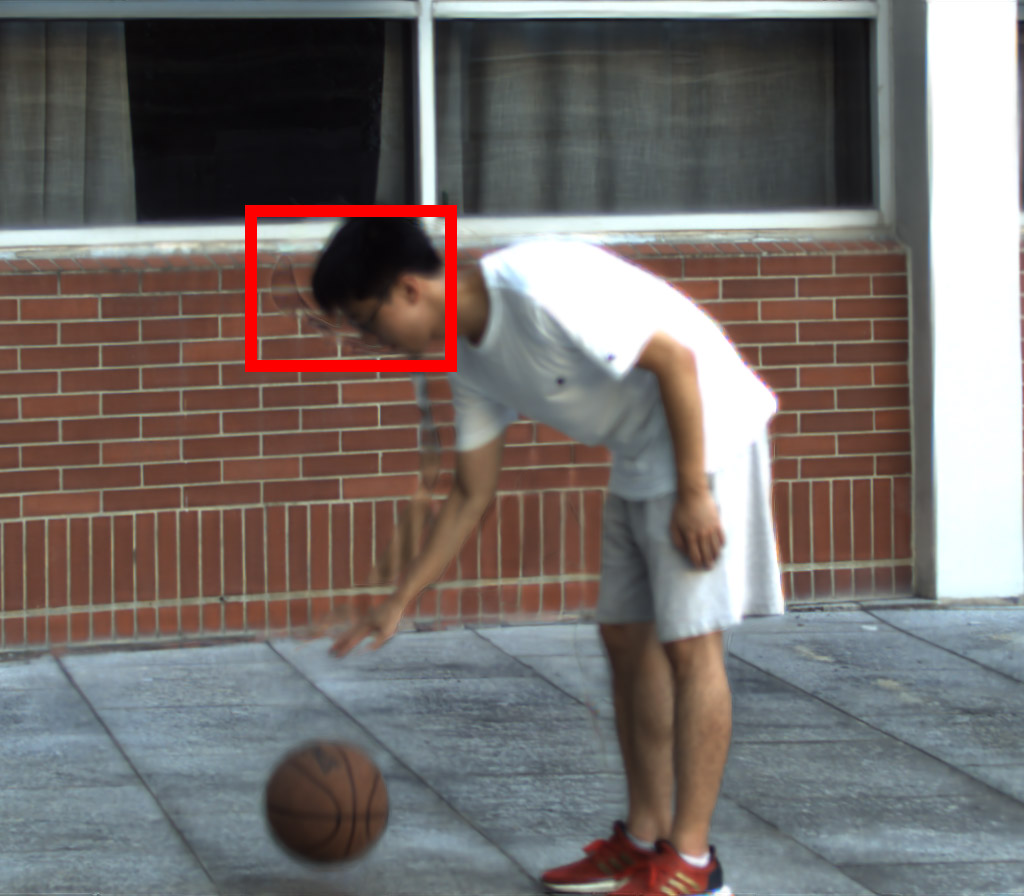} &
	\includegraphics[width=0.22\linewidth]{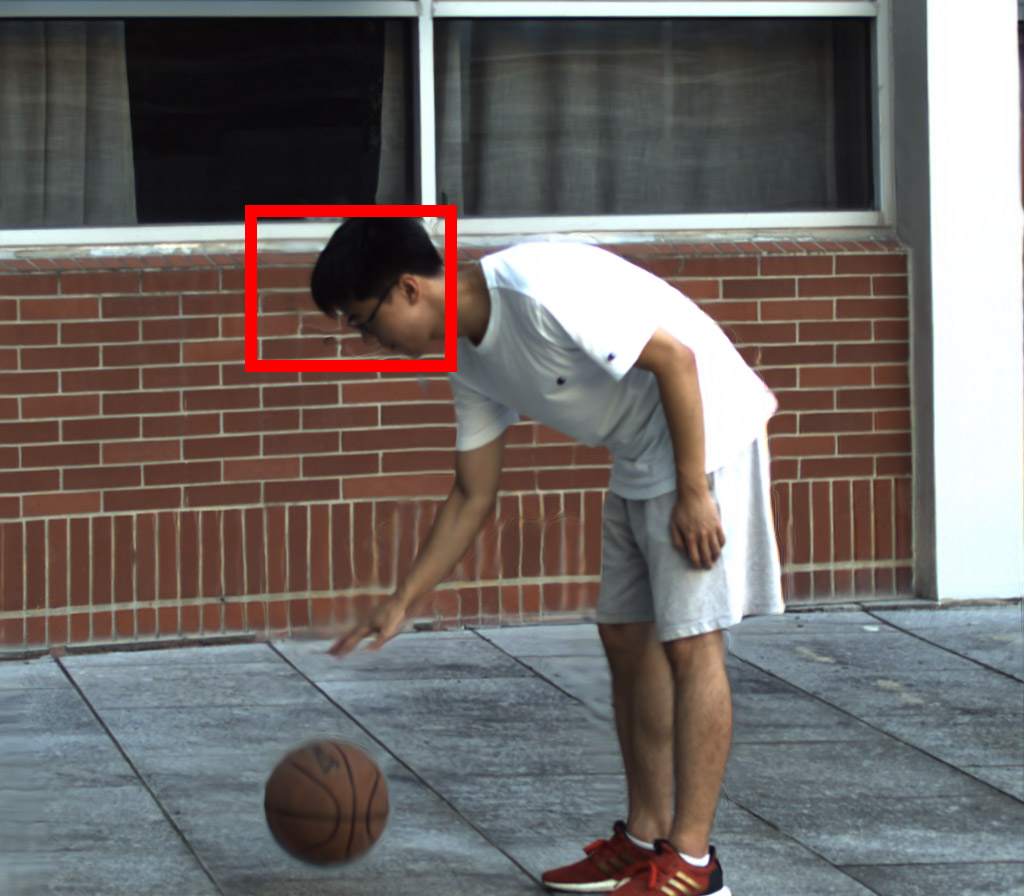} &
	\includegraphics[width=0.22\linewidth]{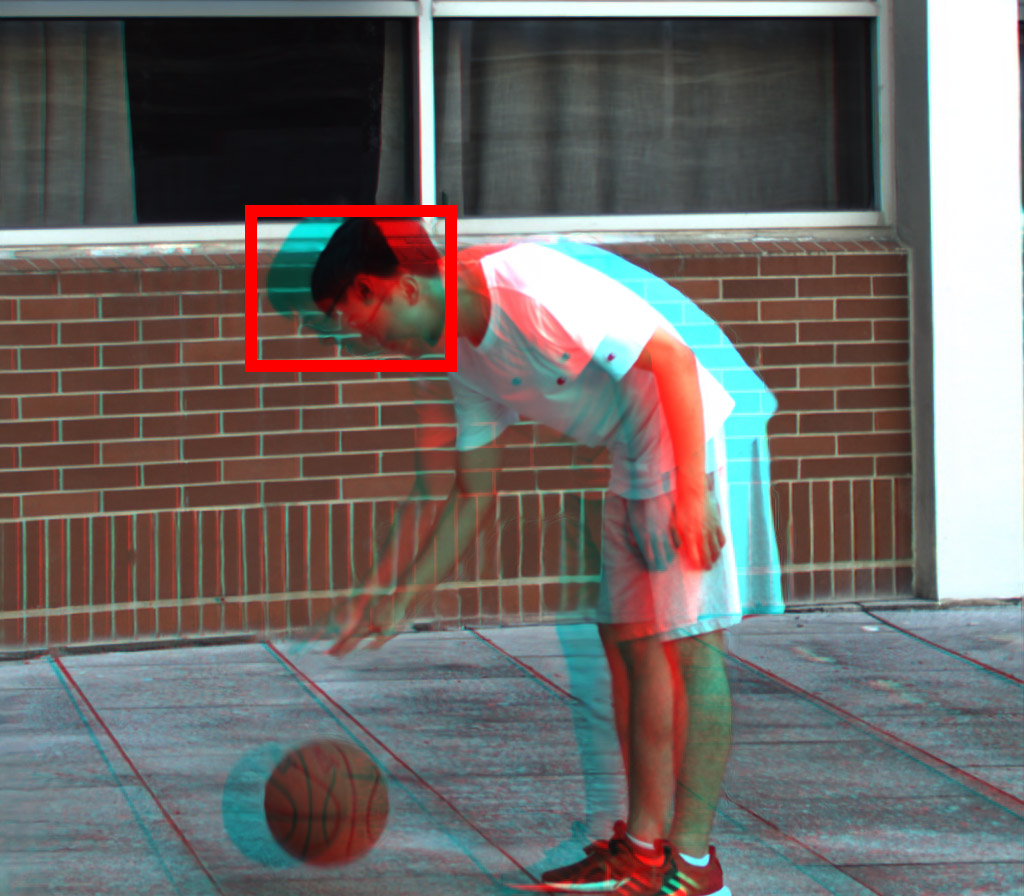}\\
	\includegraphics[width=0.22\linewidth]{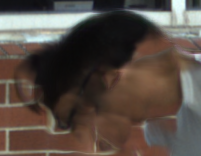} &
	\includegraphics[width=0.22\linewidth]{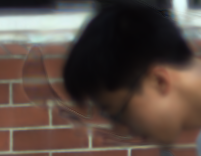} &
	\includegraphics[width=0.22\linewidth]{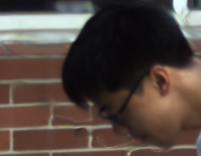} &
	\includegraphics[width=0.22\linewidth]{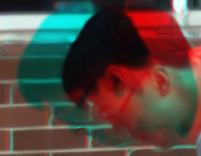}\\
	(a) & (b) & (c) & (d) \\
	\end{tabular}
    \vspace{-5pt}
	\caption{\textbf{Comparisons with Paliwa~\textit{\etal}~\cite{paliwal2020deep} and Cheng~\textit{\etal}~\cite{cheng2020dual}.} 
	(a) Result of Paliwa~\textit{\etal}~\cite{paliwal2020deep} in LSR-HFR view. (b) Result of Cheng~\textit{\etal}'s~\cite{cheng2020dual} AWnet in LSR-HFR view. (c) Result of our method in LSR-HFR view. (d) Reconstructed stereoscopic frame of our method.
	}
	\label{fig:paliwa}
	\vspace{-15pt}
\end{figure}

    There are two fundamental problems to be solved in this spatiotemporal fusion task: 
    First, how to effectively solve the problem of misalignment between LSR-HFR and HSR-LFR frames due to view difference and temporal motions?
    Second, how to remove the warping artifacts in the alignment procedure for high-quality frame reconstruction? 
    We propose a \textbf{L}earned \textbf{I}nformation \textbf{F}usion network (LIFnet) to combine the spatiotemporal information from both views to reconstruct the stereoscopic frames.
    We propose one alignment module with shared disparities and flows for both views and two fusion modules to separately integrate the warped frames for the two views.
    Existing flow-based dual-camera methods~\cite{cheng2020dual,paliwal2020deep} can well align the HSR-LFR and LSR-HFR frames if the disparity is small but fail in large disparity scenes, as shown in Figure~\ref{fig:paliwa}.
    {We propose to use disparity and flow networks to solve the problem of view difference and temporal motion separately. Compared with only using flow network for disparity and motion estimations, the introduced disparity network works well in large disparity scenes due to the geometry (one-dimensional) constraint of the disparity of the two views~\cite{lipson2021raft,slesareva2005optic}, as shown in Figure~\ref{fig:paliwa}.}
    Based on the disparity and flow networks, we propose disparity-guided flow-based warping to align the two nearest HSR-LFR frames to the LSR-HFR frame to improve the super-resolution procedure for LSR-HFR view.
    For HSR-LFR view, we propose a complementary warping strategy.
    {
    We warp two adjacent HSR-LFR frames to the intermediate time using the motion information from both views and warp the appearance information from the synchronous LSR-HFR frame to HSR-LFR view.
    }
    For image fusion, we adopt an adaptive fusion network with two-reference frames for LSR-HFR view, which has demonstrated high effectiveness in reference-based super-resolution.
    In addition, we input the disparity information into the fusion network to help distinguish the view occlusion regions.
    To alleviate occlusion-induced warping ghosts and holes, we propose to use GridNet~\cite{fourure2017residual} to fuse the aligned frames.
    To protect the multi-scale context information of the input frames, we propose to extractor trainable multi-scale features of the input frames and warp in feature domain. 
    We improve the GridNet~\cite{fourure2017residual} architecture to integrate multi scales of warped features as inputs for effective frame synthesis.
    In this way, we sequentially reconstruct the high-quality stereoscopic frames, as shown in Figure~\ref{fig:system}(d).

    We collect a high-quality stereoscopic video dataset, named Stereo Video dataset, from YouTube for training.
    We also construct a dual camera rig with large camera baseline.
    A high-speed, high-resolution stereoscopic (H$^2$-Stereo) video dataset captured by our camera rig is created for training and evaluation.
    Extensive experiments on Stereo Video dataset, H$^2$-Stereo dataset, Stereo Blur dataset~\cite{zhou2019davanet}, KITTI dataset~\cite{Menze2015CVPR} and Light Field Video dataset~\cite{guillo2018light} demonstrate the favorable performance of LIFnet against state-of-the-art methods.
    The proposed LIFnet achieves up to 1.07dB performance gain against existing state-of-the-art reference-based super-resolution method AWnet~\cite{cheng2020dual} and 8.3dB against frame interpolation method DAIN~\cite{bao2019depth}.
    Results on real data demonstrate the effectiveness of our system in practice.

    We also conduct various ablation studies to fully understand the capability of our H$^2$-Stereo system in practice, such as the impacts of the spatiotemporal resolution, camera baseline, camera desynchronization, long/short exposures and applications.
    All these studies demonstrate the effectiveness of our dual camera system for high-speed, high-resolution stereoscopic video reconstruction.

    Our main contributions can be summarized as follows:
    \begin{compactitem}
        \item We propose the first learning-based dual camera system for high-speed, high-resolution stereoscopic (H$^2$-Stereo) video acquisition, where one camera captures high-spatial-resolution low-frame-rate (HSR-LFR) videos and the other captures low-spatial-resolution high-frame-rate (LSR-HFR) videos.
        \item We propose a \textbf{L}earned \textbf{I}nformation \textbf{F}usion network (LIFnet), which utilizes disparity network to transfer spatiotemporal information across views even in large disparity scenes, and a featured-based multi-scale fusion network to minimize occlusion-induced warping ghosts and holes in HSR-LFR view. We present an image alignment system in LIFnet, including disparity-guided flow-based warping for LSR-HFR view and complementary warping for HSR-LFR view.
        \item {We construct a real dual camera rig with large baseline to demonstrate the high effectiveness of our method on real data.}
        \item Objective and subjective results and more detailed analyses on spatiotemporal resolution, camera baseline, camera desynchronization, long/short exposures and applications demonstrate the superior performance and robustness.
    \end{compactitem}

\section{Related work}

    \subsection{Super Resolution}
    Super-resolution methods, such as single-image super-resolution~\cite{lim2017enhanced,esmaeilzehi2021updresnn,esmaeilzehi2021srnmsm}, video super-resolution~\cite{xue2019video,wang2019edvr,haris2019recurrent,shen2021prediction} and multi-view image super-resolution~\cite{jeon2018enhancing,wang2019learning}, aim to restore the high-resolution images from low-resolution inputs.
    All of these approaches can not restore sharp spatial details for high-resolution reconstruction due to the severe loss of high-frequency spatial information.
    Especially when the upscaling factor is four or more, classical super-resolution methods usually lead to over-smooth results.
    To solve this problem, we use a high-resolution camera to capture the high-frequency spatial textures in another view to associate the super-resolution procedure.

    \subsection{Frame Interpolation}
    Frame interpolation algorithms synthesize missing temporal images between the input frames, including
    flow-based~\cite{bao2019depth,xue2019video,niklaus2018context,niklaus2020softmax,shen2021spatial,shen2022enhanced}, kernel-based~\cite{niklaus2017videosep,niklaus2017video,shen2020blurry,shen2020video} and phase-based methods~\cite{meyer2015phase,meyer2018phasenet}.
    However, all these algorithms are based on the basic translational and uniform motion assumption, which will be invalid in cases with complex and irregular motions.
    Therefore, these methods can not correctly interpolate the missing frames over a long temporal span with large motion.
    In our system, we use an additional high-frame-rate camera to provide temporal motions for the missing frames that enable the restoration of high-quality videos.

    \subsection{Hybrid Camera Systems}
    Hybrid camera systems, which combine different types of cameras that complement one another, have been proposed to enhance the spatiotemporal resolution of videos~\cite{cheng2020dual,paliwal2020deep,wang2017light,lin2020learning,wang2015high}.
    For instance, Sawhney~\textit{\etal}~\cite{Sawhney01hybridstereo} and Zhang~\textit{\etal}~\cite{zhang2014unified} super-resolve the low-resolution view with the help of corresponding high-resolution view to alleviate the transmission or acquisition difficulties for simultaneously processing two high-resolution and high-frame-rate streams.
    Wang~\textit{\etal}~\cite{wang2017light} combine a DSLR camera and a light field camera to enhance the resolution of light fields temporally.
    Cheng~\textit{\etal}~\cite{cheng2020dual} and Paliwal~\textit{\etal}~\cite{paliwal2020deep} propose similar camera setups, which spatially enhance the low-resolution view with the help of the high-spatial-resolution low-frame-rate view for high-speed, high-resolution video reconstruction.
    None of them discusses both spatial and temporal enhancements of multiple output views.
    
    Image alignment between multiple cameras is a critical problem in hybrid camera systems.
    Cheng~\textit{\etal}~\cite{cheng2020dual} use optical flow for preliminary alignment and dynamic filter for refinement.
    Paliwal~\textit{\etal}~\cite{paliwal2020deep} train a flow-enhancement network to refine the estimated flow.
    Their proposed flow-based alignments can partially solve the problem of small disparity, but they will fail in large disparity scenes, as shown in Figure~\ref{fig:paliwa}(a) and Figure~\ref{fig:paliwa}(b).
    We utilize disparity to transfer the spatial textures and temporal motions across views for image alignment.
    Because of the one-dimensional constraint for disparity estimation, our disparity network can solve large disparity problem compared with flow network.

    The fusion of warped frames is also a necessary module for frame reconstruction to alleviate the alignment errors and warping artifacts.
    Wang~\textit{\etal}~\cite{wang2017light} and Cheng~\textit{\etal}~\cite{cheng2020dual} integrate warped frames in pixel domain.
    Image warping in pixel domain cannot preserve the contextual information in occlusion areas.
    Especially for the frame interpolation task in HSR-LFR view, the occlusion-induced warping ghosts and holes will result in artifacts on output frames.
    Paliwal~\textit{\etal}~\cite{paliwal2020deep} fuses one scale of features extracted from the \textit{conv1} layer of ResNet-18~\cite{he2016deep} with fixed weights.
    They do not optimize the feature extractor with their training set, so the performance is limited.
    We use a trainable feature extractor for image fusion and optimize the feature extractor with our created datasets.
    Furthermore, we propose a multi-scale fusion method in feature domain to preserve more contextual information to remove occlusion-induced warping ghosts and holes.

\section{Proposed H$^2$-Stereo system and learned information fusion method}
    Our H$^2$-Stereo system is summarized in Figure~\ref{fig:system}.
    One camera captures a high-spatial-resolution low-frame-rate (HSR-LFR) video ($nH\times nW@F$ fps), and the other one captures a low-spatial-resolution high-frame-rate (LSR-HFR) video ($H\times W@mF$ fps).
    We aim to combine these two videos to synthesize a high-speed, high-resolution stereoscopic (H$^2$-Stereo) video, which contains two synchronous high spatiotemporal resolution (HSTR) videos ($nH\times nW@mF$ fps) with different views.
    {We first bicubicly upsample the LSR-HFR frames before disparity and flow estimations. Then, the upscaled LSR-HFR frames and HSR-LFR frames have the same spatial resolution.}
    We denote the bicubicly upscaled LSR-HFR frame as $\hat{L}^{t}$ and the HSR-LFR frame as $R^t$, where $t\in{\mathbb{N}}$ is the frame index, and denote two neighboring captured HSR-LFR frames by $R^0$ and $R^T$.
    Then, our problem can be formulated as: given LSR-HFR frames $\{\hat{L}^{0},\dots,\hat{L}^{T}\}$ and HSR-LFR frames $\{R^0,R^T\}$, we aim to synthesize high-resolution frames $\{\tilde{L}^0,\dots,\tilde{L}^T\}$ and $\{\tilde{R}^0,\dots,\tilde{R}^{T}\}$.
    
\begin{figure}[t]
% \vspace{-10pt}
	\footnotesize
% 	\tiny 
% 	\scriptsize        
	\centering
	\renewcommand{\tabcolsep}{0pt} % adjust horizontal space
	\renewcommand{\arraystretch}{1} % adjust vertical space
      \newcommand{\quantTit}[1]{\multicolumn{3}{c}{\scriptsize #1}}
    \newcommand{\quantSec}[1]{\scriptsize #1}
    \newcommand{\quantInd}[1]{\scriptsize #1}
    \newcommand{\quantVal}[1]{\scalebox{0.83}[1.0]{$ #1 $}}
    \newcommand{\quantBes}[1]{\scalebox{0.83}[1.0]{$\uline{ #1 }$}}
	\begin{tabular}{cc}
	\includegraphics[width=0.48\textwidth]{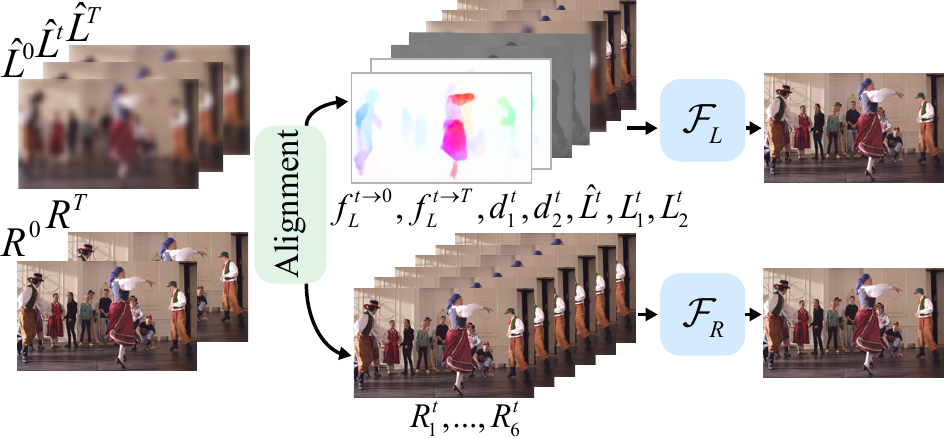} \\
	\end{tabular}
    \vspace{-5pt}
	\caption{\textbf{Learned Information Fusion network.} Our proposed Learned Information Fusion network contains one alignment module and two fusion modules, $\mathcal{F}_L$ and $\mathcal{F}_R$, for LSR-HFR and HSR-LFR views, respectively.}
	\label{fig:system_x}
	\vspace{-10pt}
\end{figure}

    To address this problem, we propose a \textbf{L}earned \textbf{I}nformation \textbf{F}usion network (LIFnet), including one alignment module and two fusion modules with separate outputs, as shown in Figure~\ref{fig:system_x} and Figure~\ref{fig:network}.
    See the green and red lines in Figure~\ref{fig:network}, we warp $R^0, R^T$ to $\hat{L}^t$ with a disparity-guided flow-based warping strategy and warp $R^0, R^T, \hat{L}^t$ to $R^t$ with a complementary warping strategy.
    The disparities and optical flows are shared for both views, as shown the black lines in Figure~\ref{fig:network}, which ensures the view and temporal consistencies of them.
    For image fusion, we propose adaptive weighting fusion for LSR-HFR view (denoted by $\mathcal{F}_L$) in Section~\ref{sec:awf} and feature-based multi-scale fusion for HSR-LFR view (denoted by $\mathcal{F}_R$) in Section~\ref{sec:fmf}.

\subsection{Alignment}\label{sec:alignment}
    
    \subsubsection{Disparity-guided flow-based warping for LSR-HFR view}
    {We propose to warp $\{R^0,R^T\}$ to $\hat{L}^t$ to assist the super-resolution process.
    The displacement between $\{R^0,R^T\}$ to $\hat{L}^t$ contains view disparity and motion flow.
    Directly flow estimation between $\{R^0,R^T\}$ to $\hat{L}^t$ is a challenging problem especially in large disparity scenes~\cite{cheng2020dual,paliwal2020deep}.
    Therefore, we propose to estimate the view disparity and motion flow separately.
    Then, the displacement is separated into view disparity and motion flow, which are smaller and easier to estimate~\cite{sun2018pwc,IMKDB17}.
    It is easier for disparity estimation than flow estimation in large disparity scenes due to the geometry (one-dimensional) constraint of disparity~\cite{lipson2021raft,slesareva2005optic}.
    Thus, we introduce a disparity network for view disparity estimation.
    Then, we concatenate the disparity and flow for image warping, namely, disparity-guided flow-based warping.
    }

\begin{figure}
% \vspace{-10pt}
\begin{center}
\includegraphics[width=0.45\textwidth]{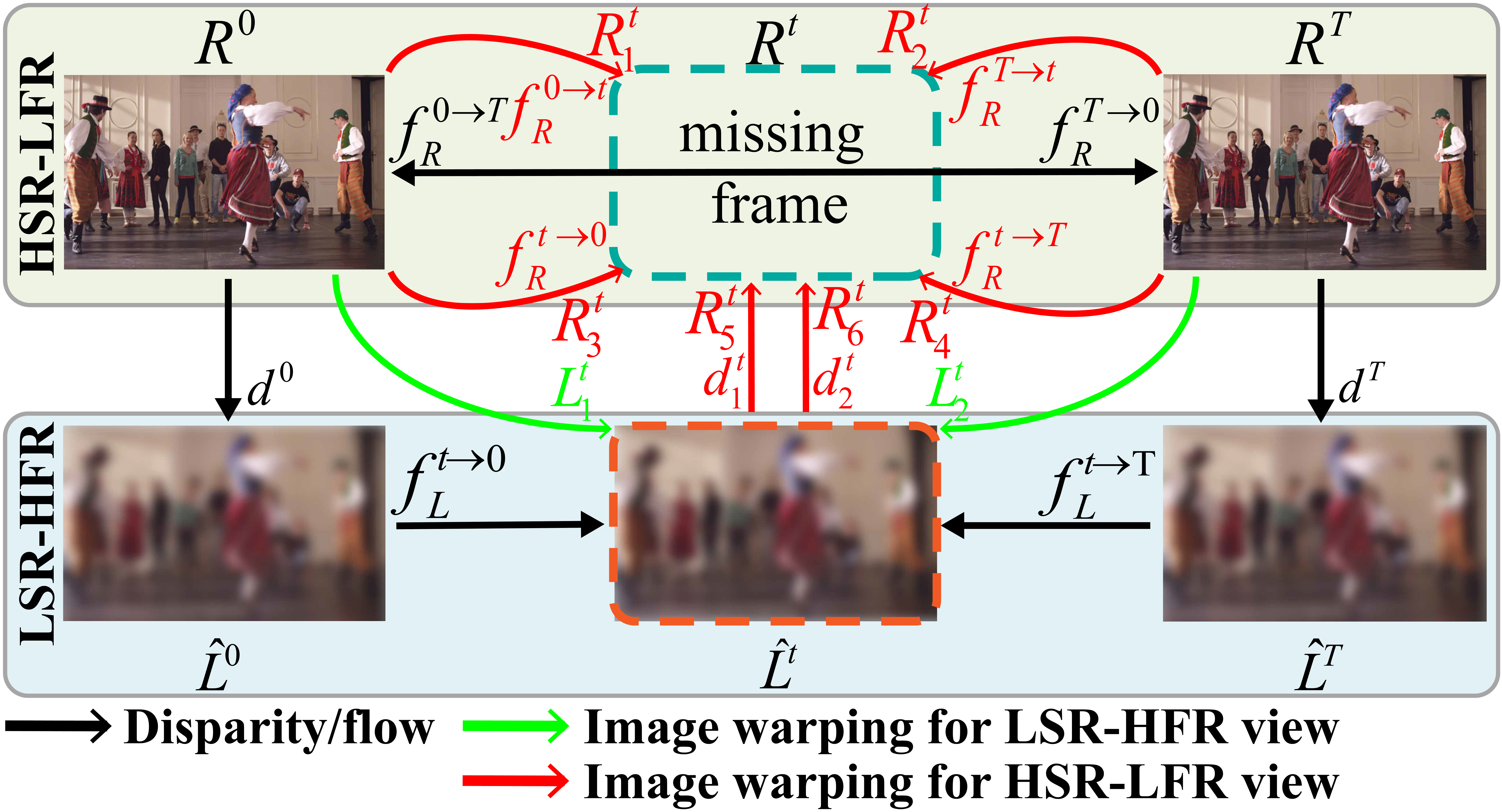}
\end{center}
  \vspace{-15pt}
  \caption{ \textbf{Alignment module.}
  The black lines illustrate the disparity and flow estimations.
  The green lines illustrate the warping process for LSR-HFR views, and red for HSR-LFR view.
  }
\label{fig:network}
\vspace{-15pt}
\end{figure}

    The black lines in Figure~\ref{fig:network} illustrate the estimations of the disparities between $R^0, R^T$ and $\hat{L}^0, \hat{L}^T$ and the optical flows from $\hat{L}^t$ to $\hat{L}^0, \hat{L}^T$.
    The disparity $d^0$ and flow $f^{t\rightarrow 0}_L$ are generated as:
    \begin{align}
        d^{0} &= {\sf DispNet}(\hat{L}^{0},R^0),\label{equ:disp1}\\
        f^{t\rightarrow 0}_L &= {\sf FlowNet}(\hat{L}^{t},\hat{L}^{0}),\label{equ:flow1}
    \end{align}
    {\flushleft where} {\sf DispNet} and {\sf FlowNet} denote the disparity and optical flow networks, respectively.
    {We adopt fast and effective disparity and flow networks to balance the computational complexity and performance, HD$^3$S~\cite{yin2019hierarchical} with one-dimensional displacement vector output for disparity estimation and PWC-Net~\cite{sun2018pwc} with two-dimensional displacement vector output for flow estimation.}
    $d^T$ and $f^{t\rightarrow T}_L$ can be formulated similarly.


    Then, we warp $R^0$ to $\hat{L}^{t}$ by concatenating $d^0$ and $f^{t\rightarrow 0}_L$ as a cascaded warping flow:
    $
        f^{t\rightarrow 0}_L+\mathcal{W}_b(d^0,f^{t\rightarrow 0}_L)
    $, where $\mathcal{W}_b$ denotes backward warping~\cite{IMKDB17}.
    {More details of the backward warping operation are presented in Section III in the supplementary material.}
    Then, we use the cascaded warping flow to warp $R^0$ to $\hat{L}^{t}$ as $L^t_1$:
    \begin{align}
        L^t_{1} = \mathcal{W}_b(R^0,f^{t\rightarrow 0}_L+\mathcal{W}_b(d^0,f^{t\rightarrow 0}_L)).\label{equ:lt1}
    \end{align}
    Similarly, we align $R^T$ to $\hat{L}^{t}$ and get the aligned frame $L^t_{2}$.
    The details of the warping process are illustrated in Figure~\ref{fig:warp}(a).

    \subsubsection{Complementary warping for HSR-LFR view}\label{sec:comwarp}
    To synthesize the missing frame $R^t$ in HSR-LFR view, we propose a complementary warping method, which consists of three complementary warping components, i.e., warping using the motion information in HSR-LFR view, warping using the motion information from LSR-HFR view and warping based on disparity.
    {Based on the warping method, we combine the motion information in LSR-HFR and HSR-LFR view and the appearance information from LSR-HFR view for effective frame interpolation.}
    We reuse the disparities and flows in Equation~\ref{equ:disp1} and Equation~\ref{equ:flow1} in the warping processes for view consistency and computational efficiency.
    
    \paragraph{{Warping using the motion information in HSR-LFR view}}
    Classic frame interpolation methods warp the input frames to the intermediate time based on bidirectional flows.
    We adopt the similar scheme that warping input frames $R^0$ and $R^T$ to time $t$ based on the bidirectional flows, {which performs effectively in static and small motion regions.}
    
\begin{figure}[t]
\begin{center}
\includegraphics[width=0.7\linewidth]{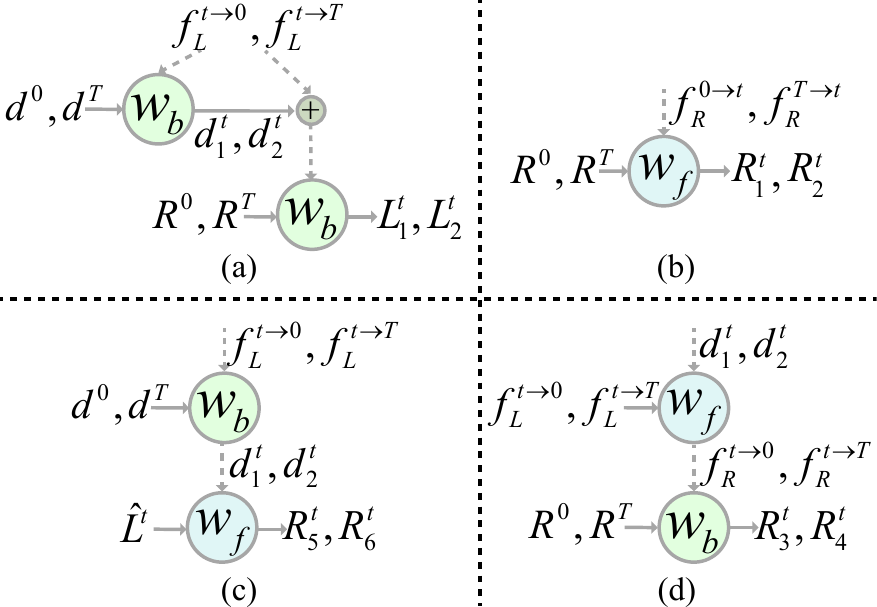}
\end{center}
  \vspace{-10pt}
  \caption{{\bf Warping details.}
  $W_b$ and $W_f$ refer to backward and forward warping respectively.
  (a) Disparity-guided flow-based warping details for LSR-HFR view.
  (b), (c) and (d) are the complementary warping details for HSR-LFR view.
  (b) We warp $R^0, R^T$ to $R^t$ based on the bidirectional flows in HSR-LFR view.
  (c) We warp $\hat{L}^t$ to $R^t$ based on the disparities at time $t$.
  (d) We warp $R^0, R^T$ to $R^t$ based on the flows from LSR-HFR view.
  }
\label{fig:warp}
 \vspace{-10pt}
\end{figure}

\begin{figure}[t]
% \vspace{-10pt}
	\footnotesize
% 	\tiny 
% 	\scriptsize        
	\centering
	\renewcommand{\tabcolsep}{1pt} % adjust horizontal space
	\renewcommand{\arraystretch}{1} % adjust vertical space
      \newcommand{\quantTit}[1]{\multicolumn{3}{c}{\scriptsize #1}}
    \newcommand{\quantSec}[1]{\scriptsize #1}
    \newcommand{\quantInd}[1]{\scriptsize #1}
    \newcommand{\quantVal}[1]{\scalebox{0.83}[1.0]{$ #1 $}}
    \newcommand{\quantBes}[1]{\scalebox{0.83}[1.0]{$\uline{ #1 }$}}
	\begin{tabular}{ccccc}
	\includegraphics[width=0.09\textwidth]{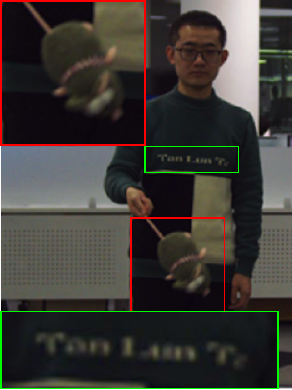} &
	\includegraphics[width=0.09\textwidth]{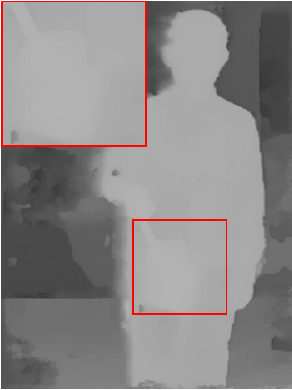} &
	\includegraphics[width=0.09\textwidth]{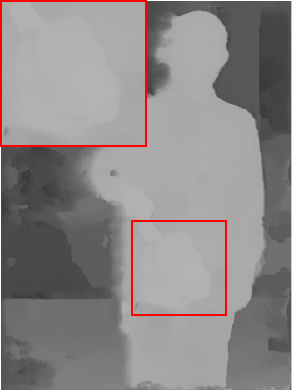} & 
	\includegraphics[width=0.09\textwidth]{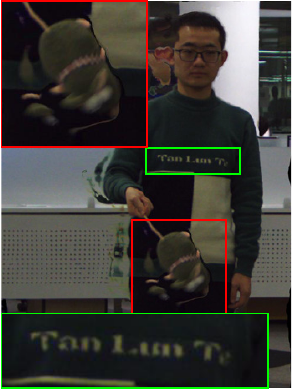} &
	\includegraphics[width=0.09\textwidth]{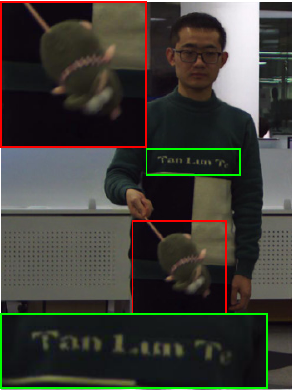}\\
	$(a) \hat{L}^t$ & (b) $d^0$ & (c) $d^t_1$ & (d) $L^t_1$ & (e) $\Tilde{L}^t$ \\
	\includegraphics[width=0.09\textwidth]{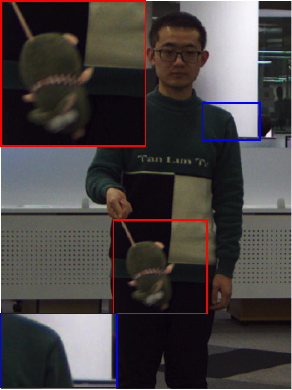} &
	\includegraphics[width=0.09\textwidth]{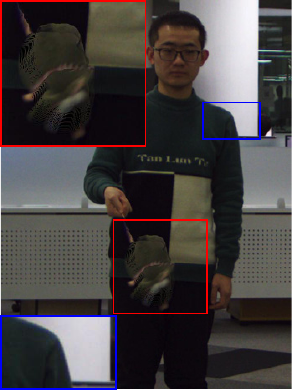} & 
	\includegraphics[width=0.09\textwidth]{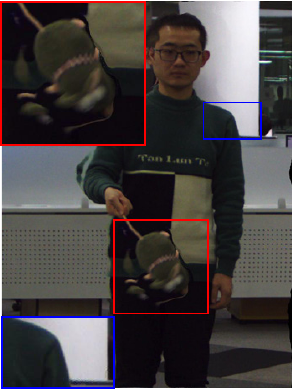} &
	\includegraphics[width=0.09\textwidth]{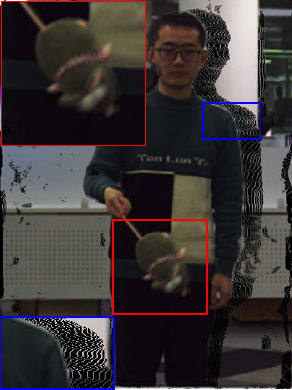} &
	\includegraphics[width=0.09\textwidth]{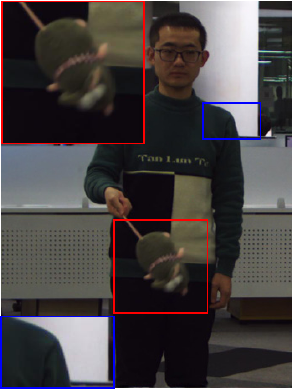} \\
	(f) $R^0$ & (g) $R^t_1$ & (h) $R^t_3$ & (i) $R^t_5$ & (j) $\Tilde{R}^t_1$\\
	\end{tabular}
    % \vspace{-5pt}
	\caption{\textbf{Inputs and outputs.} 
	(a) and (f) are the input LSR-HFR and HSR-LFR frames, respectively.
	(b) is the estimated disparity at time $0$.
	(c) is the warped disparity at time $t$.
	(d) is the warped frame in LSR-HFR view.
	(g), (h) and (i) are the warped frames in HSR-HFR view.
	(e) and (j) are the output frames of LSR-HFR and HSR-LFR views, respectively.
	}
	\label{fig:output}
	\vspace{-15pt}
\end{figure}

    The bidirectional flows $f^{0\rightarrow T}_R, f^{T\rightarrow 0}_R$ between $R^0$ and $R^T$ are generated as below:
    \begin{align}
        f^{0\rightarrow T}_R = {\sf FlowNet}(R^0,R^T),\\
        f^{T\rightarrow 0}_R = {\sf FlowNet}(R^T,R^0).
        \vspace{-5pt}
    \end{align}
    {\flushleft The} flows $f^{0\rightarrow t}_R, f^{T\rightarrow t}_R$ from $R^0$, $R^T$ to $R^t$ is calculated based on the uniform motion assumption~\cite{xue2019video,niklaus2018context}:
    \vspace{-3pt}
    \begin{align}
        f^{0\rightarrow t}_R &= t \cdot f^{0\rightarrow T}_R\label{equ:linear1},\\
        f^{T\rightarrow t}_R &= (1-t) \cdot f^{T\rightarrow 0}_R.\label{equ:linear2}
    \end{align}
    {\flushleft Next}, we use a differentiable forward warping method~\cite{niklaus2020softmax} to warp $R^0, R^T$ to $R^t$ as $R^t_{1}, R^t_{2}$:
    \begin{align}
        R^t_{1} = \mathcal{W}_f(R^0,f^{0\rightarrow t}_R)\label{equ:R1},
    \end{align}
    where $\mathcal{W}_f$ denotes forward warping. 
    {To tackle the warping conflict problem in forward warping operation~\cite{niklaus2020softmax}, which means multiple pixels splatting to one pixel, we use the property of brightness constancy to estimate the weights for the multiple splatted pixels. More details of the forward warping operation are presented in Section III in the supplementary material.}
    $R^t_{2}$ is generated similarly. 
    Figure~\ref{fig:warp}(b) illustrates the warping details based on bidirectional flows.
    Figure~\ref{fig:output}(g) is an example of $R^t_1$. 
    {In the static regions or regions with small motions, the warped frame $R^t_1$ is well aligned with the intermediate frame $R^t$. However, the shape of the fast-moving mouse is deteriorated due to flow estimation failure.}
    The reasons are large motion occlusion and suddenly changing motion vectors around the fast-moving mouse, which are challenging for flow estimation.
    In addition, the warping ghosts are severe due to the large motion occlusion.

    \paragraph{{Warping using motion information from LSR-HFR view}}
    Notice that $\hat{L}^t$ is synchronous with $R^t$, the estimated flows in LSR-HFR view from $\hat{L}^t$ to $\hat{L}^0, \hat{L}^T$ should be smaller and more accurate than the estimated bidirectional flows in HSR-LFR view.
    Therefore, we utilize the flows in LSR-HFR view to compensate for the missing motion information in HSR-LFR view.
    We transfer the flows $f^{t\rightarrow 0}_L, f^{t\rightarrow T}_L$ from LSR-HFR view to HSR-LFR view based on the disparities at time $t$.
    We cannot directly estimate the disparity at time $t$ because $R^t$ is missing. Thus, we propagate the disparity $d^0$ at time 0 to time $t$ based on the flow $f^{t\rightarrow 0}_L$ from time $t$ to time 0:
    \begin{align}
        d^t_{1} = \mathcal{W}_b(d^0,f^{t\rightarrow 0}_L).\label{equ:dt1}
    \end{align}
    Similarly, we can obtain the disparity $d^t_{2}$.
    Then, we use the disparities to warp the flows from LSR-HFR view to HSR-LFR view.
    \begin{align}
        f^{t\rightarrow 0}_R = \mathcal{W}_f(f^{t\rightarrow 0}_L,d^t_{1}).\label{equ:fr}
    \end{align}
    % }
    %
    {\flushleft We} obtain $f^{t\rightarrow T}_R$ from $R^t$ to $R^T$ similarly.
    {Here, we use the disparity information to estimate the weights for the multiple splatted pixels to solve the warping conflict problem. More details of the disparity-based forward warping process are presented in Section III in the supplementary material.}
    Then, we use $f^{t\rightarrow 0}_R$ to backward warp $R^0$ to $R^t$ as $R^t_{3}$:
    \begin{align}
        R^t_{3} = \mathcal{W}_b(R^0,f^{t\rightarrow 0}_R)\label{equ:R5},
    \end{align}
    $R^t_{4}$ is generated similarly.
    The warping details are shown in Figure~\ref{fig:warp}(d).
    From Figure~\ref{fig:output}(h), with the help of the accurate temporal flows from LSR-HFR view, the mouse has a good shape and precise location.
    {However, we also observe some slight artifacts at the edge of the mouse's forearm. Even using the flow in LSR-HFR view, flow-based alignment of the fast and small objects is still challenging due to motion occlusion.}
    
    \paragraph{Warping based on disparities}
    {In complex and large motion scenes, the flow-based alignment method may result in distortions in occlusion and edge regions around fast-moving objects, whether using motion information from LSR-HFR or HSR-LFR views. It is the limitation of flow-based alignment because of motion occlusion.}
    Notice that the LSR-HFR frame $\hat{L}^t$ is synchronous with $R^t$, which means they share the same scene, motion and shape but with a disparity.
    {Therefore, we propose to use view alignment to avoid distortions caused by motion occlusion. Because the ranges and positions of motion occlusions and view occlusions are often inconsistent with each other. The regions with severe motion occlusions may exist small view occlusions. Thus, we use well-aligned textures based on disparity to compensate for the shape of the fast-moving objects in motion occlusion regions. Then, the shapes of the fast-moving objects will be well reserved.}
    Thus, we propose to warp $\hat{L}^t$ to $R^t$ based on the disparity at time $t$ to compensate for the precise appearance of the fast-moving objects.
    To effectively utilize the appearance information from $\hat{L}^t$, we use both of $d^t_1$ and $d^t_2$ for warping.
    The subsequent fusion network will intelligently extract useful appearance information from the warped frames.
    The warped frames $R^t_{5}$ and $R^t_{6}$ are generated as:
    \begin{align}
        R^t_{5} = \mathcal{W}_f(\hat{L}^{t},d^t_{1}),\label{equ:rt3}\\
        R^t_{6} = \mathcal{W}_f(\hat{L}^{t},d^t_{2}).
    \end{align}
    Figure~\ref{fig:output}(i) is an example of $R^t_5$. 
    {Here, we use the disparity information to estimate the weights for the multiple splatted pixels to solve the warping conflict problem.
    }
    {As shown in Figure~\ref{fig:output}(i), the shape of the mouse's forearm is well reserved.}
    However, there are severe warping holes around the man due to view occlusions and forward warping.
    
    {In summary, warping using the motion information in HSR-LFR view works well in static and small motion regions. Warping using the motion information from LSR-HFR view works well in large motion regions but still may result in distortions at the edge of fast-moving objects due to motion occlusions. Warping based on disparity can avoid the warping artifacts due to motion occlusions. Thus, it works well in large motion scenes. However, the spatial information losses of the LSR-HFR frames deteriorate the quality of the disparity-based warped frames. In addition, the view occlusions result in warping ghosts and holes in the disparity-based warped frames.
    Thus, we combine the warped frames based on motion information in HSR-LFR view, the warped frames based on motion information from LSR-HFR view and the warped frames based on disparity to complement each other. Then, our method works well in all the regions with small motions, fast motions, motion occlusions and view occlusions.
    We name this warping strategy ``complementary warping strategy''.
    }

\subsection{Fusion}

    \subsubsection{Adaptive weighting fusion for LSR-HFR view}\label{sec:awf}
    We combine the warped frames $L^t_{1}, L^t_{2}$ together with $\hat{L}^{t}$ to reconstruct the high-resolution frame $\tilde{L}^t$.
    Inevitably, there will be misaligned pixels in the warped frames $L^t_{1}, L^t_{2}$ due to motion or view occlusions and the errors of disparities and flows. To effectively exploring the spatial information from the warped frames $L^t_{1}, L^t_{2}$, we adopt the adaptive weighting fusion network~\cite{cheng2020dual} for image fusion, which effectively refines the alignment and weighted fuses the warped frames $L^t_{1}, L^t_{2}$ and the low-resolution frame $\hat{L}^{t}$.
    The adaptive weighting fusion network is a U-net architecture and outputs dynamic filters and weighting masks for image fusion~\cite{cheng2020dual}.
    The adaptive weighting fusion network~\cite{cheng2020dual} weighted fuses warped frames and the low-resolution frame in pixel-domain, which can preserve the fine-grained spatial details of warped frames.
    We take disparities $d^t_1, d^t_2$ in addition to flows $f^{t\rightarrow 0}_L, f^{t\rightarrow T}_L$ into the fusion network as guidances because the disparities could help to hint at view occlusions and estimate the appropriate weights for image fusion.
    The input flow maps are used to hint at motion occlusions.
    The adaptive fusion network outputs 53 channels of features, which we use to calculate the dynamic filters and adaptive weighting masks.
    \begin{align}
        \mathbf{\mathrm{Fm}}=\mathcal{F}_L(\hat{L}^t,L^t_1,L^t_2,d^t_1,d^t_2,f^{t\rightarrow 0}_L,f^{t\rightarrow T}_L),
    \end{align}
    where $\mathbf{\mathrm{Fm}}$ denotes the output features with the same spatial size with the target frame $\tilde{L}^t$.
    We use the first 50 channels to produce two $5\times 5$ dynamic filters (one for $L^t_1$ and the other for $L^t_2$), and the last three channels to create the masks to weighted fuse $L^t_1, L^t_2$ and $\hat{L}^t$.
    The dynamic filter $K_{1}(x,y)$ for $L^t_1$ at pixel $(x,y)$ is a $5\times 5$ matrix:
    \begin{equation}
        \begin{aligned}
            K_{1}(x,y,i,j)=\mathbf{\mathrm{Fm}}(x,y,5i+j), i,j\in\{0,\dots,4\}.
        \end{aligned}
    \end{equation}
    In a similar way, we can obtain the dynamic filter $K_{2}(x,y)$ for $L^t_2$.
    We use the dynamic filters to rectify $L^t_1$ as below:
    \begin{equation}
        L^t_{1k}(x,y)=\sum\limits_{i,j=0}^{4}K_{1}(x,y,i,j)L^t_1(x-2+i,y-2+j).
    \end{equation}
    Similarly, we refine $L^t_2$ to get $L^t_{2k}$.
    We use the last three channels of $\mathbf{\mathrm{Fm}}$ to create the weighted masks:
    \begin{align}
        M(x,y,0:2)=\mathrm{softmax}(Fm(x,y,50:52)),
    \end{align}
    where $M(x,y,0:2)$ denotes the weighted masks for $L^t_{1k}$, $L^t_{2k}$ and $\hat{L}^t$, respectively.
    Then, the output frame $\tilde{L}^t$ can be formulated as:
    \begin{align}
        \tilde{L}^t(x,y) = &M(x,y,0)L^t_{1k}(x,y)+M(x,y,1)L^t_{2k}(x,y)\nonumber\\
            +&M(x,y,2)\hat{L}^t(x,y),
    \end{align}
    We use the same dynamic filter and weighted mask for all the three color channels of one frame.

\begin{figure}[t]
% \vspace{-20pt}
	\footnotesize
% 	\tiny 
% 	\scriptsize        
	\centering
	\renewcommand{\tabcolsep}{5.0pt} % adjust horizontal space
	\renewcommand{\arraystretch}{1.0} % adjust vertical space
	\begin{tabular}{cc}
	\includegraphics[width=0.35\linewidth]{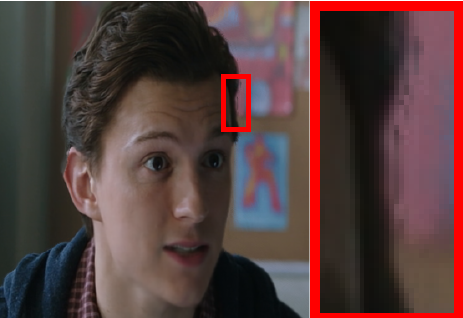} &
	\includegraphics[width=0.35\linewidth]{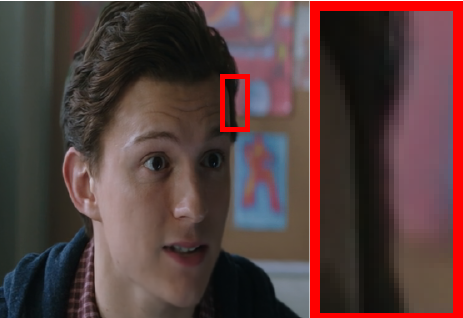} \\
	(a) Adaptive weighting fusion & (b) Feature-based multi-scale fusion\\
	\end{tabular}
    % \vspace{-5pt}
	\caption{\textbf{Fusion for HSR-LFR view.} 
	}
	\label{fig:fusion}
	\vspace{-15pt}
\end{figure} 
\begin{figure*}
\vspace{-5pt}
% \begin{center}
\centering
    \includegraphics[width=0.85\textwidth]{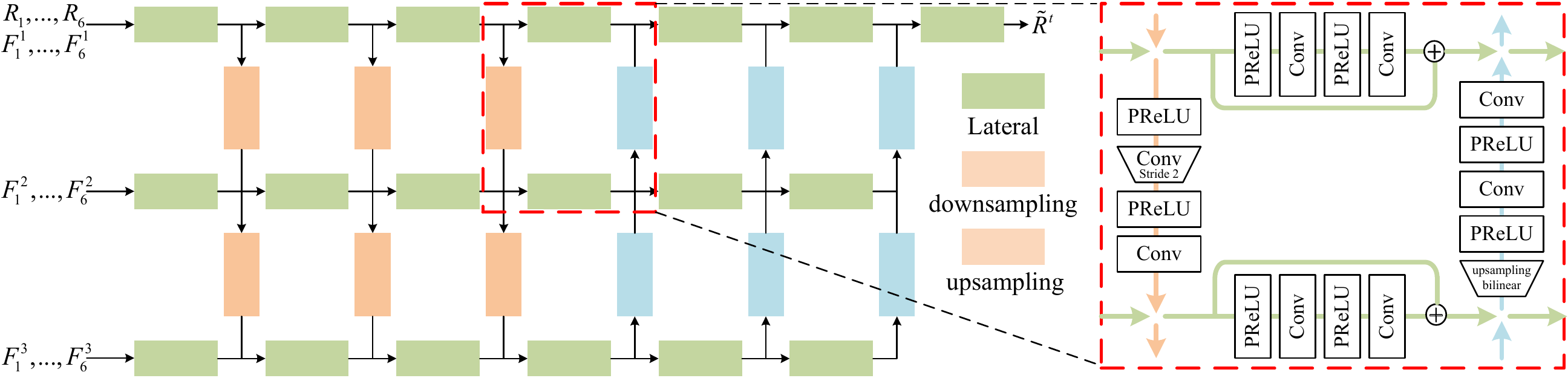}
% \end{center}
\vspace{-5pt}
    \caption{\textbf{Feature-based multi-scale fusion network.} The inputs of the feature-based multi-scale fusion network are warped frames $\{R^t_i~|~i=1,\dots,6\}$ and corresponding three scales of warped features $\{F^s_i~|~s=1,2,3,i=1,\dots,6\}$. The output is the reconstructed frame $\tilde{R}^t$.}
    \vspace{-15pt}
    \label{fig:gridnet}
\end{figure*}
 
    \subsubsection{Feature-based multi-scale fusion for HSR-LFR view}\label{sec:fmf}
    In HSR-LFR view, we fuse all the six warped frames $\{ R^t_{i}~|~{i=1,\dots,6}\}$ to synthesize $\tilde{R}^t$.
    As shown in Figure~\ref{fig:output}(g)-(i), there are severe ghosts and holes on the warped frames, especially on $R^t_1$ and $R^t_5$.
    {It is challenging for a fusion network to intelligently fuse all the warped frames, remove the warping artifacts and reconstruct accurate smooth textures.}
    The adaptive weighting fusion method, which combines the warped images in pixel-domain, can not produce smooth textures in occlusion regions with warping holes, as shown in Figure~\ref{fig:fusion}(a).
    Thus, we propose a feature-based multi-scale fusion network for HSR-LFR view.
    {Compared with warping in pixel domain, feature-based warping can reserve more context information of the warped frames and restore smoother textures. Multi-scale features further preserve more multi-scale context information.}
    Thus, before the feature warping and feature fusion, we use a convolutional feature extractor, which contains six convolution layers and two of them with the stride of 2, to extract three scales of features of the input HSR-LFR frames $R^0, R^T$ and the synchronized LSR-HFR frame $\hat{L}^t$. 
    After the warping process, we get the warped frames $\{R^t_i~|~i=1,\dots,6\}$ and corresponding three scales of warped features $\{F^s_i~|~s=1,2,3,i=1,\dots,6\}$.
    {Unet is the commonly used multi-scale fusion network, an encoder-decoder architecture with skip connections. However, the features of different scales can only propagate in a single convolutional path or skip connections.
    To further improve the interaction between multi-scale features, we adopt the GridNet~\cite{fourure2017residual} for multi-scale feature fusion,  where the multi-scale features are interconnected by convolutional blocks in a grid way.}
    We remove the batch-normalization layer following existing image enhancement works~\cite{lim2017enhanced}, because batch-normalization will lead to color distortion in testing process.
    We replace the deconvolution layers with bilinear upsampling layers to preserve smooth textures~\cite{he2015delving}. 
    We replace ReLU activation layer in the original GridNet~\cite{fourure2017residual} with PReLU~\cite{he2015delving}.
    The ``Dead ReLU'' problem limits the performance of the ReLU activation layer because of its zero slope in the negative part~\cite{he2015delving}.
    The slope of PReLU in the negative part is not zero. Therefore, it will not be ``dead''. In addition, the learnable slope in the negative part of PReLU results in more specialized activations and better representation ability.
    PReLU has demonstrated high effectiveness in various visual tasks, such as classification~\cite{he2015delving} and super-resolution~\cite{dong2016accelerating}.
    We configure three rows and six columns for GridNet~\cite{fourure2017residual} to balance the performance and computation consuming, as shown in Figure~\ref{fig:gridnet}.
    We concatenate all the warped features of $\{ R^t_{i}~|~{i=1,\dots,6}\}$ at each scale and input them into the fusion network.
    Then, the target frame $\tilde{R}^t$ can be generated as below:
    \begin{align}
        \tilde{R}^t = \mathcal{F}_R ( \{ R^t_{i},F^s_i~|~{i=1,\dots,6,s=1,2,3}\} ),
    \end{align}
    where $\mathcal{F}_R$ denotes the fusion network for HSR-LFR view.
    As shown in Figure~\ref{fig:fusion}(b), the feature-based multi-scale fusion network reconstructs smooth textures in occlusion regions.

    \subsection{Losses}

    We train our model with reconstruction losses, warping losses and smoothness loss. The overall loss is the weighted sum of them.
    
    \paragraph{Reconstruction losses}
    We use the widely used pixel-wise $\mathcal{L}_1$ loss between the generated frames $\tilde{L}^t, \tilde{R}^t$ and the ground truth frames $L^t, R^t$ to train our network.
    \begin{align}
        \mathcal{L}_L = ||L^t-\tilde{L}^t||_1,\label{equ:LSR} \\
        \mathcal{L}_R = ||R^t-\tilde{R}^t||_1,\label{equ:HSR}
    \end{align}
    where $\mathcal{L}_L$ and $\mathcal{L}_R$ refer to the reconstruction losses for LSR-HFR and HSR-LFR views, respectively.
    
    \paragraph{Warping losses}
    We use warping losses to fine-tune the disparity and optical flow networks:
    \begin{equation}
        \begin{aligned}
        \mathcal{L}_d = ||L^0-\mathcal{W}_b(R^0,d^0)||_1+||L^T-\mathcal{W}_b(R^T,d^T)||_1,\label{equ:warpdisp}
        \end{aligned}
    \end{equation}
    \begin{equation}
        \begin{aligned}
        \mathcal{L}_f = ||L^t-L^t_1||_1+||L^t-L^t_2||_1.\label{equ:warpflow}
        \end{aligned}
    \end{equation}
    
    \paragraph{Smoothness loss}
    We minimize the total variation loss on disparities to enforce the view consistency of stereo outputs:
    \begin{equation}
        \begin{aligned}
        \mathcal{L}_{s} = &||\nabla d^0||_1+||\nabla d^T||_1,\label{equ:tvloss}
        \end{aligned}
    \end{equation}
    where $\nabla$ denotes first derivative.

\begin{figure}[t]
    \centering
    \includegraphics[width=0.3\textwidth]{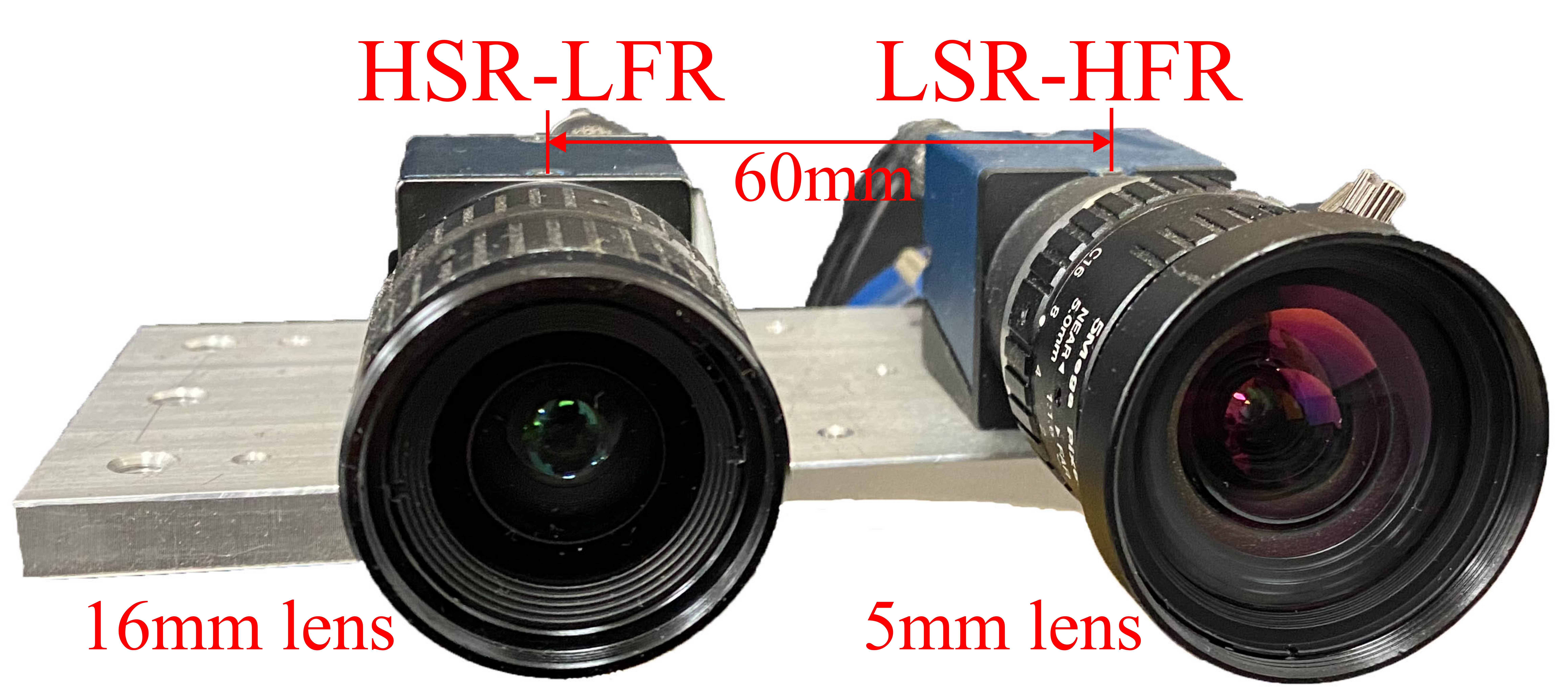}
    \caption{\textbf{Stereo camera system rig.}}
    \label{fig:cameras}
    \vspace{-15pt}
\end{figure}

\section{Experiment} 

    \subsection{Real Camera system}

    \subsubsection{Real dual camera rig}
    We use two MER-160-227U3C cameras fitted with one 16mm lens for HSR-LFR view and one 5mm lens for LSR-HFR view, as shown in Figure~\ref{fig:cameras}.
    The resolution gap between them is nearly three.
    We use the HSR-LFR camera to capture videos of $1440\times 1080@24$ fps and the LSR-HFR camera to capture videos of $720\times 540@384$ fps with region of Interest (ROI) mode.
    We fix the two cameras on a customized metal board to capture stereoscopic videos with left and right views with a large baseline of 60mm.
    A hardware trigger is adopted for time synchronization.

    \subsubsection{Camera calibration}
    We use a checkerboard and adopt Zhang's~\cite{zhang2000flexible} method for camera calibration.
    We propose a two-step strategy to improve the accuracy and robustness of calibration on our captured heterogeneous stereo pair.
    The first step is zoom-compensation.
    We first use the Camera Calibration Toolbox in Matlab to estimate the focal lengths of both views.
    We upscale the LSR-HFR images with the HSR-LFR to LSR-HFR focal length ratio.
    Then, we crop the center $1440\times 1080$ patch from the up-scaled LSR-HFR images.
    After the zoom-compensation, we obtain homogeneous stereo pairs.
    In the second step, we reuse the Camera Calibration Toolbox for the final camera calibration and stereo rectification.
    
    \subsubsection{Color calibration}
    Multi-camera systems will face inconsistent color responses across sensors, leading to unsatisfactory performance.
    We utilize Ding~\textit{\etal}'s color calibration method~\cite{ding2020multi}, which can effectively prevent over exposure.

    \subsection{Stereo Video Dataset and H$^2$-Stereo Dataset}
    The existing stereo video datasets, such as KITTI~\cite{Menze2015CVPR}, Cityscapes~\cite{cordts2016cityscapes} and Stereo Blur dataset~\cite{zhou2019davanet}, are not suitable for our task due to their limited scenes, low frame rate and low image quality.
    To train and evaluate our method in wide variety of scenes and real-world scenes, we create two high-quality datasets.
    One is Stereo Video dataset, which is collected from YouTube.
    The other is a High-speed, high-resolution Stereoscopic (H$^2$-Stereo) video dataset, which is captured by our constructed dual camera rig with large camera baseline.
    
    \subsubsection{Stereo Video Dataset}
    We download 100 high-quality stereoscopic videos from YouTube to construct a stereoscopic video dataset, named Stereo Video Dataset.
    As many videos contain shot cut, we use a threshold-based shot detection algorithm to break each video into multiple shots~\cite{xue2019video}.
    Further, we use the GIST feature~\cite{oliva2001modeling} to remove shots with similar scene contents.
    As a result, our Stereo Video dataset consists of 40509 stereo image pairs in total, including 1929 diverse clips of dynamic scenes and 21 frames in each clip.
    We resize all frames to the fixed resolution of $960\times 540$ to {improve per-pixel quality} and standardize the input.
    We randomly select 400 clips for testing and the remaining for training.
    
    \subsubsection{High-speed, High-resolution Stereoscopic (H$^2$-Stereo) Video Dataset}
    We use our dual-camera rig fitted with two 16mm lenses to capture a high-speed, high-resolution ($1440\times 1080@200$ fps) stereoscopic video dataset with large disparity.
    We directly store raw video data after the acquisition to speed up the video acquisition and storage. 
    ISP processing is implemented on a computer after the acquisition process.
    We use a fixed and small aperture for a large depth of field.
    Due to the small aperture and limited exposure time, there will be noise on the captured videos. 
    Therefore, we use the state-of-the-art video denoising algorithm, fastDVD~\cite{Tassano_2020_CVPR}, to denoise our captured videos.
    There are different kinds of motions in our dataset, including object movements, camera shakes and mixed movements.
    There are 24 sequences and 600 frames for each sequence, where 20 videos are in the training set and the other four videos are in the testing set.
    To the best of our knowledge, this dataset is the first high-speed, high-resolution stereoscopic (H$^2$-Stereo) video dataset currently available.

\subsection{Training}
    
\subsubsection{Training data}

    We randomly extract a sequence with continuous 3, 5 or 7 frames from Stereo Video dataset as a training sample.
    We set the first and the last frames in the right view as HSR-LFR frames.
    In this way, we reduce the frame rate of the right view from its original HSTR video by $2\times$, $4\times$ or $6\times$ to generate the HSR-LFR input.
    For the LSR-HFR frames, we downsample all the frames in the left view by $4\times$, $6\times$ or $8\times$ using \textit{bicubic} method.
    The training patch size is $576\times 448$.
    We augment the training data by horizontal and vertical flipping as well as reversing the temporal order.

    Our model trained on synthetic data can not be applied to real data directly due to the domain gap between synthetic data and real data.
    Pre-denoising and color calibration are generally helpful for our system but not enough.
    Thus, we augment the LSR-HFR training data by adding random noises and color jitters to further reduce the domain gap and increase the robustness of our network on real data, which has been adopted in existing real-world enhancement works~\cite{paliwal2020deep,cheng2020dual}.

\begin{table*}[t]
% \vspace{-20pt}
\caption{\textbf{Quantitative evaluations on stereoscopic video datasets.} 
The "--" in the table means that we did not test it.}
\label{table:stereoresults}
\centering
\footnotesize
  \newcommand{\quantTit}[1]{\multicolumn{3}{c}{\scriptsize #1}}
    \newcommand{\quantSec}[1]{\scriptsize #1}
    \newcommand{\quantInd}[1]{\scriptsize #1}
    \newcommand{\quantVal}[1]{\scalebox{0.83}[1.0]{$ #1 $}}
    \newcommand{\quantBes}[1]{\scalebox{0.83}[1.0]{$\uline{ #1 }$}}
\renewcommand{\tabcolsep}{3.5pt} % adjust horizontal space
\renewcommand{\arraystretch}{0.9} % adjust vertical space
\vspace{-5pt}
\begin{tabular}{ l c c c c c c c c c c c c c c c c}
\toprule
\multirow{3}{*}{Methods} & \multicolumn{4}{c}{H$^2$-Stereo ($1440\times 1080$)} & \multicolumn{4}{c}{Stereo Blur~\cite{zhou2019davanet} ($1280\times 720$)} & \multicolumn{4}{c}{Stereo Video ($960\times 540$)} & \multicolumn{4}{c}{KITTI~\cite{Menze2015CVPR} ($1242\times 375$)}\\
\cmidrule{3-4} \cmidrule{7-8} \cmidrule{11-12} \cmidrule{15-16}
 & \multicolumn{2}{c}{LSR-HFR} & \multicolumn{2}{c}{HSR-LFR} & \multicolumn{2}{c}{LSR-HFR} & \multicolumn{2}{c}{HSR-LFR} & \multicolumn{2}{c}{LSR-HFR} & \multicolumn{2}{c}{HSR-LFR} & \multicolumn{2}{c}{LSR-HFR} & \multicolumn{2}{c}{HSR-LFR}\\
\cmidrule{2-17}
& PSNR & SSIM & PSNR & SSIM & PSNR & SSIM & PSNR & SSIM & PSNR & SSIM & PSNR & SSIM & PSNR & SSIM & PSNR & SSIM\\
    \midrule
    EDVR~\cite{wang2019edvr}
        & 39.42      
        & 0.9581
        &--
        &--
        & 32.58     
        & 0.9373
        & --
        & --
        & 37.56      
        & 0.9777
        &--
        &--
        & 25.10     
        & 0.8675
        &--
        &--
        \\
    RBPN~\cite{haris2019recurrent}
        & 39.07
        & 0.9573
        &--
        &--
        & 32.63
        & 0.9373
        &--
        &--
        &36.30
        &0.9718
        &--
        &--
        & 25.63
        & 0.8834
        &--
        &--
        \\
    %  \midrule
    %
    CrossNet~\cite{zheng2018crossnet}
        & 39.19
        & 0.9613
        &--
        &--
        & 32.63
        & 0.9287
        &--
        &--
        &39.58
        &0.9850
        &--
        &--
        & 25.49
        & 0.8775
        &--
        &--
        \\
    AWnet~\cite{cheng2020dual}
        & \underline{41.18}
        & \underline{0.9779}
        &--
        &--
        & \underline{34.77}
        &\underline{0.9542}
        &--
        &--
        &\underline{40.15}
        &\textbf{0.9880}
        &--
        &--
        & \underline{26.27}
        & \underline{0.9011}
        &--
        &--
        \\
    \midrule
    DAIN~\cite{bao2019depth}
        &--
        &--
        & {36.22}
        & {0.9398}
        &--
        &--
        & {22.16}
        & {0.7202}
        &--
        &--
        & {33.29}
        & {0.9456}
        &--
        &--
        & {22.80}
        & \underline{0.8518}
        \\
    BMBC~\cite{park2020bmbc}
        &--
        &--
        & 34.97
        & 0.9266
        &--
        &--
        & 20.70
        & 0.6924
        &--
        &--
        &33.03
        &0.9417
        &--
        &--
        & 21.48
        & 0.7475
        \\
    \midrule
    StereoSR~\cite{jeon2018enhancing}
        & 34.27
        & 0.9213
        & 33.51
        & 0.9145
        & 29.12
        & 0.8958
        & 29.15
        & 0.8973
        & 33.69
        & 0.9521
        & 33.68
        & 0.9521
        & 23.96
        & 0.7737
        & \underline{24.90}
        & 0.7856
        \\
    PASSRnet~\cite{wang2019learning}
        & 37.59
        & 0.9484
        & \underline{36.90}
        & \underline{0.9447}
        & 31.78
        & 0.9215
        & \textbf{31.82}
        & \underline{0.9226}
        & 34.84
        & 0.9599
        & \underline{34.86}
        & \underline{0.9598}
        & 25.13
        & 0.7956
        & \textbf{26.22}
        & 0.8288
        \\
    \midrule
        LIFnet (Ours)
        & \textbf{42.25}
        & \textbf{0.9798}
        & \textbf{40.37}
        & \textbf{0.9798}
        & \textbf{34.97}
        & \textbf{0.9552}
        & \underline{30.46}
        & \textbf{0.9303}
        &\textbf{40.22}
        &\textbf{0.9879}
        &\textbf{39.19}
        &\textbf{0.9843}
        & \textbf{26.40}
        & \textbf{0.9014}
        & 24.07
        & \textbf{0.8807}
        \\
\bottomrule
\end{tabular}
\vspace{-15pt}
\end{table*}    
\begin{table}[t]
% \vspace{-2pt}
\caption{\textbf{Quantitative evaluations on light field video dataset~\cite{guillo2018light} ($960\times 540$) in psnr $\textup{[dB]}$.}}
% \vspace{-5pt}
\label{table:LFV}
\centering
\footnotesize
      \newcommand{\quantTit}[1]{\multicolumn{3}{c}{\scriptsize #1}}
    \newcommand{\quantSec}[1]{\scriptsize #1}
    \newcommand{\quantInd}[1]{\scalebox{0.95}[1.0]{\scriptsize #1}}
    \newcommand{\quantVal}[1]{\scalebox{0.83}[1.0]{$ #1 $}}
    \newcommand{\quantBes}[1]{\scalebox{0.83}[1.0]{$\uline{ #1 }$}}
\renewcommand{\tabcolsep}{1.0pt} % adjust horizontal space
\renewcommand{\arraystretch}{0.9} % adjust vertical space
\vspace{-5pt}
\begin{tabular}{lcccccc}
\toprule
Methods & \multicolumn{2}{c}{Boxer} & \multicolumn{2}{c}{Chess} & \multicolumn{2}{c}{ChessPieces} \\
\cmidrule{2-7}
  & \quantInd{LSR-HFR$|$} & \quantInd{HSR-LFR$|$} 
  & \quantInd{LSR-HFR$|$} & \quantInd{HSR-LFR$|$} 
  & \quantInd{LSR-HFR$|$} & \quantInd{HSR-LFR} \\
    \midrule 
    EDVR~\cite{wang2019edvr}
        & 31.18 & --
        & 32.96 & -- 
        & 35.42 & --
        \\
    RBPN~\cite{haris2019recurrent}
        & 32.41&--
        & 34.11&--
        & 35.21&--
        \\
    % \midrule
    %
    CrossNet~\cite{zheng2018crossnet}
        & 36.07&--
        & 36.61&--
        & 33.82&--
        \\
    AWnet~\cite{cheng2020dual}
        & \underline{36.37}&--
        & \underline{37.05}&--
        & \underline{35.44}&--
        \\
    \midrule
    DAIN~\cite{bao2019depth}
        & --& \underline{33.70}
        & --& \underline{35.18}
        & --&24.81
        \\
    BMBC~\cite{park2020bmbc}
        & --&33.64
        & --&33.48
        & --&24.74
        \\
    \midrule
    StereoSR~\cite{jeon2018enhancing}
        & 29.12 & 28.48
        & 30.47 & 29.72
        & 31.49 & 31.07
        \\
    PASSRnet~\cite{wang2019learning}
        & 28.74 & 28.15
        & 30.68 & 29.96
        & 31.53 &  \underline{31.10}
        \\
    \midrule  
        LIFnet (Ours)
        & \textbf{36.53}&\textbf{37.40}
        & \textbf{37.13}&\textbf{37.76}
        & \textbf{35.69}&\textbf{34.99}
        \\
\bottomrule
\end{tabular}
\vspace{-16pt}
\end{table}

    \subsubsection{Training scheme}
    We utilize the Adam~\cite{kingma2014adam} optimizer with $\beta_1=0.9$ and $\beta_2=0.999$.
    We use a batch size of 2 and initialize the learning rate of fusion networks to $1e^{-4}$ and disparity and optical flow networks to $1e^{-6}$.
    We first fine-tune the pre-loaded disparity and flow networks for 50k iterations with warping and smoothness loss in Equation~(\ref{equ:warpdisp}), Equation~(\ref{equ:warpflow}) and Equation~(\ref{equ:tvloss}).
    The weights for $\mathcal{L}_d$, $\mathcal{L}_f$ and $\mathcal{L}_s$ are 1.0, 1.0 and 0.005, respectively.
    Then, we train the fusion modules using reconstruction losses in~Equation~(\ref{equ:LSR}) and~Equation~(\ref{equ:HSR}) for 300k iterations. 
    We set the weights for $\mathcal{L}_L$ and $\mathcal{L}_R$ both to 1.0.
    Finally, we fine-tune the whole network with a reduced learning rate by a factor of 0.1 for another 200k iterations with reconstruction losses.

\subsection{Evaluations}
    
\subsubsection{Anchors}
    
    We evaluate our method with synthetic data and real data.
    The inputs of LIFnet are LSR-HFR and HSR-LFR videos, and the outputs are two views of HSTR videos.
    For comparisons, we use video super-resolution methods (EDVR~\cite{wang2019edvr}, RBPN~\cite{haris2019recurrent}) and reference-based super-resolution methods (CrossNet~\cite{zheng2018crossnet}, AWnet~\cite{cheng2020dual}) to reconstruct high-resolution frames in LSR-HSR view, and use frame interpolation methods (DAIN~\cite{bao2019depth}, BMBC~\cite{park2020bmbc}) to reconstruct the missing frames in HSR-LFR view.
    We also compare our method with stereo image super-resolution methods (StereoSR~\cite{jeon2018enhancing}, PASSRnet~\cite{wang2019learning}). 
    {We take two LSR-HFR videos as the inputs of stereo image super-resolution methods for comparison. The outputs are two high spatiotemporal resolution videos.}
    We use two reference frames for AWnet~\cite{cheng2020dual} for a fair comparison.
    We retrain the reference-based super-resolution methods Crossnet~\cite{wang2019learning} and AWnet~\cite{cheng2020dual} with Stereo Video dataset.
    
\subsubsection{Datasets}
    We first implement quantitatively evaluations on stereoscopic video datasets. 
    Then, we extend our method to light-field multiview videos.
    Subjective results on real data are also given in this section.
    
    \paragraph{Stereo Video dataset}
    The testset of Stereo Video dataset consists of 400 clips.
    We downscale the resolution of the left view by $4\times$ and reduce the frame rate of the right view by $4\times$ to simulate LSR-HFR and HSR-LFR inputs.

    \paragraph{Stereo Blur dataset}
    Zhou~\textit{\etal}~\cite{zhou2019davanet} use ZED stereo camera~\cite{Stereolabs} to create a stereoscopic video dataset of $1280\times 720@60$ fps, named Stereo Blur dataset.
    We downscale the resolution of the left view by $4\times$ and reduce the frame rate of the right view by $2\times$ for testing.

    \paragraph{KITTI}
    The KITTI dataset~\cite{Menze2015CVPR} is captured by vehicular dual camera system with the image size of $1242\times 375$ at 10 fps. We downscale the resolution of the left view by $4\times$ and reduce the frame rate of the right view by $2\times$ for testing.
    
    \paragraph{H$^2$-Stereo dataset}
    We downscale the left view of H$^2$-Stereo testset by $4\times$ and reduce the frame rate of the right view by $8\times$ to simulate LSR-HFR and HSR-LFR videos.
    
    \paragraph{Light Field Video dataset}
    Guillo~\textit{\etal}~\cite{guillo2018light} collect a Light Field Video dataset with $5\times 5$ angular resolution, which is captured by an R8 Raytrix camera fitted with a 35mm lens.
    We set $(0,0)$ as the LSR-HFR view, and $(1,0)$ as the HSR-LFR view.
    We downscale the resolution of LSR-HFR frames by $4\times$ and reduce the frame rate of HSR-LFR videos by $10\times$.
    In the same way, we combine the views $\{(0,0),\dots,(0,4)\}$ with the views $\{(1,0),\dots,(1,4)\}$ to generate a multiview video with 5 stereoscopic video pairs for evaluation.
    
    \paragraph{Real data}
    We use our dual camera setup to capture real videos of different moving scenes for evaluation, such as moving cars, playing basketball, jumping and so on.

\subsubsection{Results}
    The objective evaluations and subjective comparisons are shown in this section.
    
    \paragraph{Objective results}
    As shown in Table~\ref{table:stereoresults} and~Table~\ref{table:LFV}, our method outperforms all the spatiotemporal enhancement methods in terms of PSNR and SSIM on a wide variety of datasets.
    Our method achieves significant gains against the state-of-the-art super-resolution and frame interpolation methods, 2.66dB gain against EDVR~\cite{wang2019edvr} and 5.9dB gain against DAIN~\cite{bao2019depth} on Stereo Video dataset.
    The reason is that EDVR and DAIN can not exploit the compensated spatiotemporal information from the other camera.
    In contrast, our LIFnet combines the complementary spatiotemporal information from both cameras, which significantly contributes to reconstructing of both views.
    Compared with existing reference-based super-resolution methods, our LIFnet achieves significant performance gains on large disparity scenes with the help of disparity network (3.06dB and 1.07dB gains against Crossnet~\cite{zheng2018crossnet} and AWnet~\cite{cheng2020dual}, respectively, on H$^2$-Stereo dataset).
    From the results, we can also find that the larger the image size, the more significant the gain of our method in the LSR-HFR view. Because larger images tend to have larger disparities.
    Thanks to the disparity network, our method demonstrates high effectiveness in large disparity scenes.
    {The stereo image super-resolution methods have explored the spatial complementarity of the input two views. However, they still can not reconstruct high-quality, high-frequency details due to spatial information losses on both views. Thanks to our hybrid inputs, our method achieves 4.86dB gain on average against PASSRnet~\cite{wang2019learning} on Stereo Video dataset.}

    From the results of KITTI~\cite{Menze2015CVPR}, the performances of all the methods are low because of complex outdoor lighting, fast car movement and too low frame rate of KITTI~\cite{Menze2015CVPR}.
    Nevertheless, our method outperforms all the other methods.

\begin{figure}[t]
% \vspace{-3pt}
	\footnotesize
% 	\tiny 
% 	\scriptsize        
	\centering
	\renewcommand{\tabcolsep}{0.8pt} % adjust horizontal space
	\renewcommand{\arraystretch}{1} % adjust vertical space
      \newcommand{\quantTit}[1]{\multicolumn{3}{c}{\scriptsize #1}}
    \newcommand{\quantSec}[1]{\scriptsize #1}
    \newcommand{\quantInd}[1]{\scriptsize #1}
    \newcommand{\quantVal}[1]{\scalebox{0.83}[1.0]{$ #1 $}}
    \newcommand{\quantBes}[1]{\scalebox{0.83}[1.0]{$\uline{ #1 }$}}
	\begin{tabular}{ccccc}
	\includegraphics[width=0.12\textwidth]{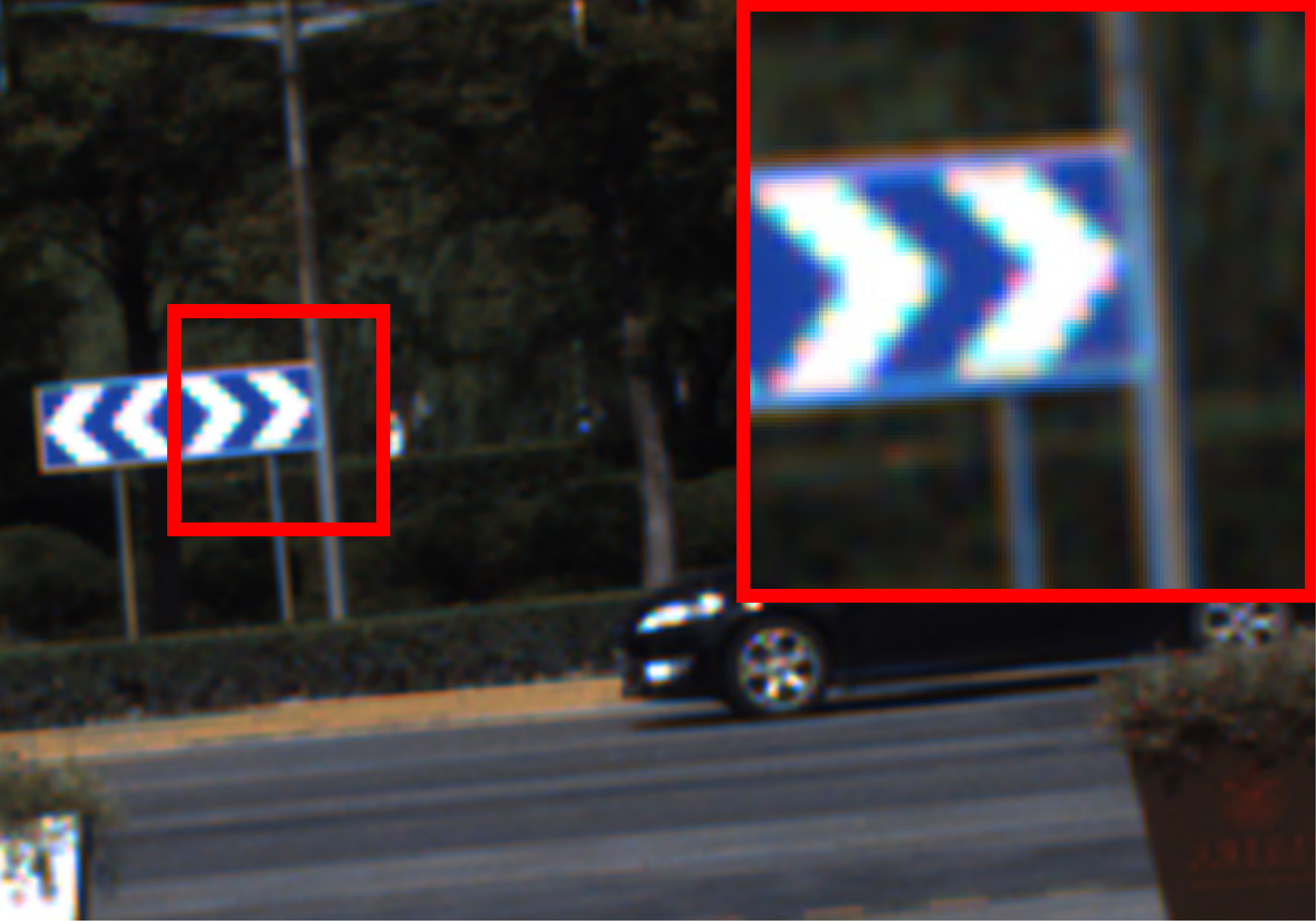} &
    \includegraphics[width=0.12\textwidth]{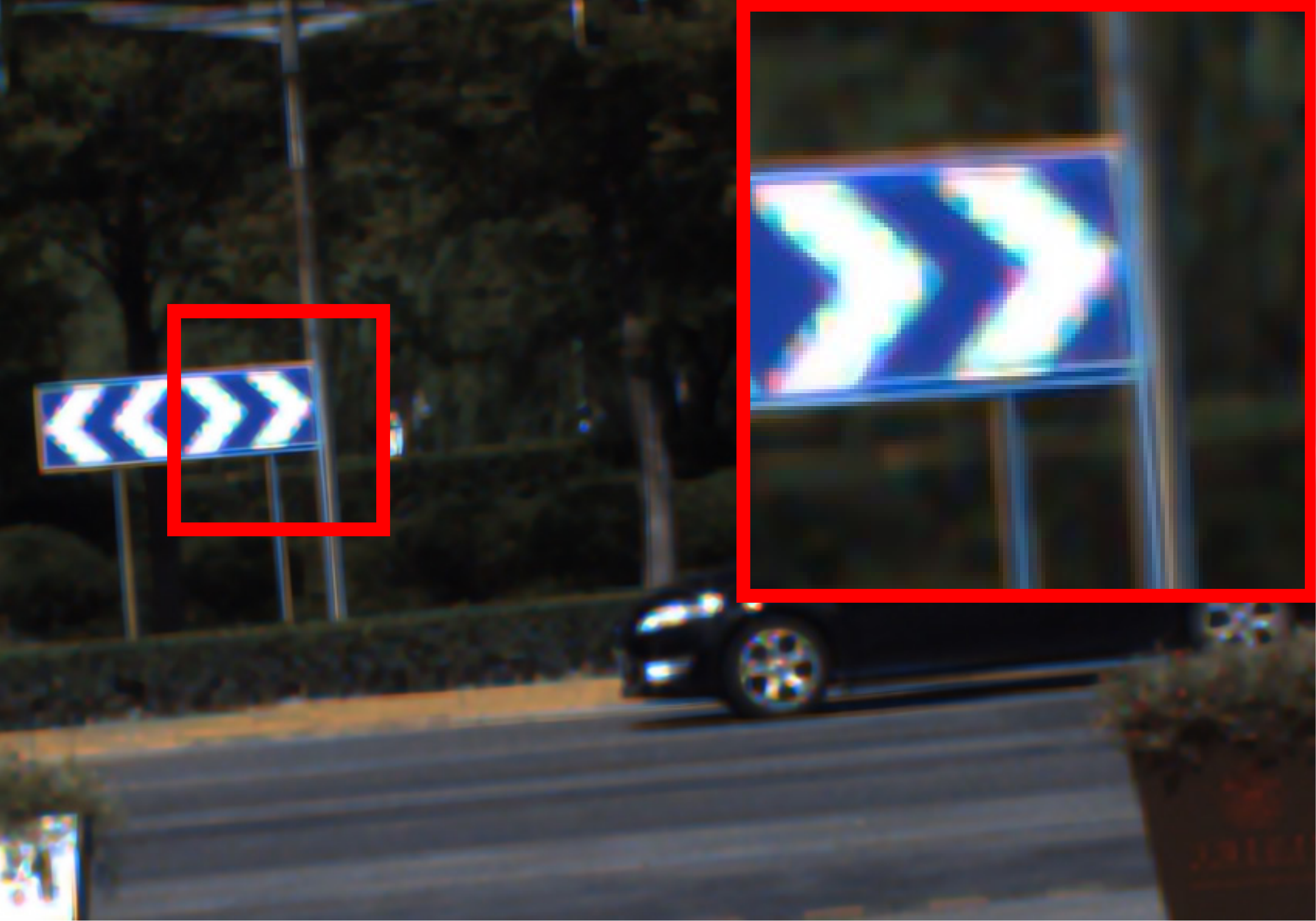} &
    \includegraphics[width=0.12\textwidth]{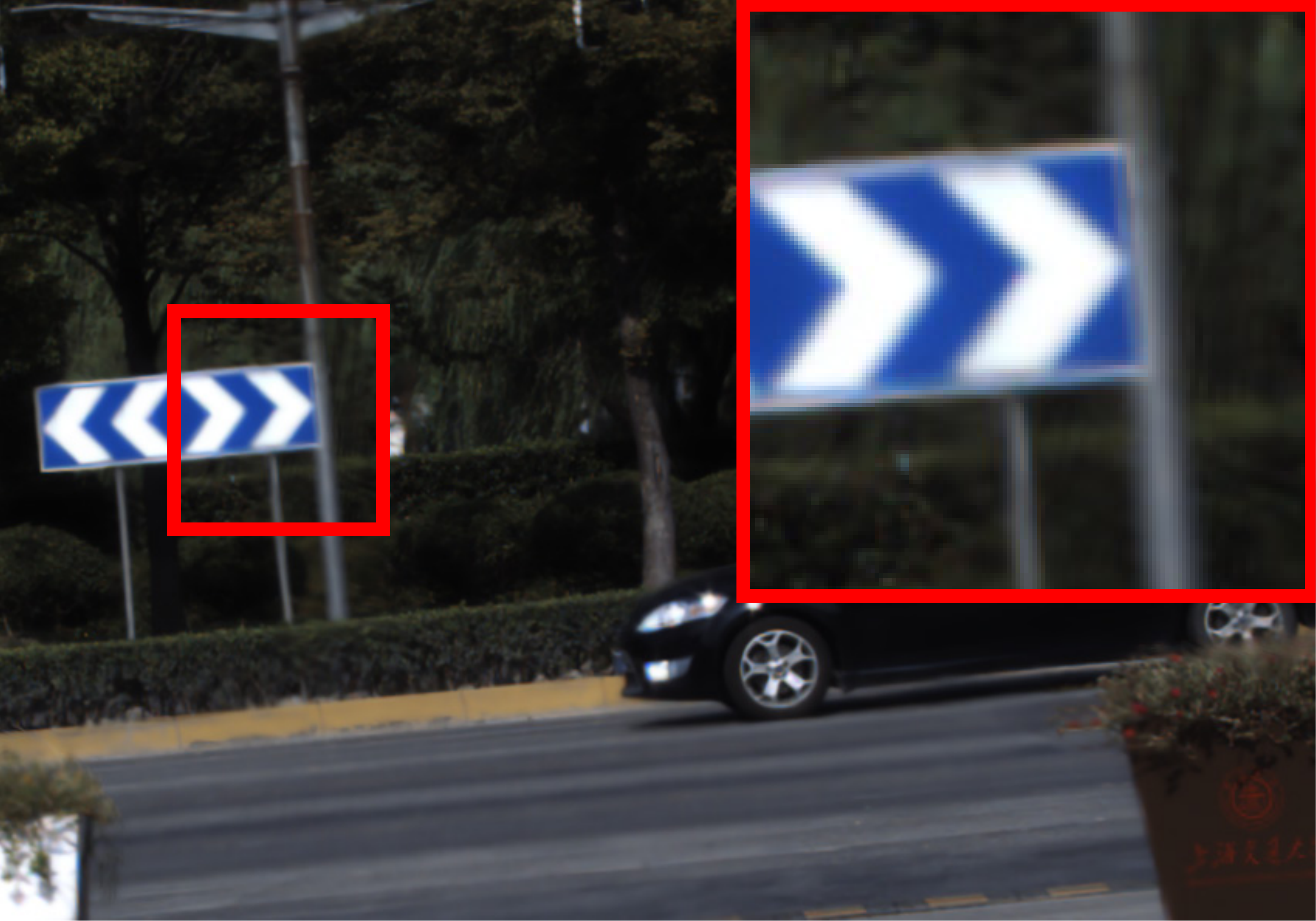} &
    \includegraphics[width=0.12\textwidth]{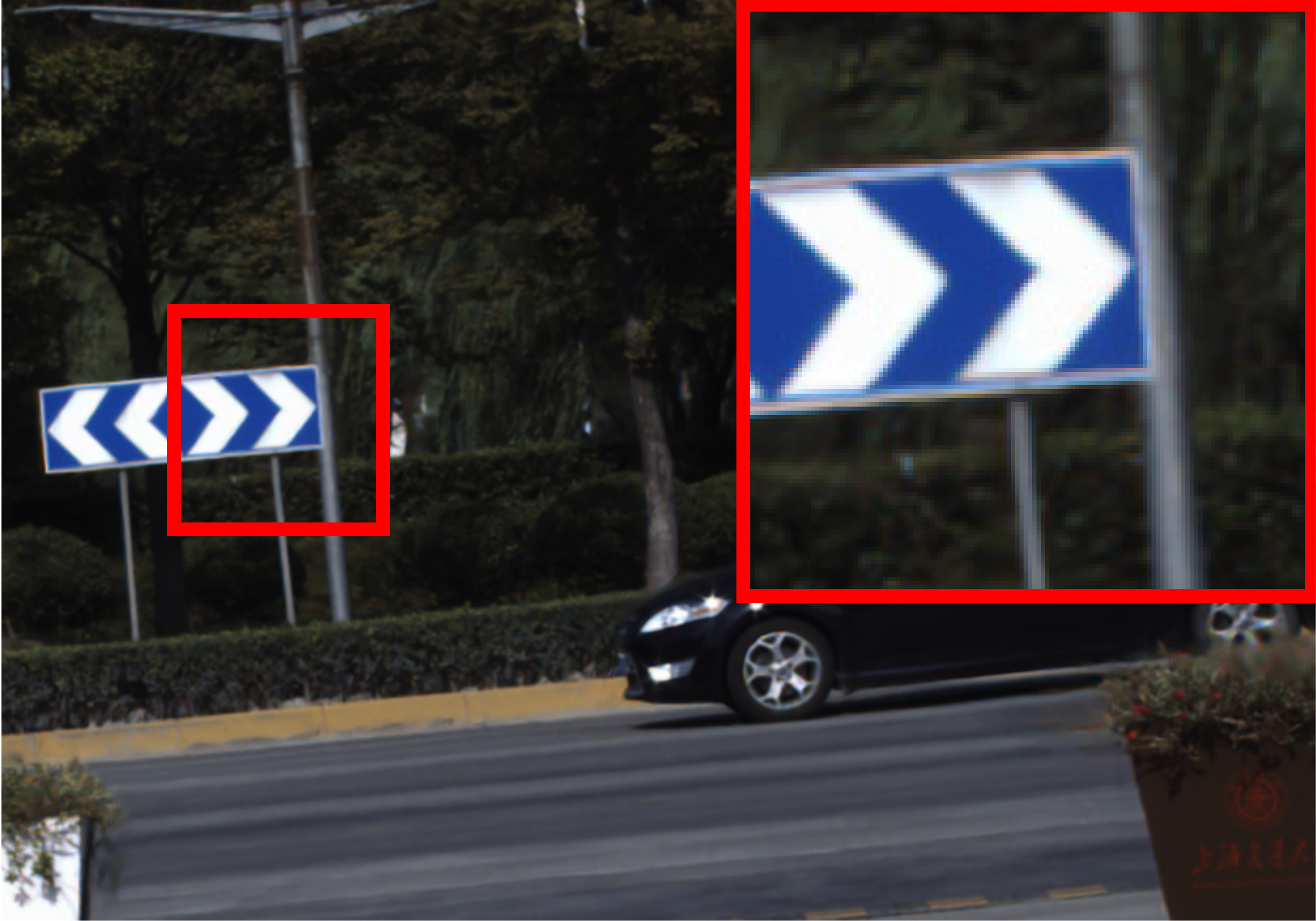} \\
    (a)LSR-HFR input & (b) EDVR~\cite{wang2019edvr} & (c) AWnet~\cite{cheng2020dual} & (d)Ours(LSR-HFR) \\
	\includegraphics[width=0.12\textwidth]{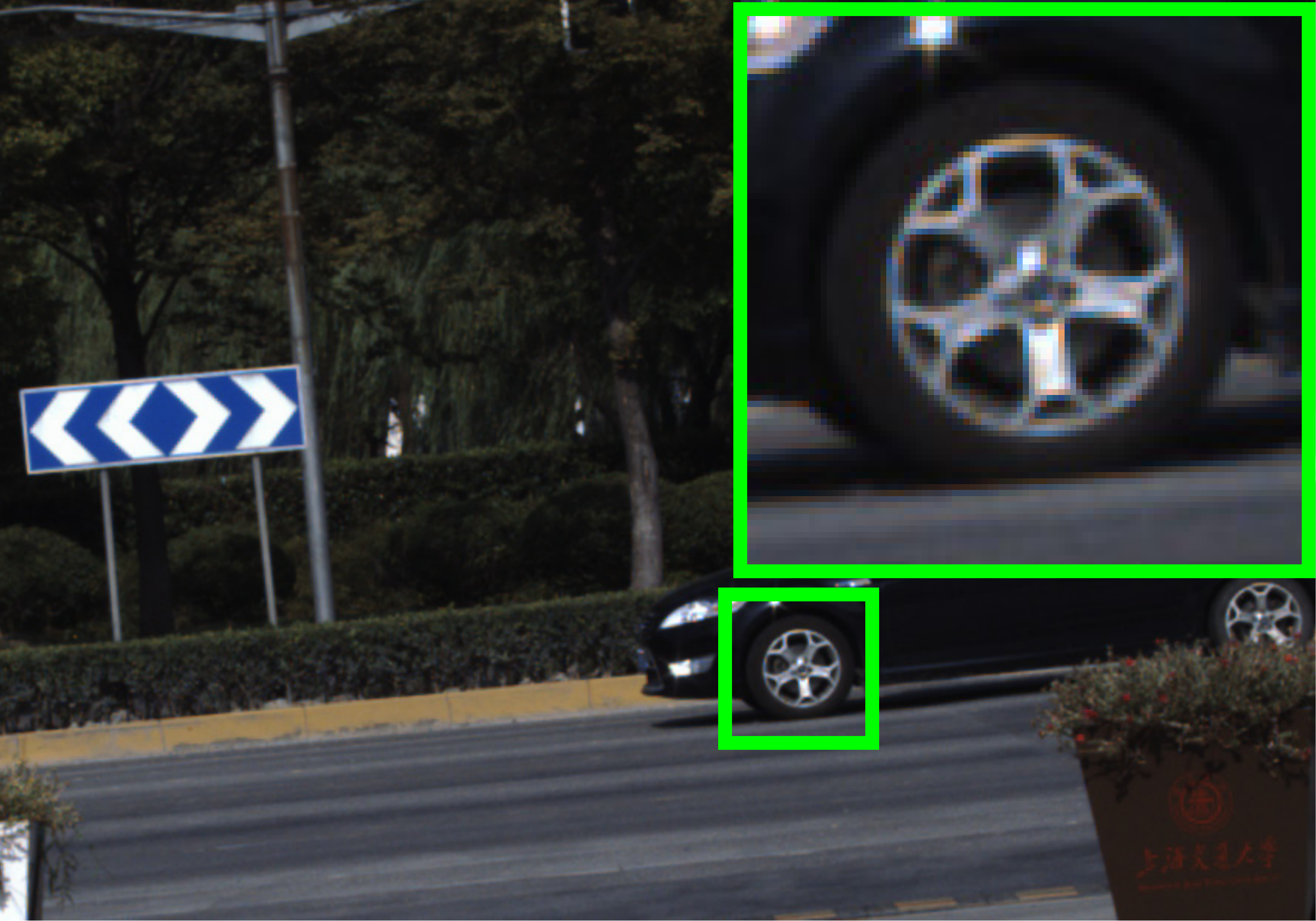} &
    \includegraphics[width=0.12\textwidth]{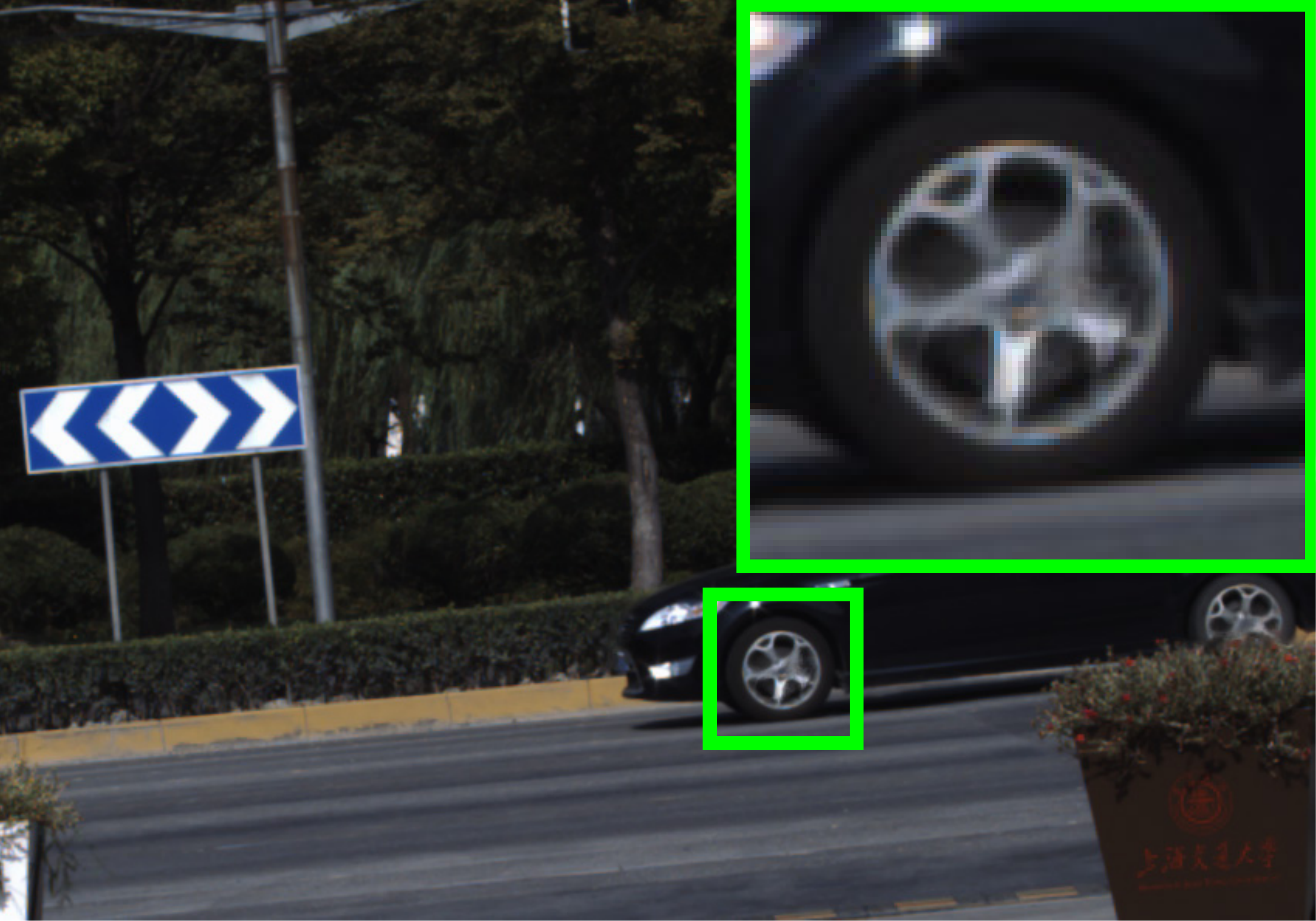} &
    \includegraphics[width=0.12\textwidth]{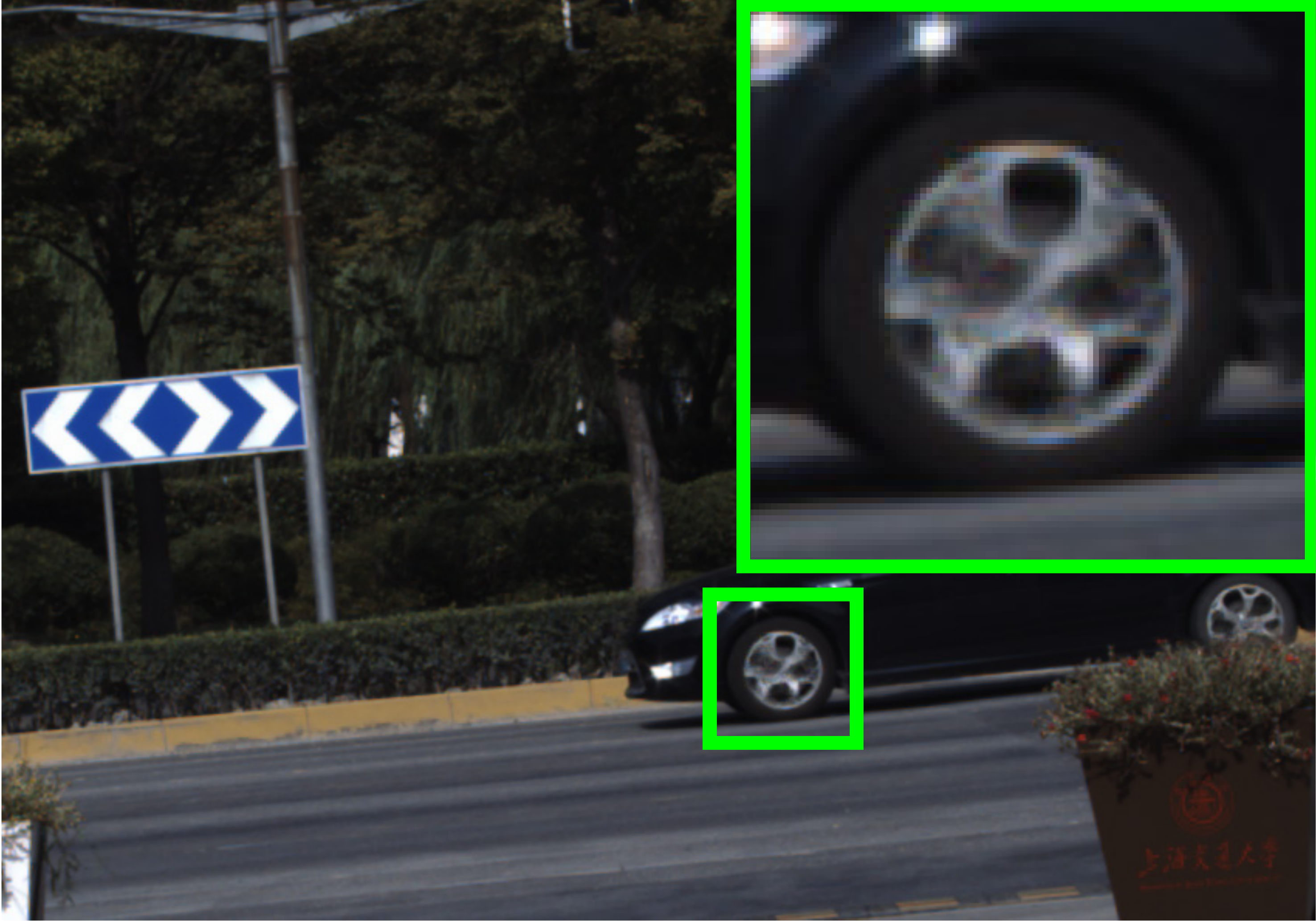} &
    \includegraphics[width=0.12\textwidth]{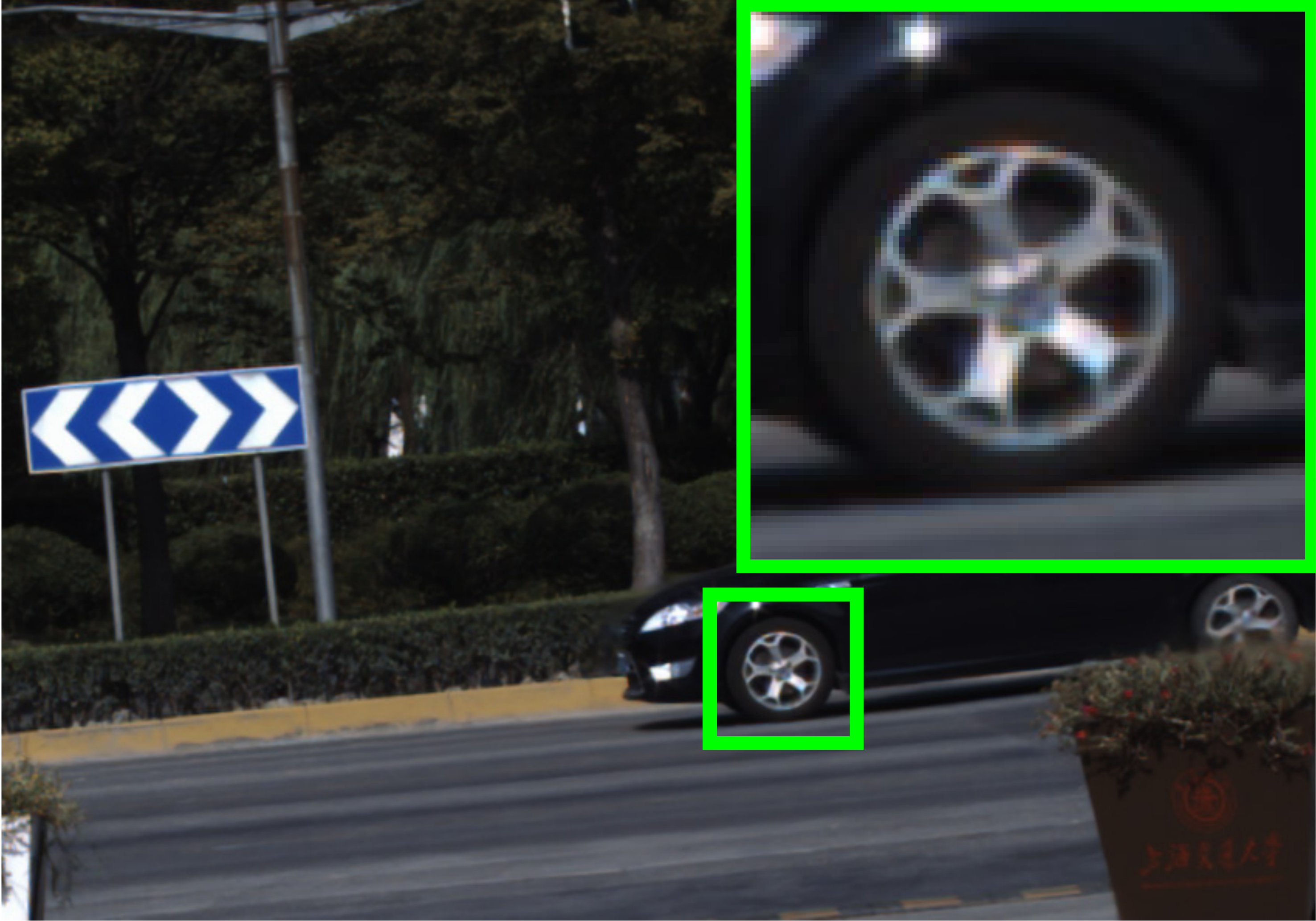} \\
    (e)HSR-LFR input & (f) BMBC~\cite{park2020bmbc} & (g) DAIN~\cite{bao2019depth} & (h)Ours(HSR-LFR) \\
	\includegraphics[width=0.12\textwidth]{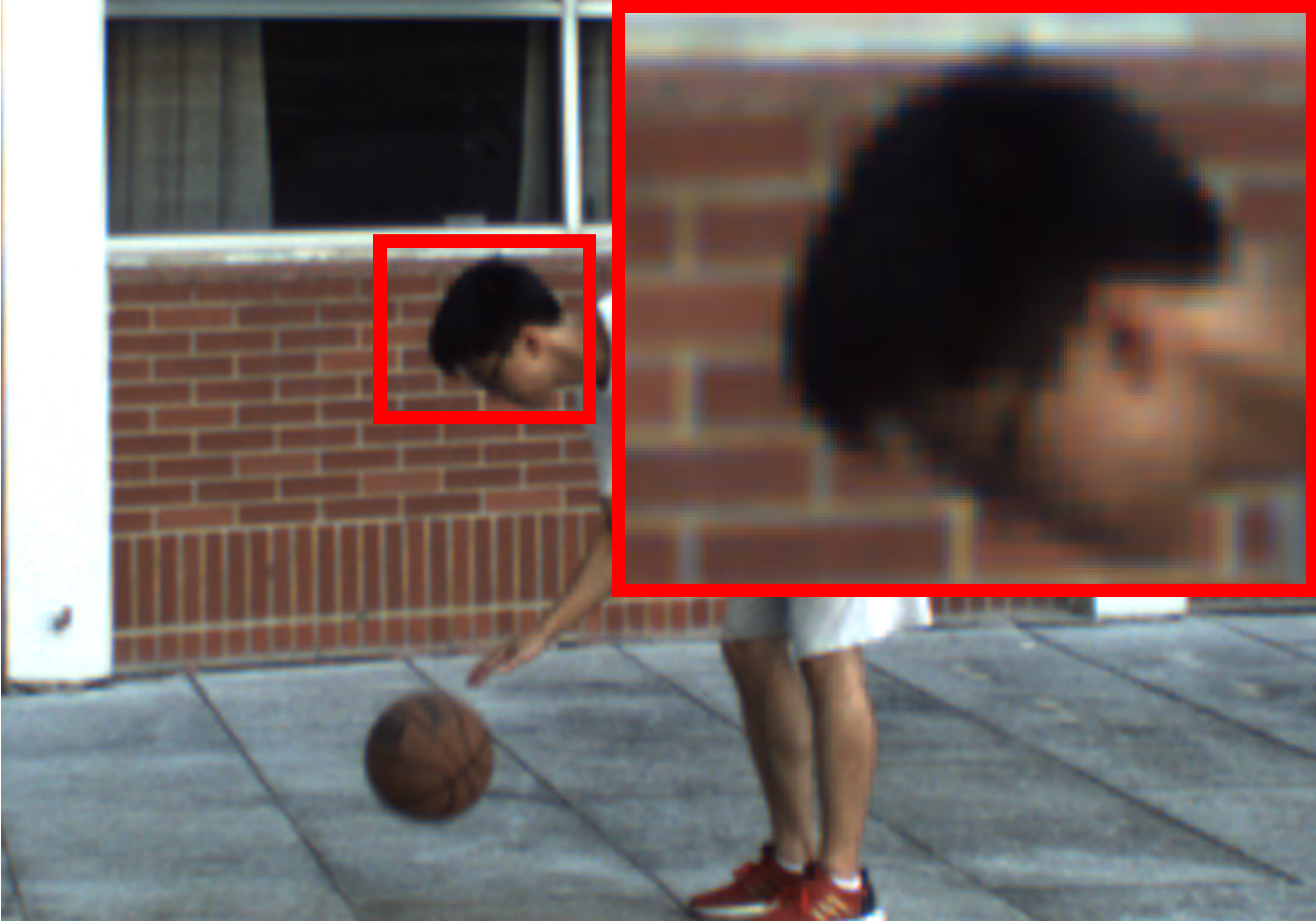} &
    \includegraphics[width=0.12\textwidth]{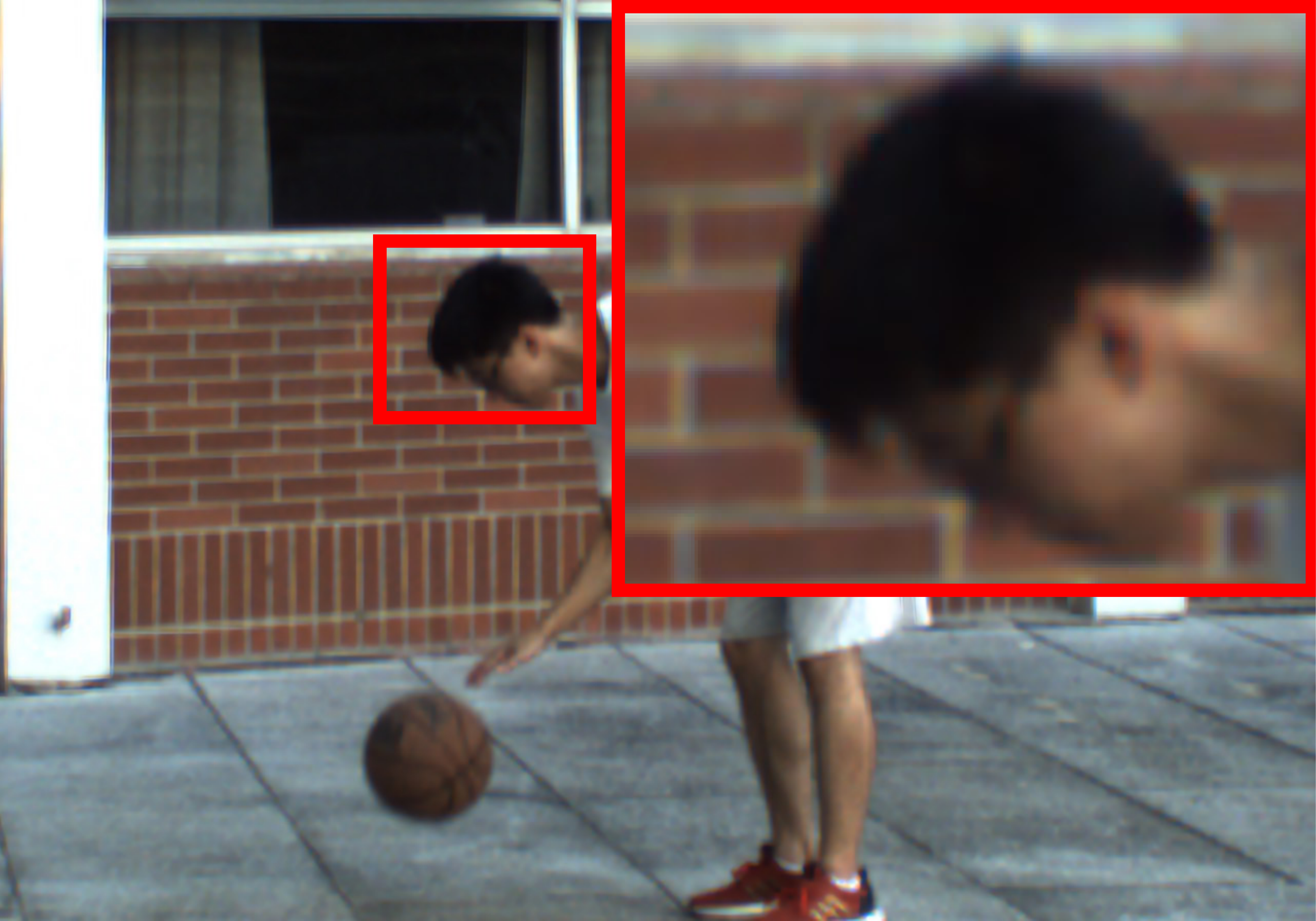} &
    \includegraphics[width=0.12\textwidth]{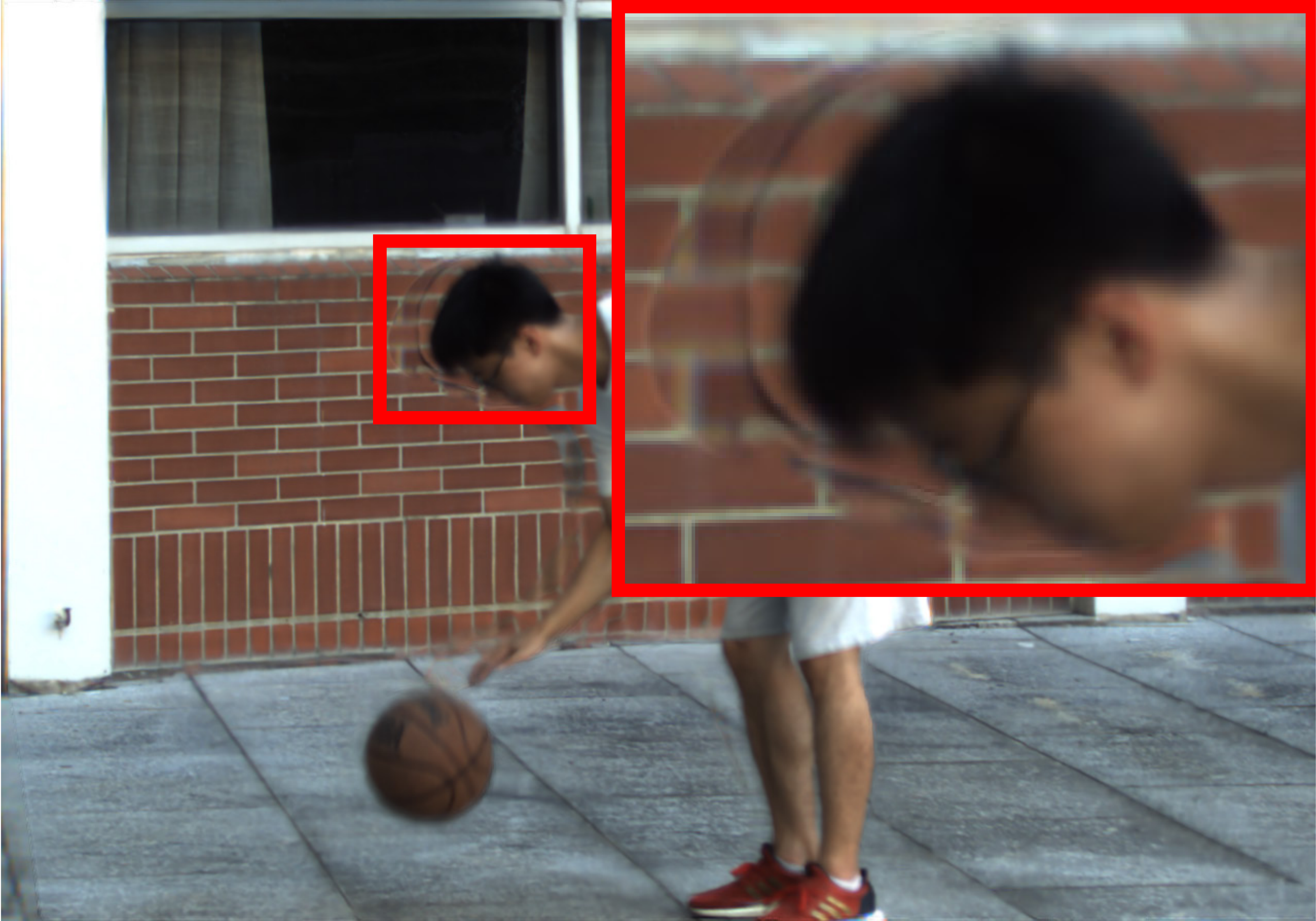} &
    \includegraphics[width=0.12\textwidth]{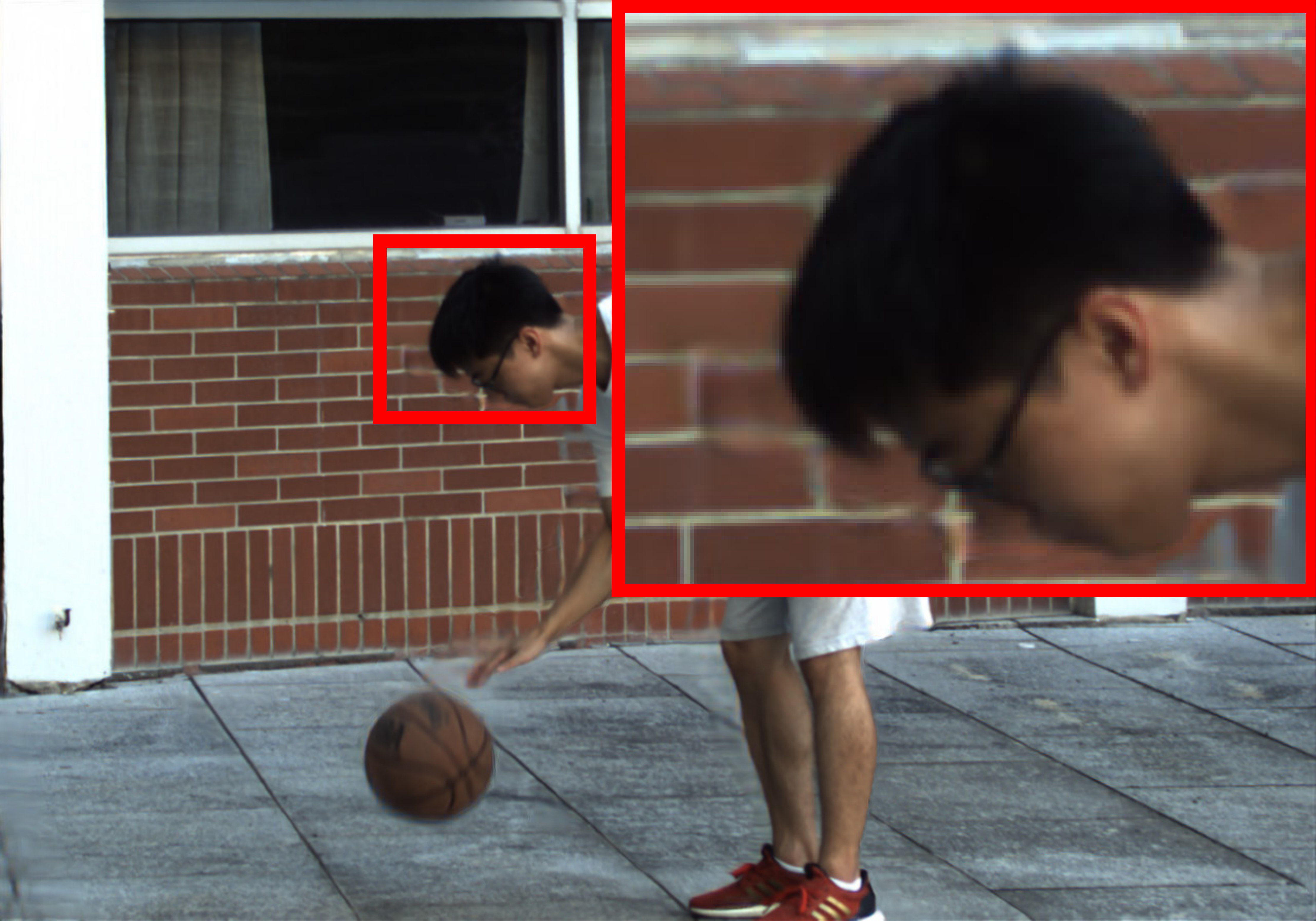} \\
	(a)LSR-HFR input & (b) EDVR~\cite{wang2019edvr} & (c) AWnet~\cite{cheng2020dual} & (d)Ours(LSR-HFR) \\
	\includegraphics[width=0.12\textwidth]{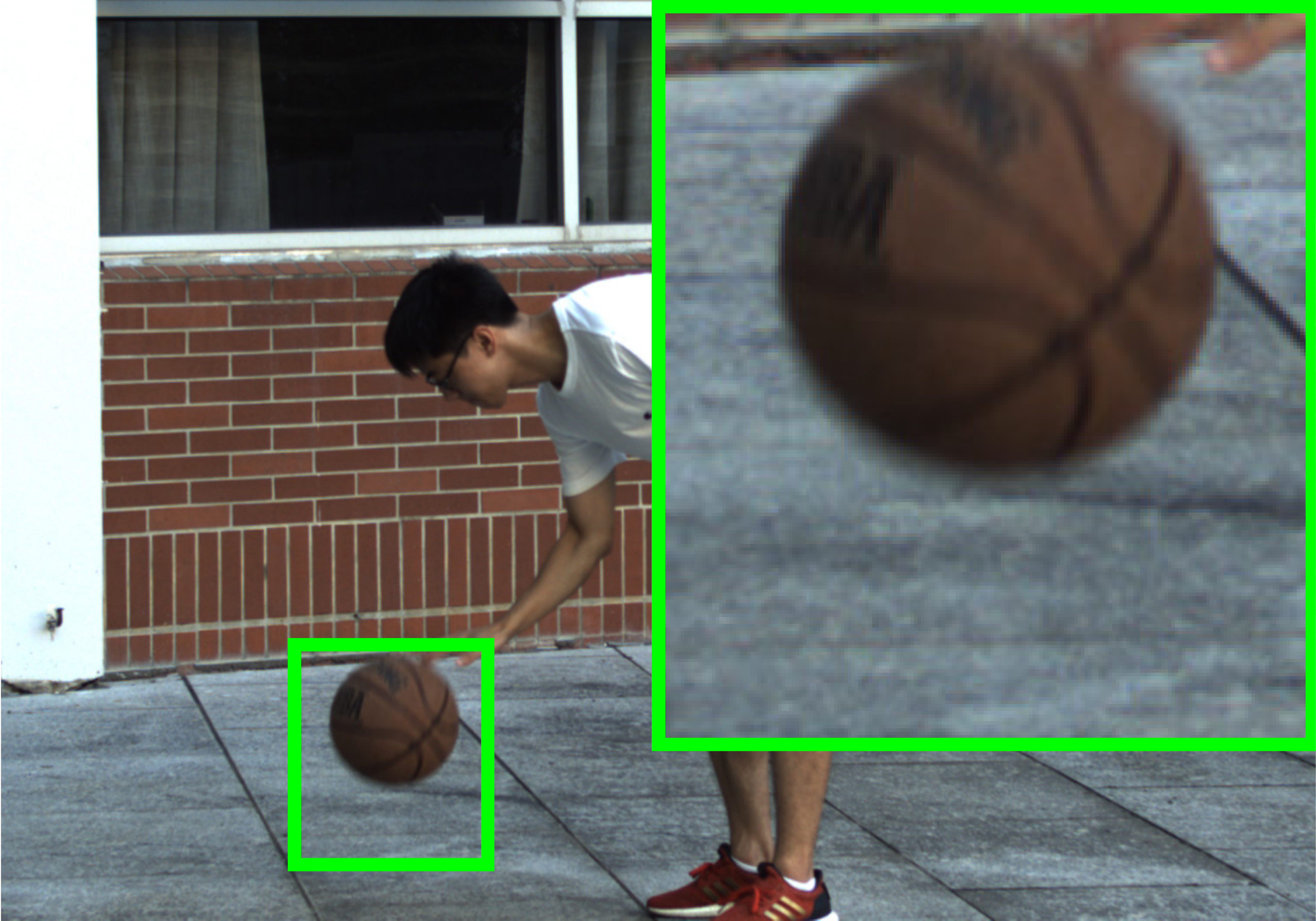} &
    \includegraphics[width=0.12\textwidth]{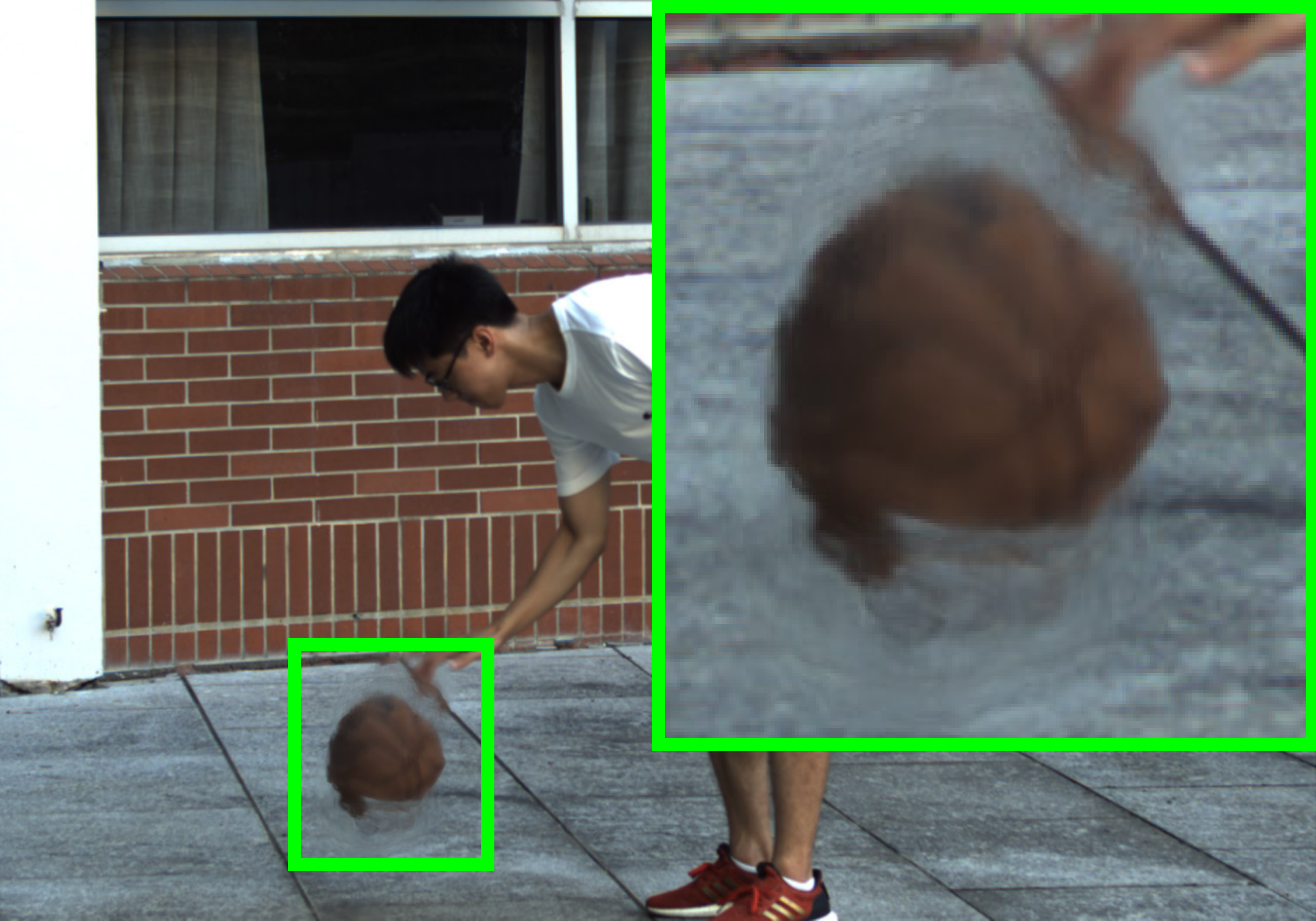} &
    \includegraphics[width=0.12\textwidth]{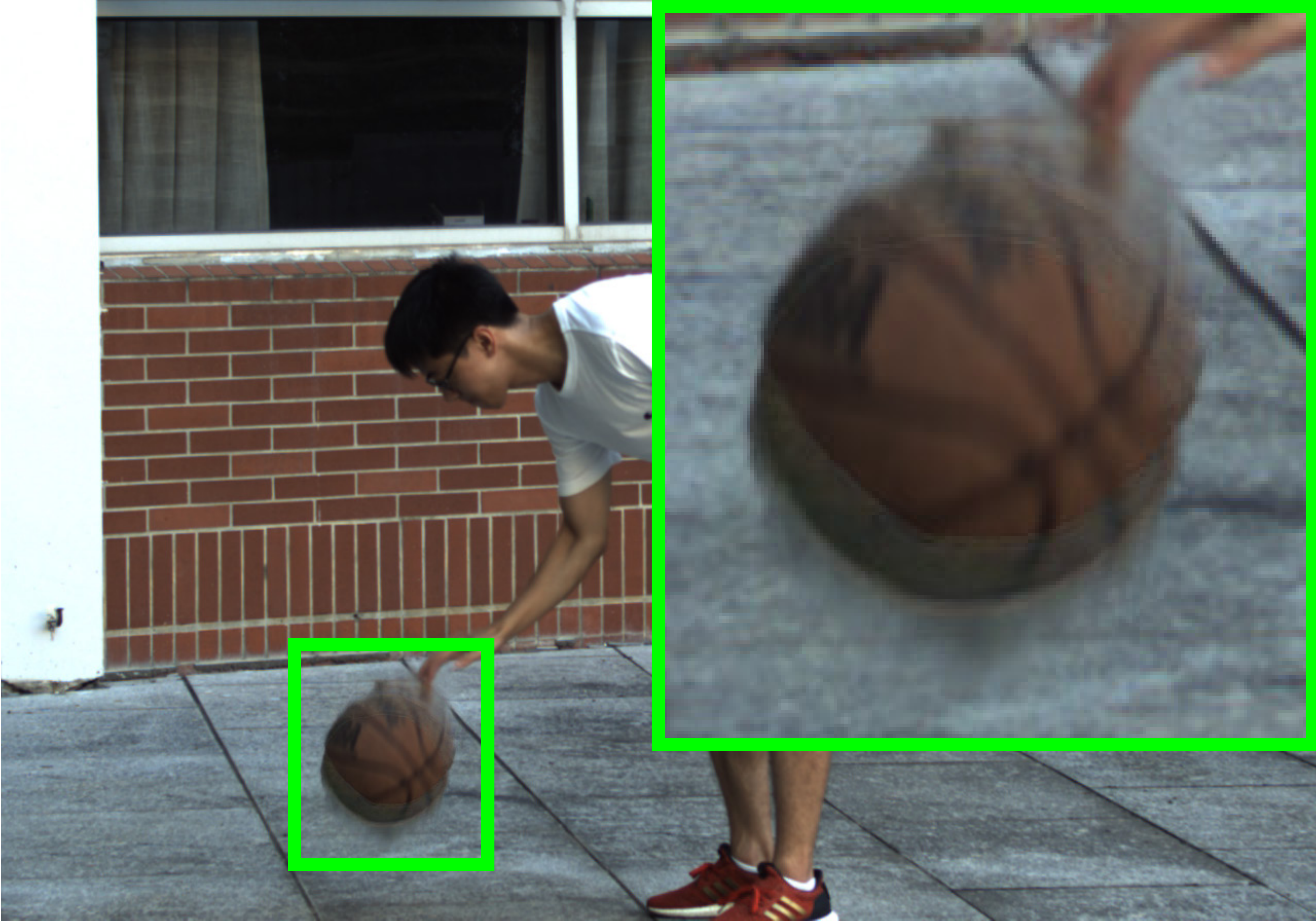} &
    \includegraphics[width=0.12\textwidth]{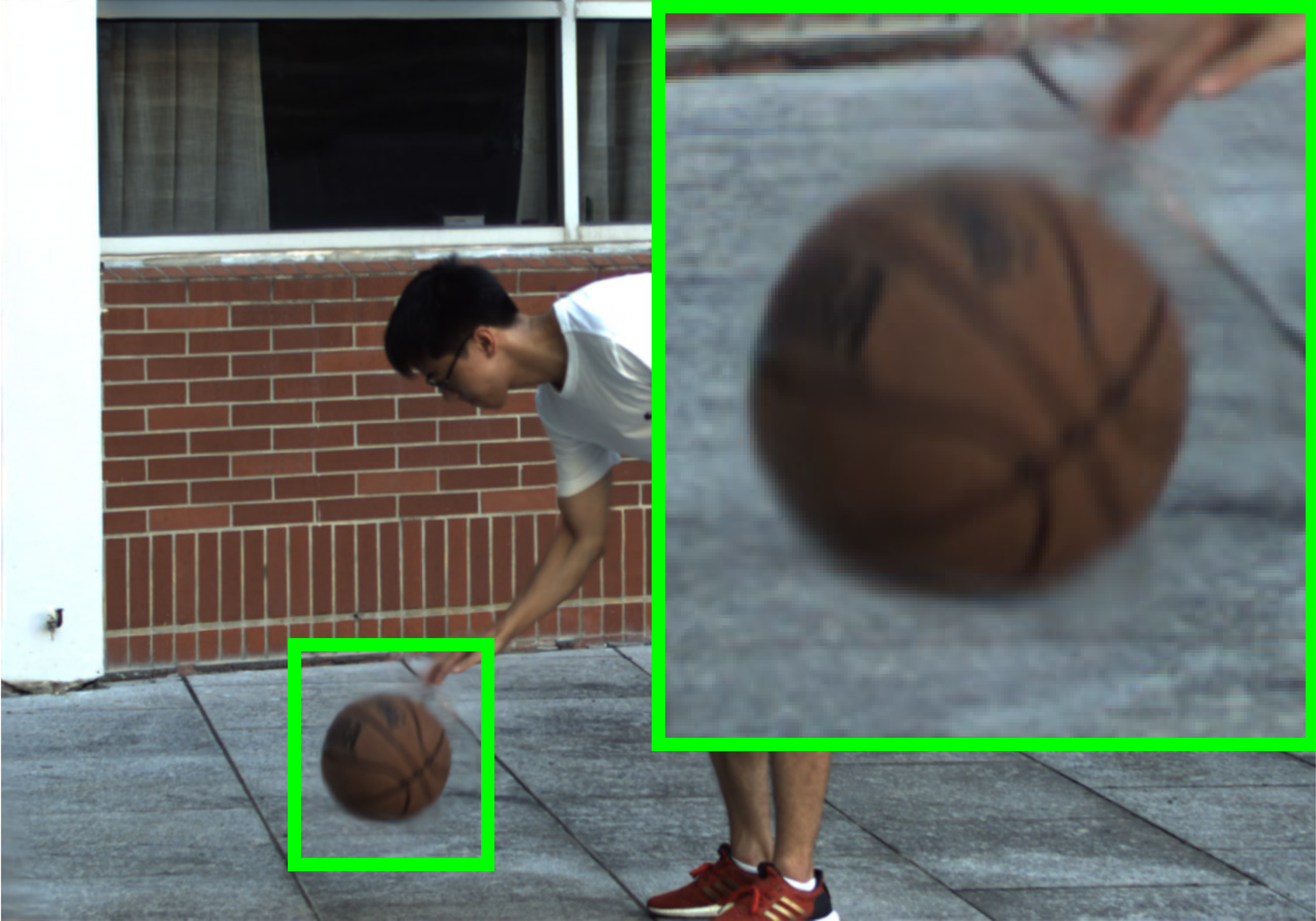} \\
	(e)HSR-LFR input & (f) BMBC~\cite{park2020bmbc} & (g) DAIN~\cite{bao2019depth} & (h)Ours(HSR-LFR) \\
	\end{tabular}
    % \vspace{-5pt}
	\caption{\textbf{Visual comparisons on real data.} 
	(a)-(d) show the images in LSR-HFR view. (e)-(h) are the images in HSR-LFR view. Our LIFnet reconstructs the fine-grained spatial textures and real temporal motions of both views.
    }
	\label{fig:real_data}
	\vspace{-15pt}
\end{figure}

\begin{figure}[t]
% \vspace{-20pt}
	\footnotesize
% 	\tiny 
% 	\scriptsize        
	\centering
	\renewcommand{\tabcolsep}{1.1pt} % adjust horizontal space
	\renewcommand{\arraystretch}{1} % adjust vertical space
      \newcommand{\quantTit}[1]{\multicolumn{3}{c}{\scriptsize #1}}
    \newcommand{\quantSec}[1]{\scriptsize #1}
    \newcommand{\quantInd}[1]{\scriptsize #1}
    \newcommand{\quantVal}[1]{\scalebox{0.83}[1.0]{$ #1 $}}
    \newcommand{\quantBes}[1]{\scalebox{0.83}[1.0]{$\uline{ #1 }$}}
	\begin{tabular}{cccc}
	\includegraphics[width=0.11\textwidth]{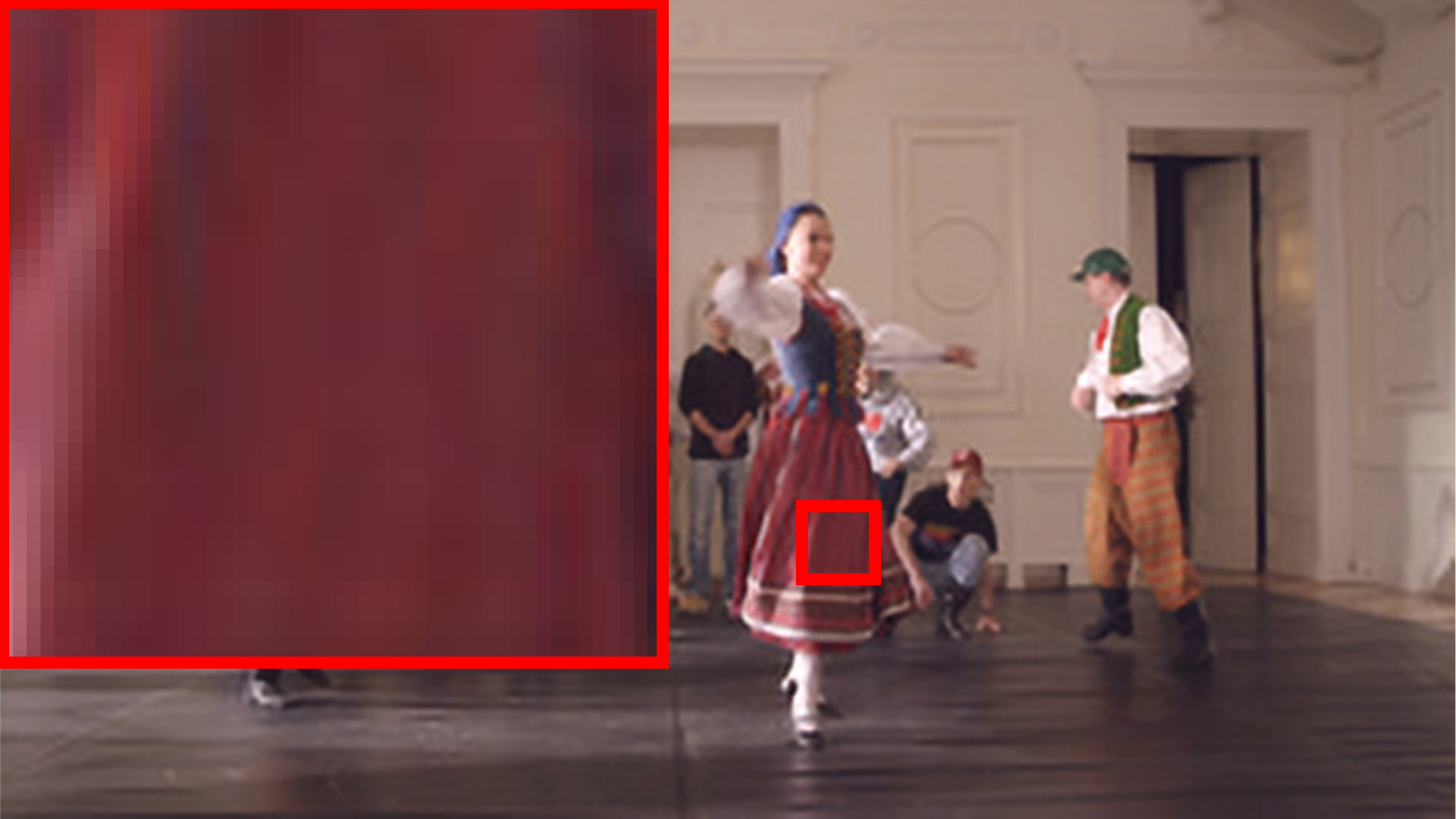} &
    \includegraphics[width=0.11\textwidth]{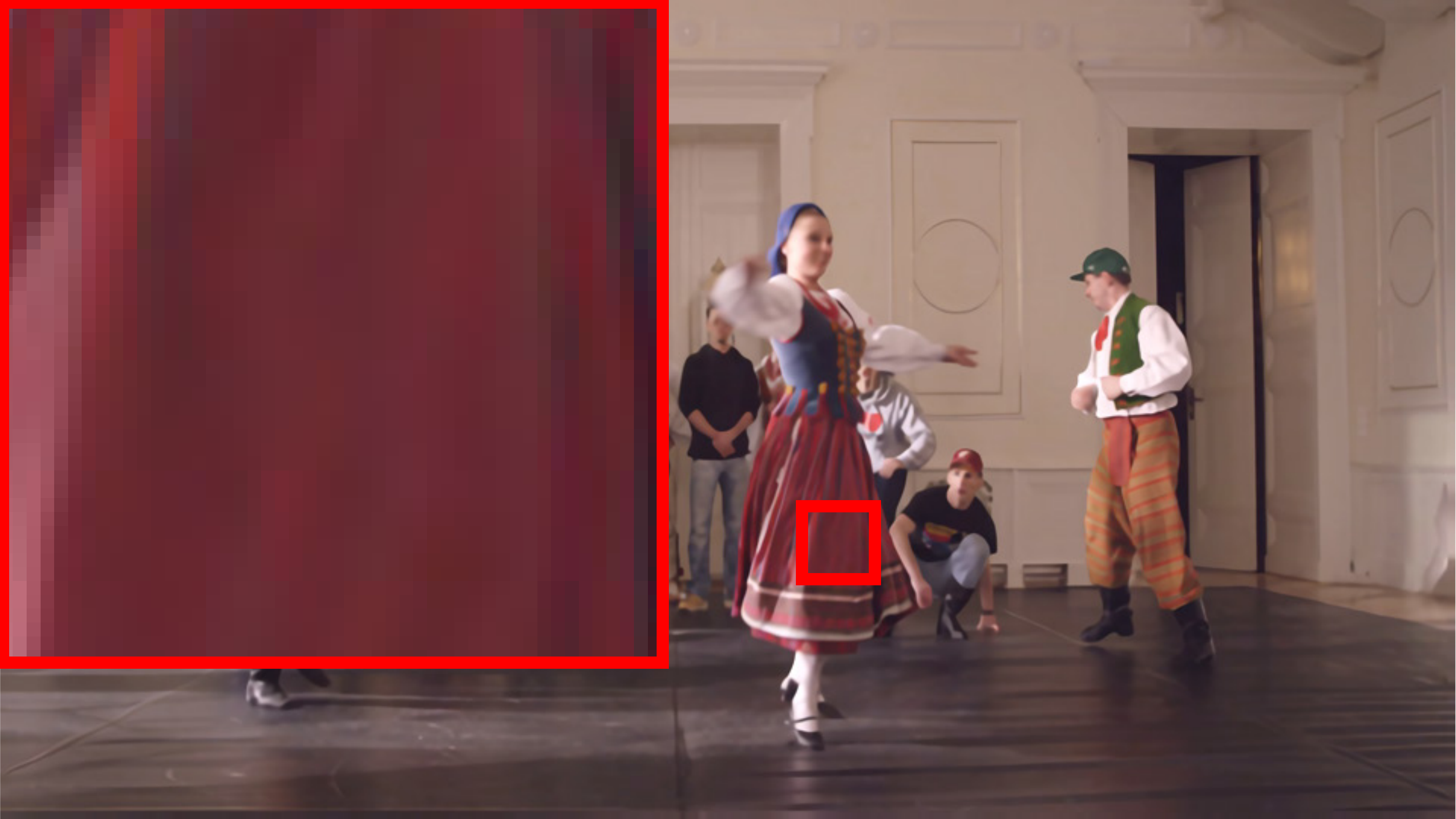} &
    \includegraphics[width=0.11\textwidth]{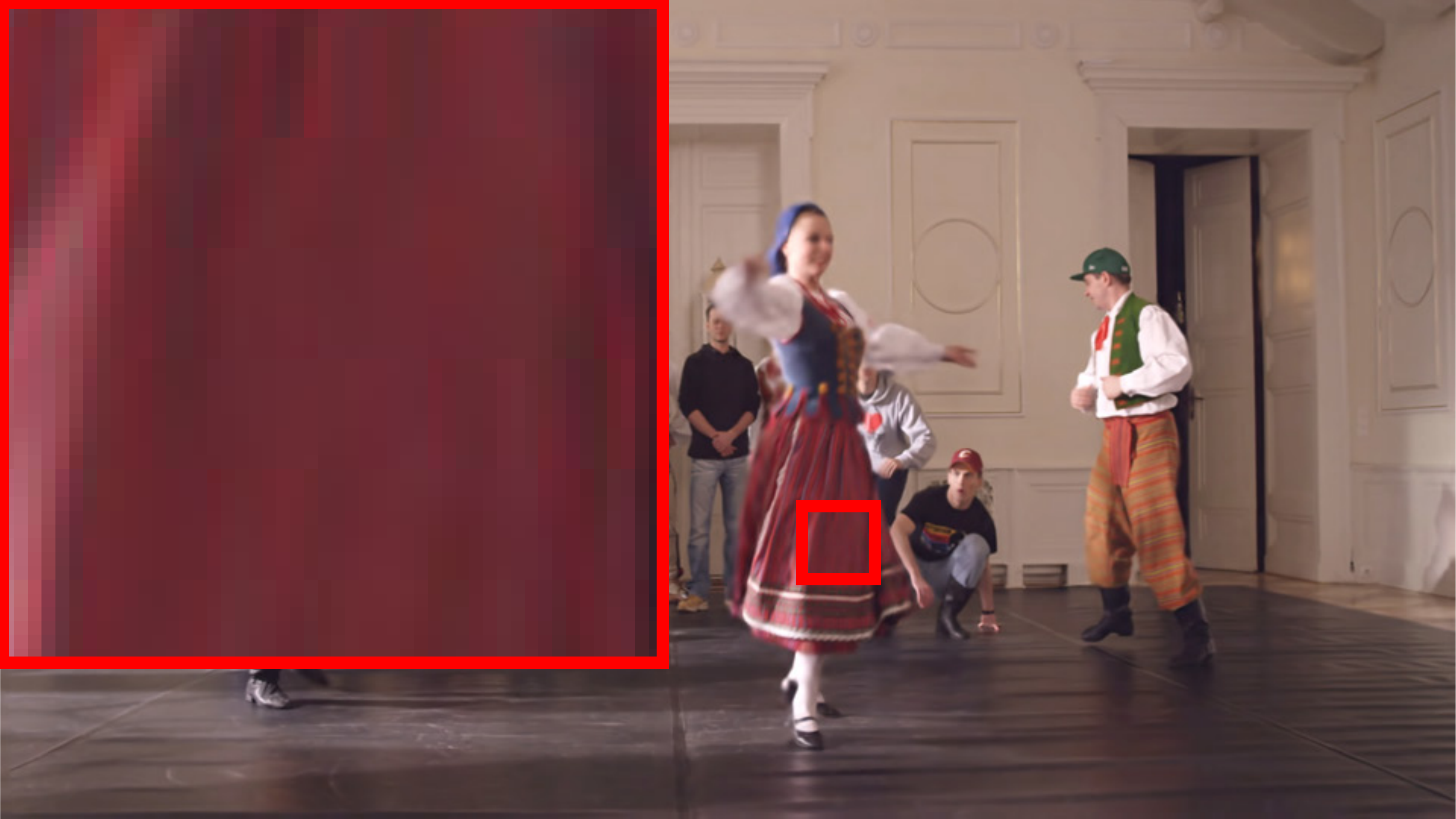} &
    \includegraphics[width=0.11\textwidth]{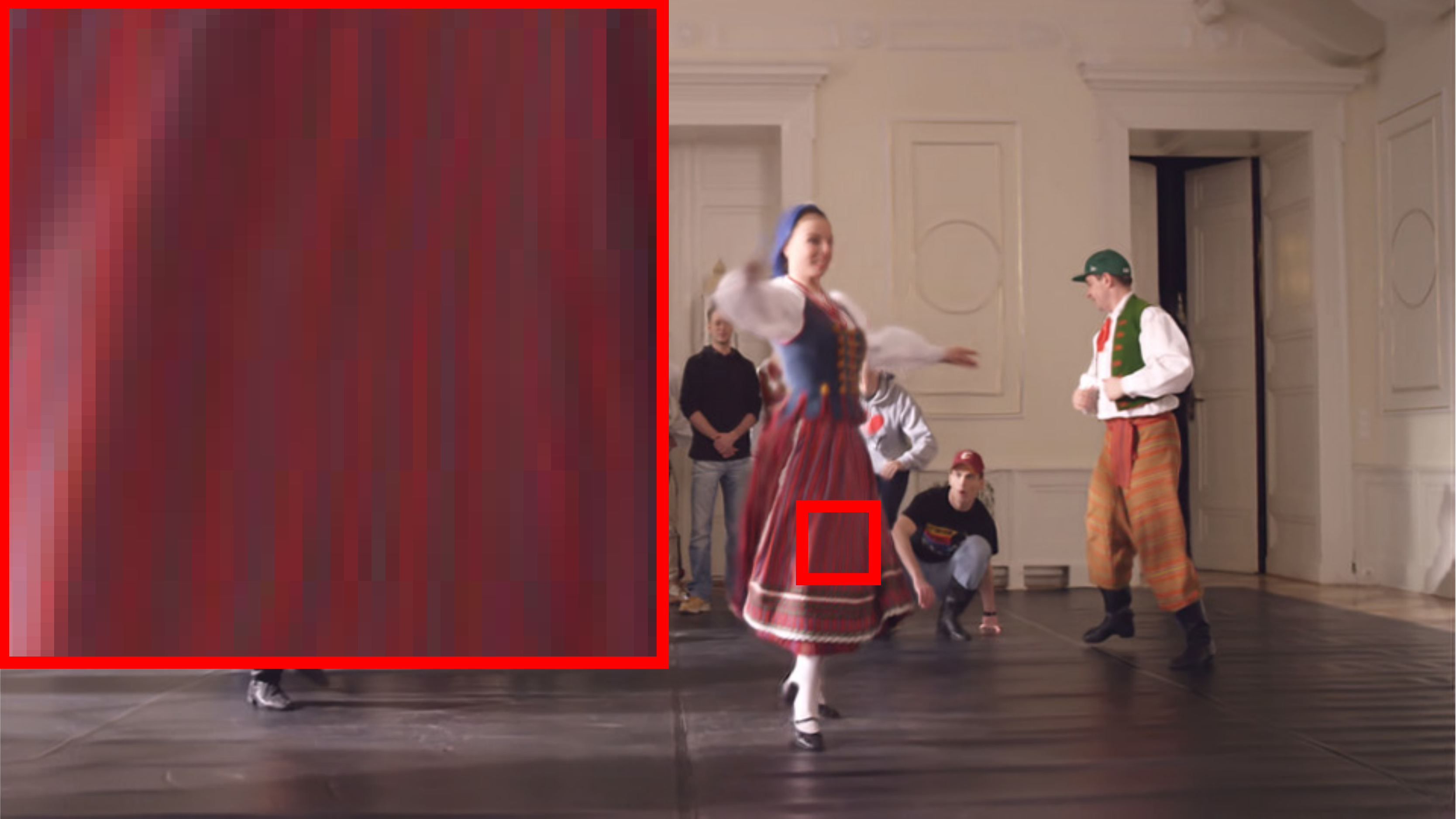}\\
    (a) PASSRnet~\cite{wang2019learning} & (b) EDVR~\cite{wang2019edvr} & (c) AWnet~\cite{cheng2020dual} & (d)Ours\scriptsize{(LSR-HFR)} \\
    \includegraphics[width=0.11\textwidth]{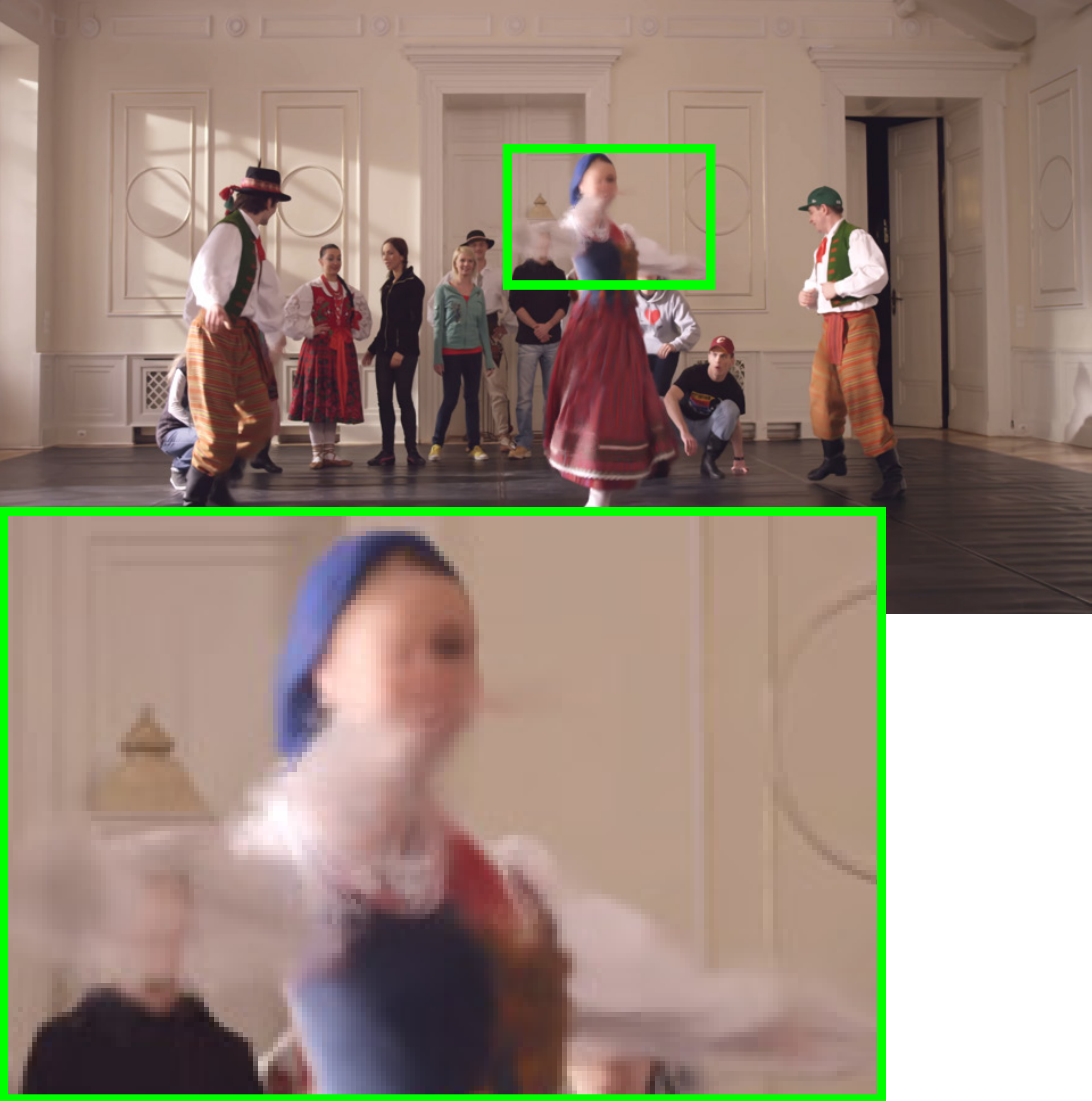} &
    \includegraphics[width=0.11\textwidth]{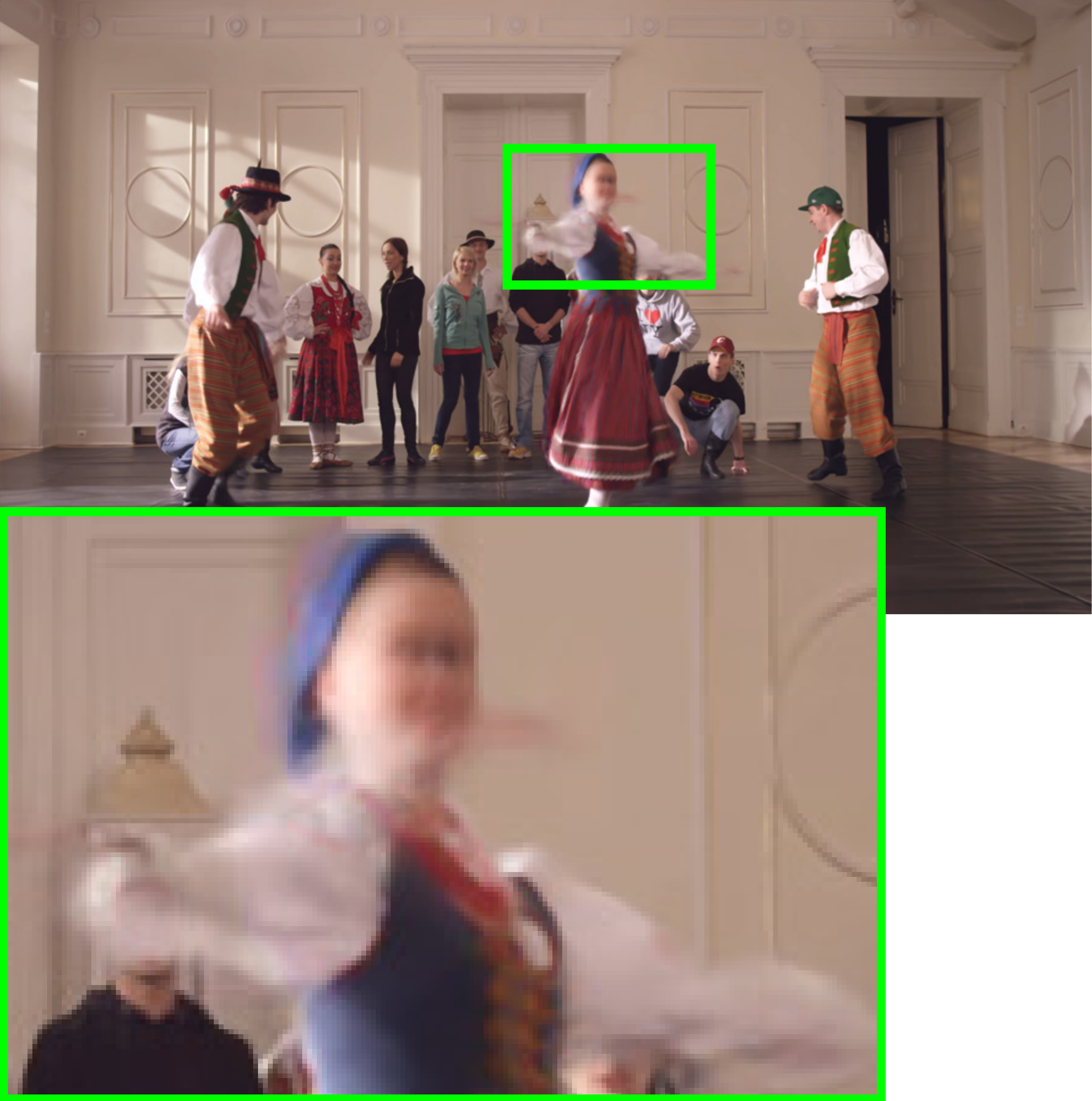} &
    \includegraphics[width=0.11\textwidth]{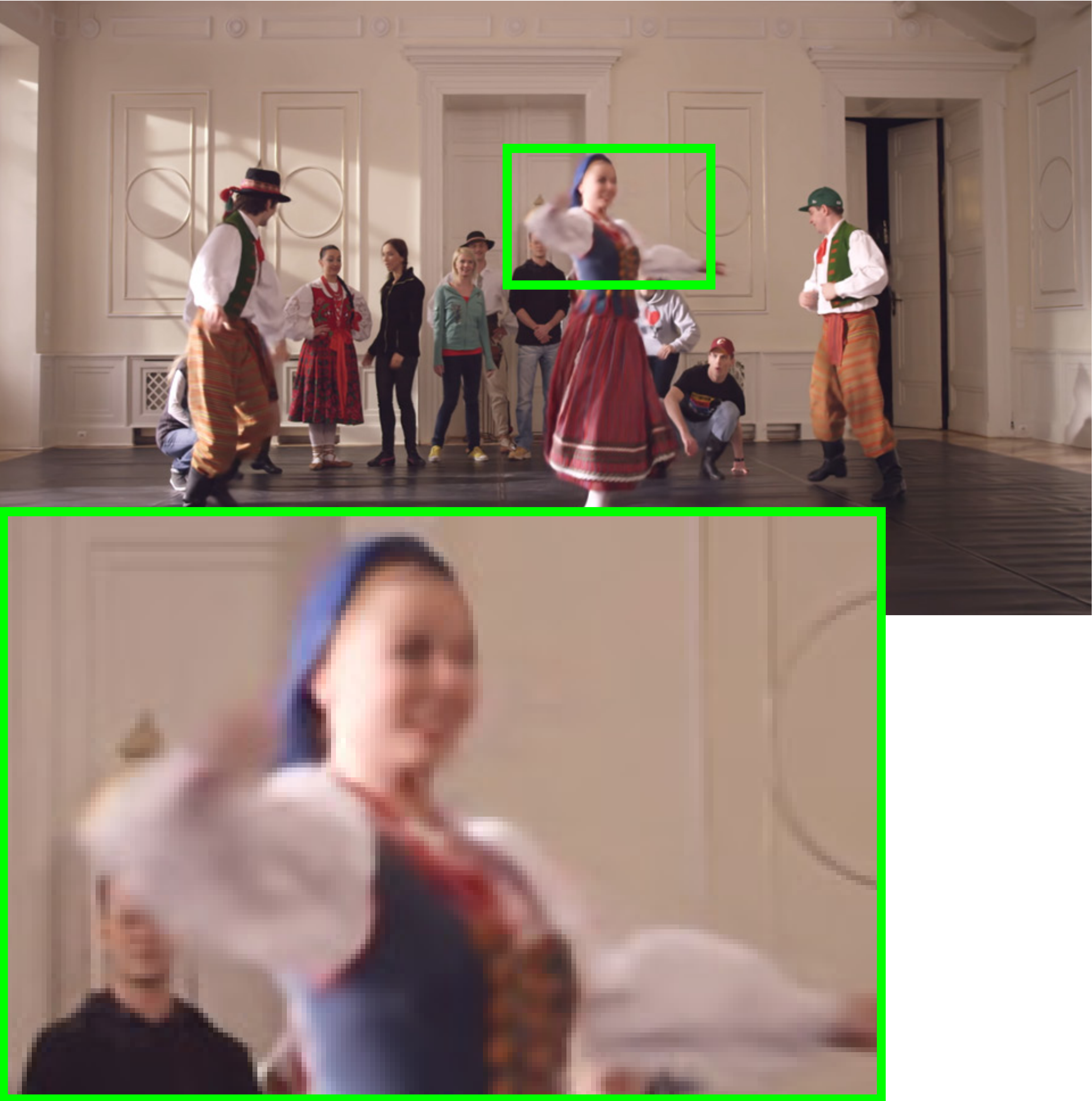} &
    \includegraphics[width=0.11\textwidth]{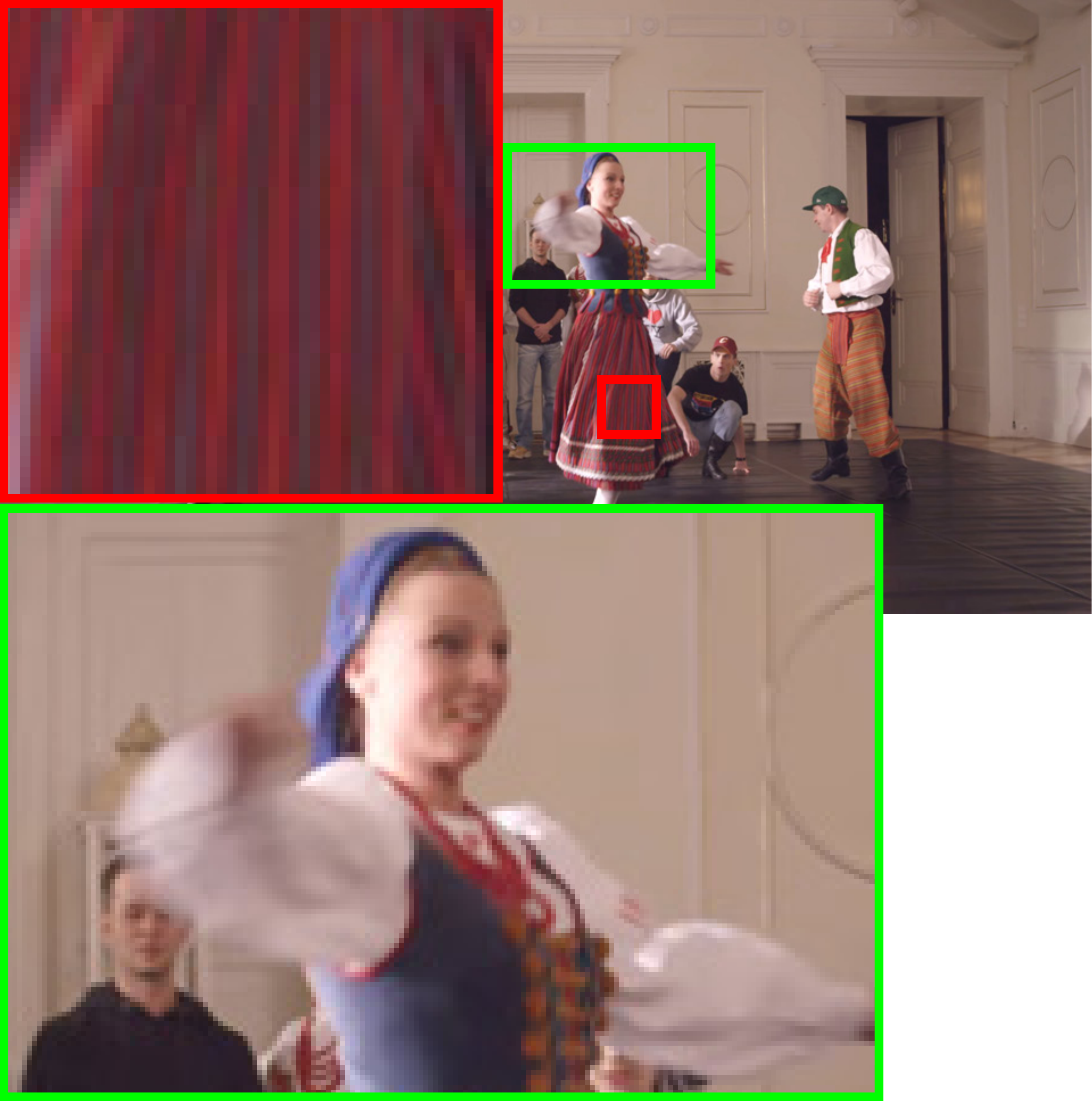}\\
    (e) BMBC~\cite{park2020bmbc} & (f) DAIN~\cite{bao2019depth} & (g) Ours\scriptsize{(HSR-LFR)} & (h) GT\scriptsize{(LSR-HFR)}\\
	\end{tabular}
    \vspace{-5pt}
	\caption{\textbf{Visual comparisons on synthetic data.} 
    We show the images of LSR-HFR view in (a)-(d) and HSR-LFR view in (e)-(g).
    (h) is the ground truth image in LSR-HFR view. Our LIFnet reconstructs fine-grained spatial textures and smooth temporal motions of both views.
	}
	\label{fig:syn_data}
	\vspace{-10pt}
\end{figure}

    \paragraph{Subjective results}
    We compare our LIFnet with other methods subjectively with synthetic data and real data.
    Figure~\ref{fig:syn_data} shows the subjective comparisons on synthetic data.
    In LSR-HFR view, our LIFnet reconstructs much sharper spatial textures comparing with PASSRnet~\cite{wang2019learning}, EDVR~\cite{wang2019edvr} and AWnet~\cite{cheng2020dual}, see the dress.
    In HSR-LFR view, existing frame interpolation methods fail to reconstruct the smooth motion of the dancer. However, our LIFnet can restore smooth and accurate motions.
    As shown in Figure~\ref{fig:real_data}, LIFnet outperforms all the spatiotemporal enhancement methods on real data.
    In LSR-HFR view, existing super-resolution algorithms can not obtain good results due to the spatial information losses and the complex degradation of LSR-HFR frames in real data.
    {Due to the large disparities, flow-based AWnet~\cite{cheng2020dual} fails to align the high-frequency details of the road sign and the athlete's head from HSR-LFR view to LSR-HFR view, thus, resulting in low-definition reconstruction.
    In contrast, our disparity-guided flow-based method can well align the road sign and the athlete's head, resulting in fine-grained spatial textures.
    In addition, there exist warping ghosts due to view occlusions around the athlete's head in the result of AWnet~\cite{cheng2020dual}, as shown in Figure~\ref{fig:real_data}(c). However, our method completely removes the warping ghosts because we use the disparity map to assist the fusion network in detecting view occlusions.
    }
    In HSR-LFR view, the temporal motions of the fast-moving objects are missed by frame interpolation methods, while our method preserves good shapes of the moving objects, see the tyre and basketball.
    {Our method transfers the accurate temporal and appearance information of the fast-moving objects from LSR-HFR view to HSR-LFR view based on complementary warping, which contributes to the frame interpolation process.
    }

% \renewcommand\thetable{\Alph{table}}
% \vspace{-10pt}
\begin{table}[t]
% \vspace{-10pt}
\caption{\textbf{Running times on image of size $640\times 512$.}}
\label{table:com}
\centering
\footnotesize
\newcommand{\quantVal}[1]{\scalebox{1.0}[1.0]{#1}}
\newcommand{\quantInd}[1]{\scalebox{1.0}[1.0]{\scriptsize #1}}
\renewcommand{\tabcolsep}{1.8pt} % adjust horizontal space
\renewcommand{\arraystretch}{0.9} % adjust vertical space
\vspace{-5pt}
% \specialrule{0em}{1pt}{1pt}
\begin{tabular}{cccccc}
\toprule
\multirow{2}{*}{Methods} & Two Views & \multicolumn{2}{c}{LSR-HFR} & \multicolumn{2}{c}{HSR-LFR} \\
\cmidrule{2-6}
& LIFnet (ours) & \quantVal{AWnet~\cite{cheng2020dual}} & \quantVal{EDVR~\cite{wang2019edvr}} & \quantVal{DAIN~\cite{bao2019depth}} & \quantVal{BMBC~\cite{park2020bmbc}} \\
\midrule

\scriptsize{running time (s)} & 0.58 & 0.19 & 0.20 & 0.69 & 4.03 \\
\bottomrule
\end{tabular}
\vspace{-15pt}
\end{table}

    \paragraph{Model efficiency}
    We compare the model efficiencies using an RTX-1080 Ti GPU card on the input image size of $640\times 512$.
    We compare the running time of LIFnet with the sum of the running times of the two views of existing methods for a fair comparison.
    As shown in Table~\ref{table:com}, we achieve $2.99$ dB performance gain on average (Stereo Video dataset in Table~\ref{table:stereoresults}) and reduces $34\%$ running time compared with AWnet~\cite{cheng2020dual} + DAIN~\cite{bao2019depth}.

\section{Analyses}\label{sec:ana}

    In this section, we perform ablation studies to understand the system capability and model robustness.
    
\subsection{Weights for loss functions.}

\begin{table}[t]
% \vspace{-20pt}
\vspace{-5pt}
\caption{\textbf{Analysis of the weights for the reconstruction losses.}}
\vspace{-5pt}
\label{table:weighted_loss}
\centering
\footnotesize
\renewcommand{\tabcolsep}{5pt} % adjust horizontal space
\renewcommand{\arraystretch}{0.9}
% \vspace{-5pt}

\begin{tabular}{ l c c c c}
\toprule
\multirow{2}{*}{Weights} & \multicolumn{2}{c}{Stereo Video} & \multicolumn{2}{c}{KITTI} \\
& LSR-HFR & HSR-LFR & LSR-HFR & HSR-LFR \\
\midrule
$\lambda_L=10,\lambda_R=1$ & 40.48 & 39.49 & 26.43 & 24.10\\
$\lambda_L=1, \lambda_R=10$ & 40.02 & 39.13 & 26.33 & 23.87 \\
$\lambda_L=1, \lambda_R=1$ & 40.22 & 39.19 & 26.40 & 24.07 \\
\bottomrule
\end{tabular}
\vspace{-10pt}
\end{table} 

  	Our LIFnet is a multi-task learning network.
  	Thus, the weights of the reconstruction losses of the two views are critical for the performance of the shared disparity and flow networks.
  	To analyze the optimal loss weight $\lambda_L$ for LSR-HFR view and $\lambda_R$ for HSR-LFR view, we retrain our network with different weights of losses with , as shown in Table~\ref{table:weighted_loss}. 
  	Our performances for LSR-HFR and HSR-LFR views improve by 0.26dB and 0.3dB, respectively, by adjusting the loss weights from $\{\lambda_L=1,\lambda_R=1\}$ to $\{\lambda_L=10,\lambda_R=1\}$. 
  	The results on KITTI~\cite{Menze2015CVPR} also prove the advantage of the adjusted loss weights.
  	That means we should pay more attention to the backward gradient from the LSR-HFR view to fine-tune the disparity and flow networks.
    We warp disparities based on flows and warp flows based on disparities, then mix the warped disparities and flows for complementary image warping, which is hard to train without sufficient supervisors.
    In contrast, we directly concatenate the estimated disparities and flows among the input images to reconstruct the LSR-HFR view, which is simpler and more accurate than the HSR-LFR view.
    Thus, paying more attention to LSR-HFR view can result in more accurate disparities and flows.
    In following ablation studies, we train the models with $\{\lambda_L=1,\lambda_R=1\}$ for comparison.
    
    \begin{table}[t]
% \vspace{-5pt}
\caption{\textbf{Effect of disparity guidance for lsr-hfr view in psnr $\textup{[dB]}$.}}
\label{table:disp_gui}
\centering
\footnotesize
\newcommand{\quantInd}[1]{\scalebox{0.8}[1.0]{\scriptsize #1}}
\renewcommand{\tabcolsep}{2pt} % adjust horizontal space
\renewcommand{\arraystretch}{0.9} % adjust vertical space
\vspace{-5pt}
\begin{tabular}{lcccccccc}
\toprule
Resolution & \multicolumn{4}{c}{$960\times 540$} & \multicolumn{4}{c}{$1920\times 1080$} \\
\cmidrule{3-4} \cmidrule{7-8}
View & (1,0) & (2,0) & (3,0) & (4,0) & (1,0) & (2,0) & (3,0) & (4,0) \\
\midrule
AWnet~\cite{cheng2020dual} $w/o$ $disp$ & 36.29 & 35.24 & 34.41 & 33.81 & 38.50 & 37.62 & 36.91 & 36.44 \\
Ours $w$ $disp$ & 36.45 & 35.49 & 34.74 & 34.21  & 38.85 & 38.18 & 37.58 & 37.03
\\
\bottomrule
\end{tabular}
\vspace{-15pt}
\end{table}
    \begin{figure}[t]
% \vspace{-20pt}
	\footnotesize
% 	\tiny 
% 	\scriptsize        
	\centering
	\renewcommand{\tabcolsep}{1.0pt} % adjust horizontal space
	\renewcommand{\arraystretch}{1.0} % adjust vertical space
	\begin{tabular}{cc}
	\includegraphics[width=0.5\linewidth]{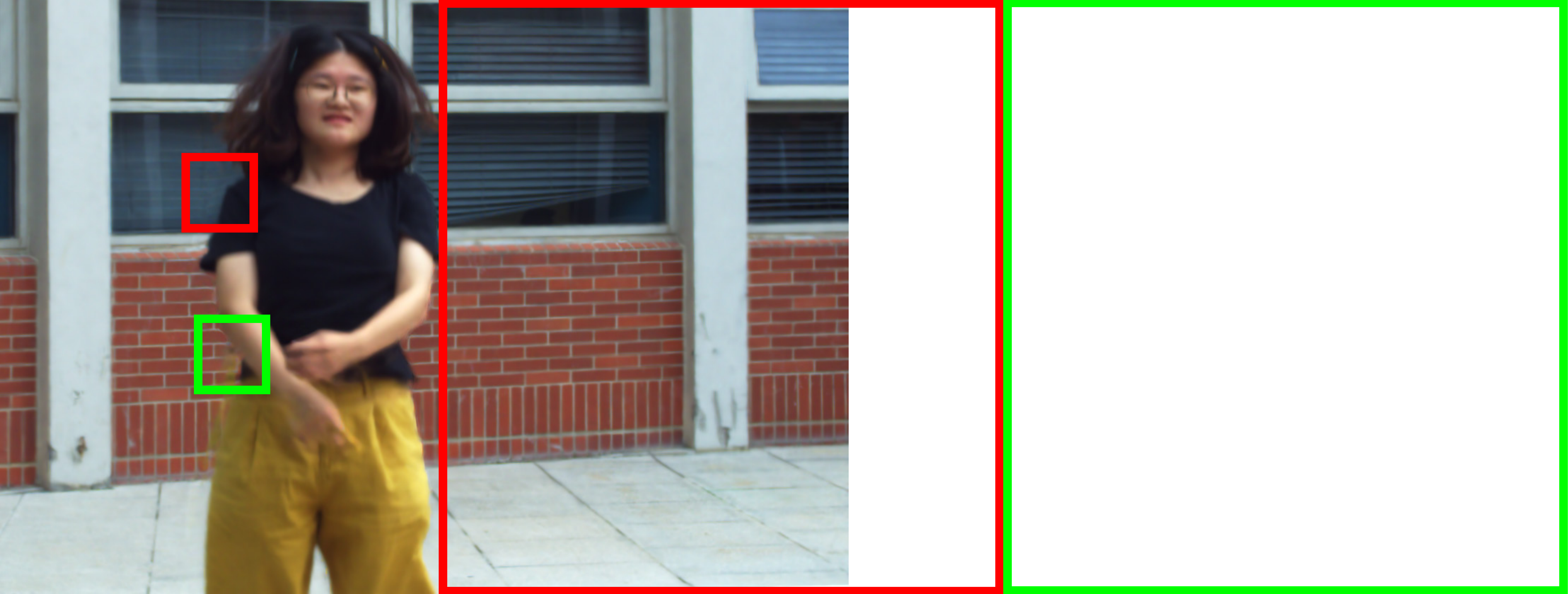} &
	\includegraphics[width=0.5\linewidth]{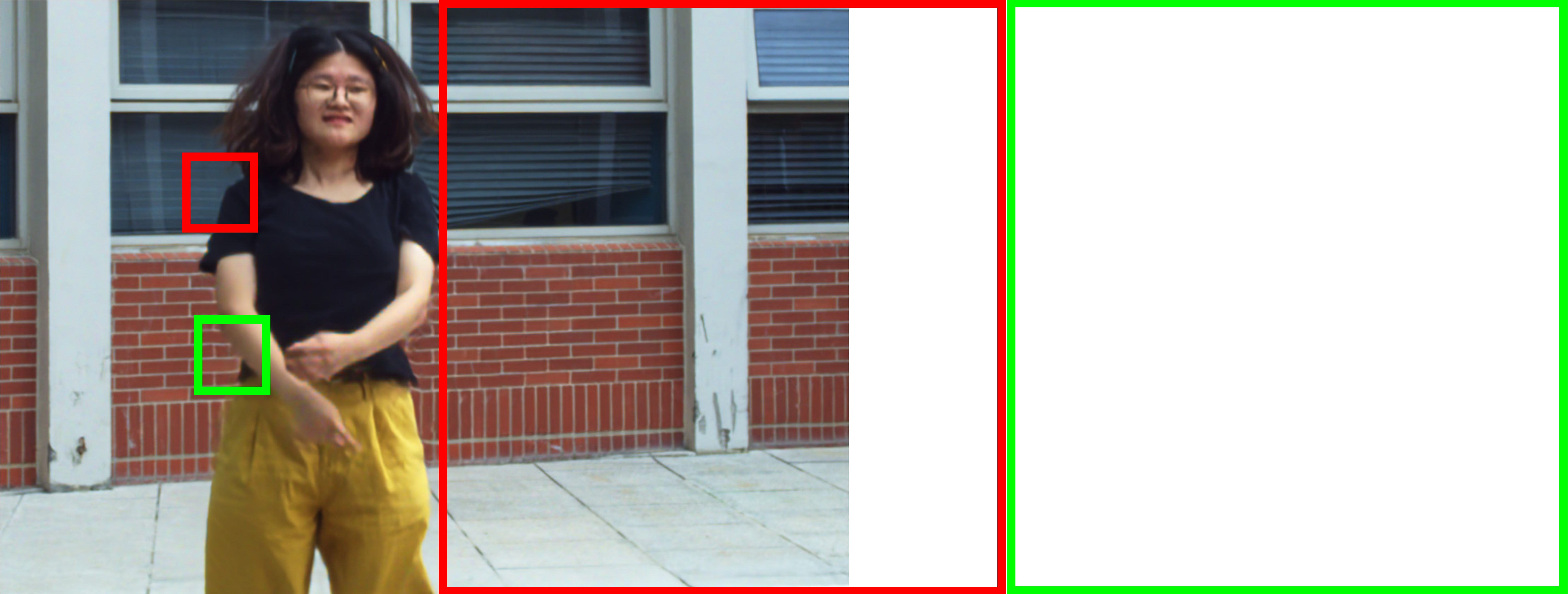} \\
	(a) AWnet~\cite{cheng2020dual} $w/o$ $disp$ & (b) Ours $w$ $disp$\\
	\end{tabular}
    \vspace{-10pt}
	\caption{\textbf{Disparity-guided flow-based warping for LSR-HFR view.} (a) Result of AWnet~\cite{cheng2020dual}. (b) Result of ours. Our method reconstructs high quality result without deformations and ghosts.}
	\label{fig:disp_LSR}
	\vspace{-10pt}
\end{figure}

    \subsection{{The Effect of Disparity Network for LSR-HFR View}}
    We evaluate the effect of the disparity network for LSR-HFR view (denoted by Ours $w$ $disp$) by comparing with the model without disparity, AWnet~\cite{cheng2020dual}, named AWnet~\cite{cheng2020dual} $w/o$ $disp$.
    We test on Light Field Video dataset~\cite{guillo2018light} with different camera baselines and image resolutions, which are two key factors affecting the value of disparity.
    Larger camera baselines or larger image resolutions will lead to larger disparities.
    The original image resolution of Light Field Video dataset~\cite{guillo2018light} is $1920\times 1080$. We use the original video and the downscaled video ($960\times 540$) as the ground truth videos for evaluation.
    We downscale the resolution of the view $(0,0)$ by $4\times$ as LSR-HFR input and reduce the frame rates by $10\times$ of the views $\{(1,0),\dots,(4,0)\}$ as HSR-LFR inputs.
    From Table~\ref{table:disp_gui}, the performance gain of disparity guidance increases as the increasing of camera baseline and image resolution.
    Our proposed disparity-guided flow-based warping achieves up to $0.67$dB performance gain against the flow-based method in AWnet~\cite{cheng2020dual}.
    There are three reasons for the outperformance of disparity-guided flow-based warping.
    1. Our disparity-guided flow-based warping method can alleviate the problem of large displacement by dividing the displacement into two smaller displacements, one is disparity and the other is flow, which will both be smaller and easier to estimate.
    2. Disparity estimation often achieves better results than flow estimation in large disparity scenes due to the one-dimension constraints of disparity.
    See the red blocks in Figure~\ref{fig:disp_LSR}, the reconstructed shutter window of AWnet~\cite{cheng2020dual} is deformed but retains straight with our disparity-guided method.
    3. The estimated disparity can help the fusion network to distinguish the view occlusion areas, which can alleviate the warping ghosts, especially in large disparity scenes.
  	See the green blocks in Figure~\ref{fig:disp_LSR}, our method can remove the warping ghosts around the arm but AWnet~\cite{cheng2020dual} can not.

\begin{figure}[t]
% \vspace{-10pt}
	\footnotesize
% 	\tiny 
% 	\scriptsize        
	\centering
	\renewcommand{\tabcolsep}{1pt} % adjust horizontal space
	\renewcommand{\arraystretch}{1} % adjust vertical space
      \newcommand{\quantTit}[1]{\multicolumn{3}{c}{\scriptsize #1}}
    \newcommand{\quantSec}[1]{\scriptsize #1}
    \newcommand{\quantInd}[1]{\scriptsize #1}
    \newcommand{\quantVal}[1]{\scalebox{0.83}[1.0]{$ #1 $}}
    \newcommand{\quantBes}[1]{\scalebox{0.83}[1.0]{$\uline{ #1 }$}}
	\begin{tabular}{cccccc}
	\multicolumn{2}{c}{\includegraphics[width=0.16\textwidth]{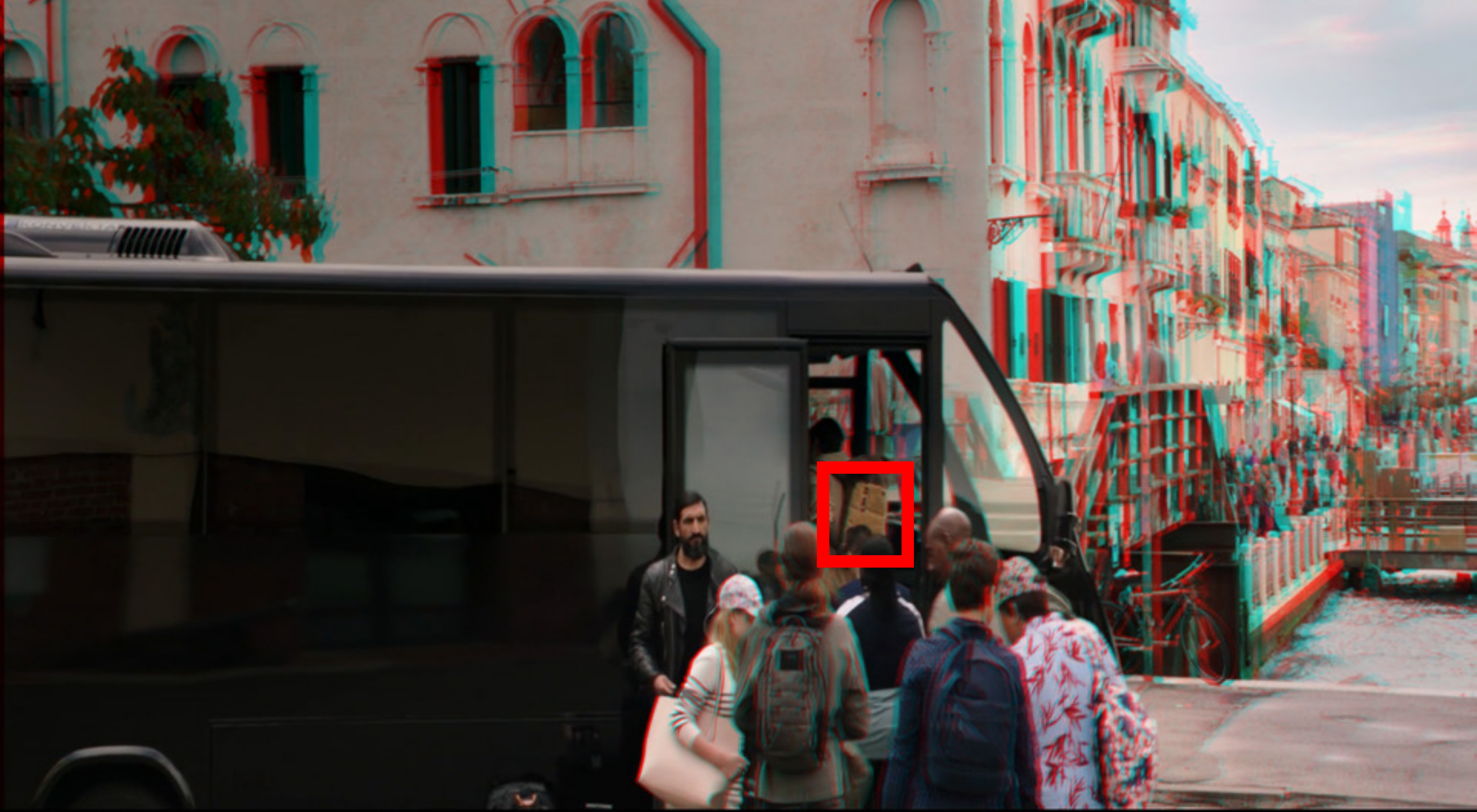}} &
	\multicolumn{2}{c}{\includegraphics[width=0.16\textwidth]{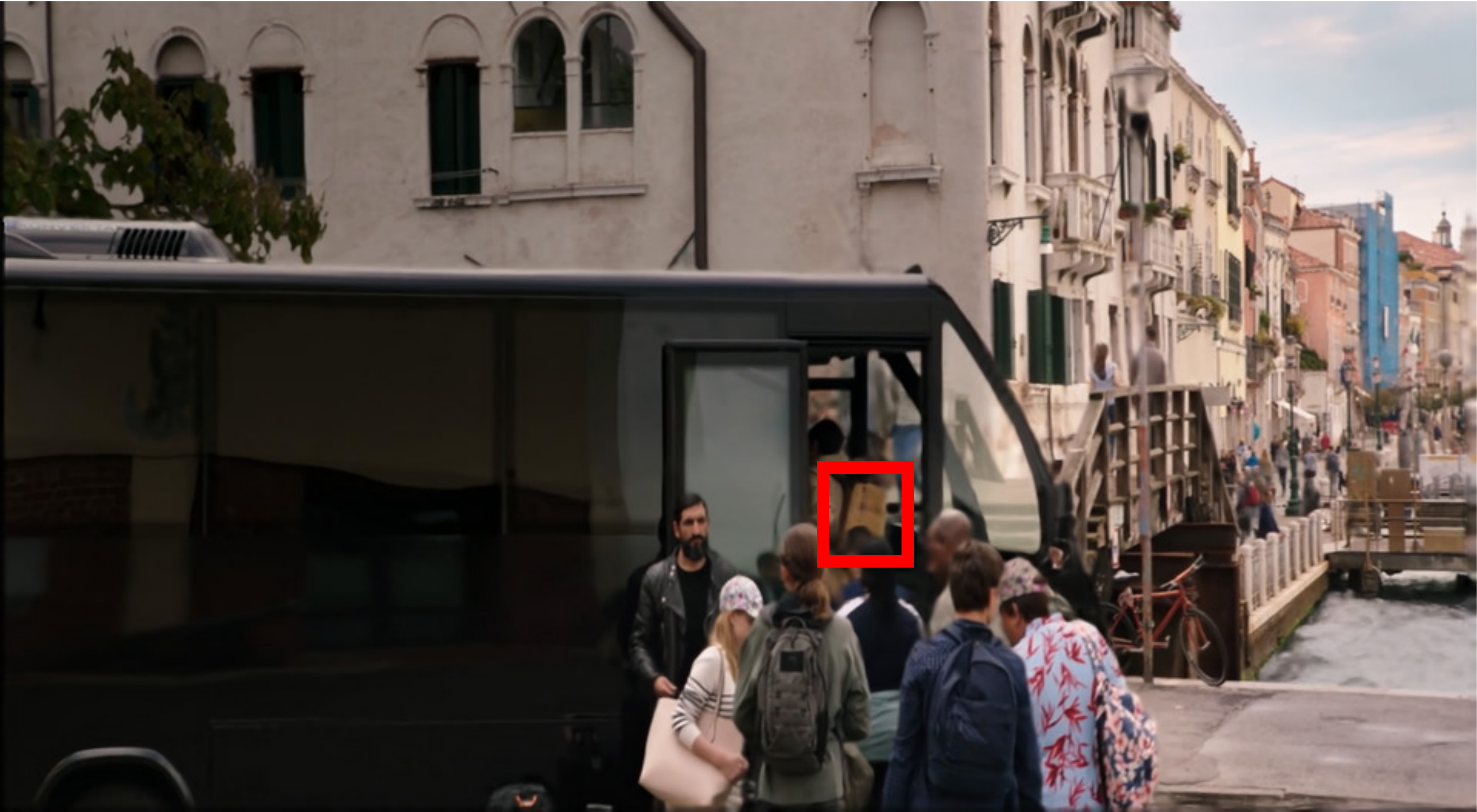}} &
    \multicolumn{2}{c}{\includegraphics[width=0.16\textwidth]{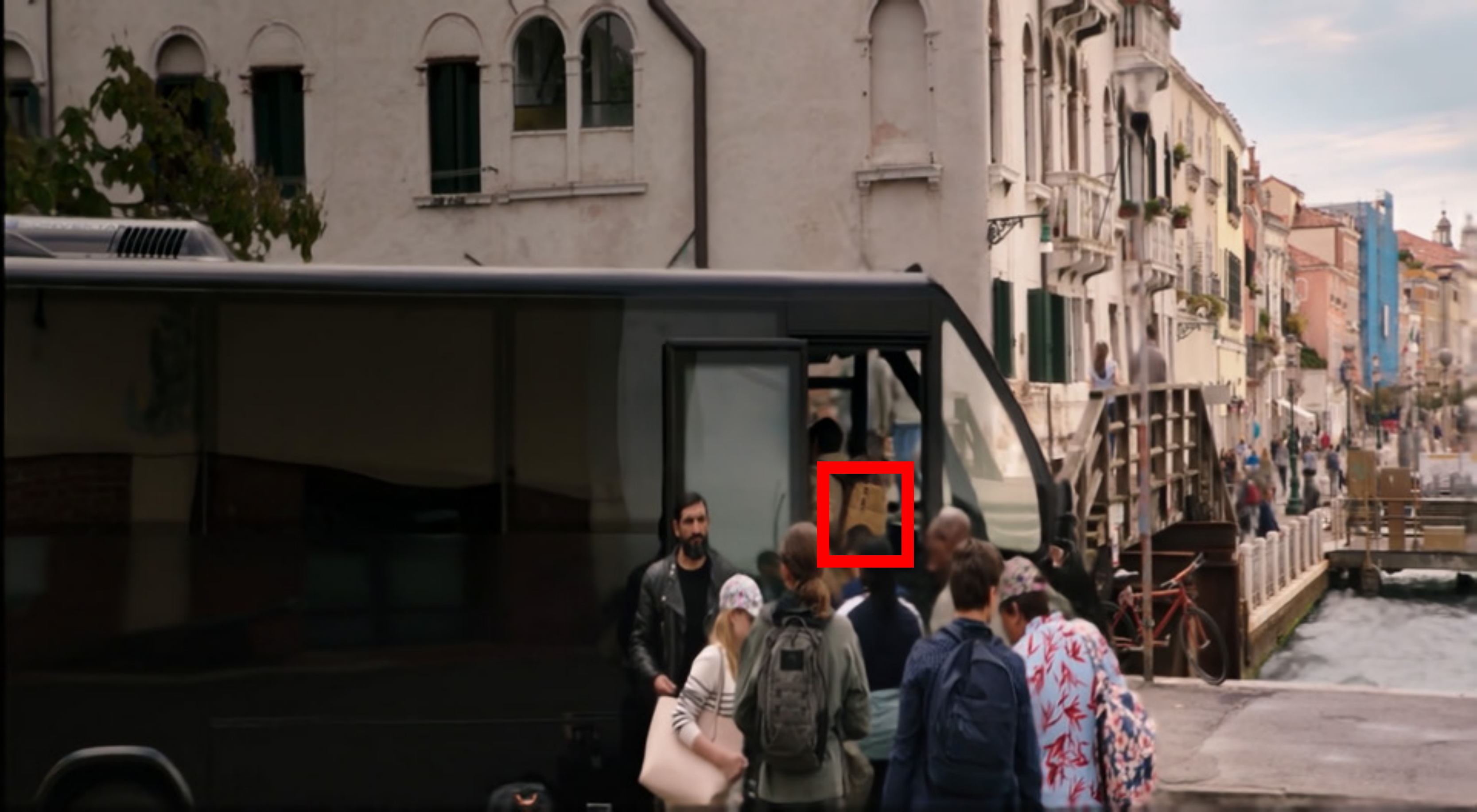}} \\
	\includegraphics[width=0.08\textwidth]{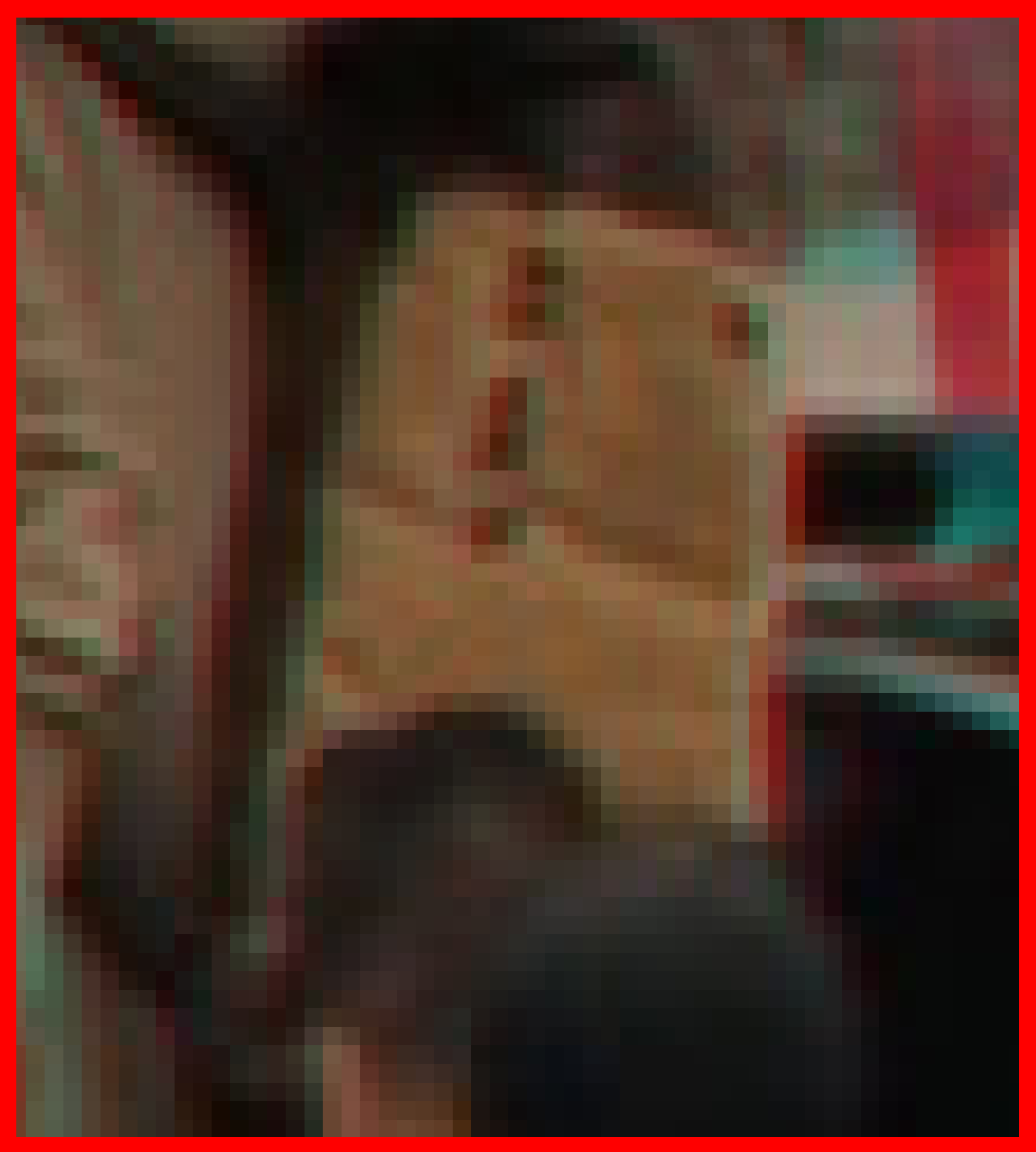} &
	\includegraphics[width=0.08\textwidth]{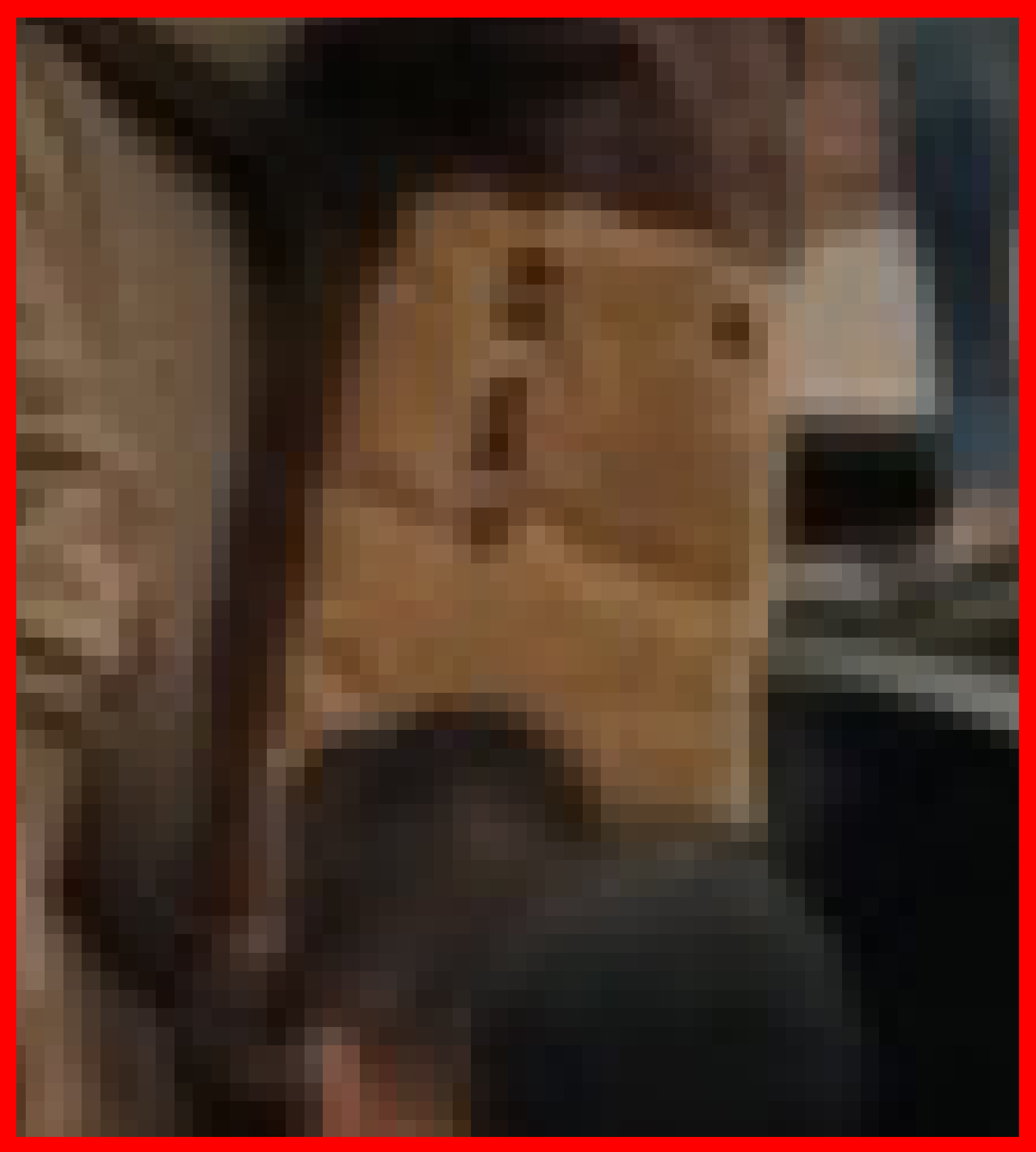} &
	\includegraphics[width=0.08\textwidth]{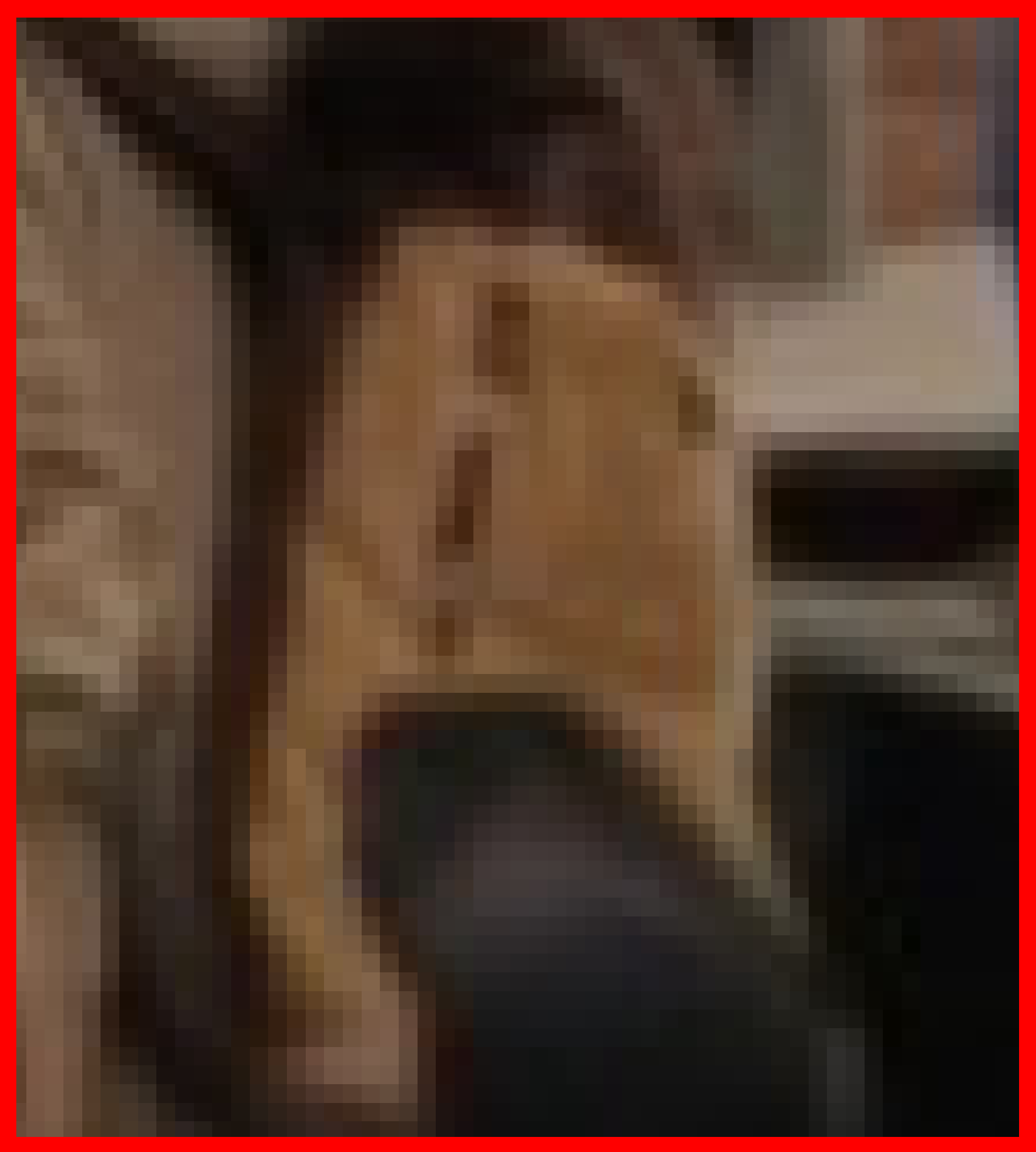} &
	\includegraphics[width=0.08\textwidth]{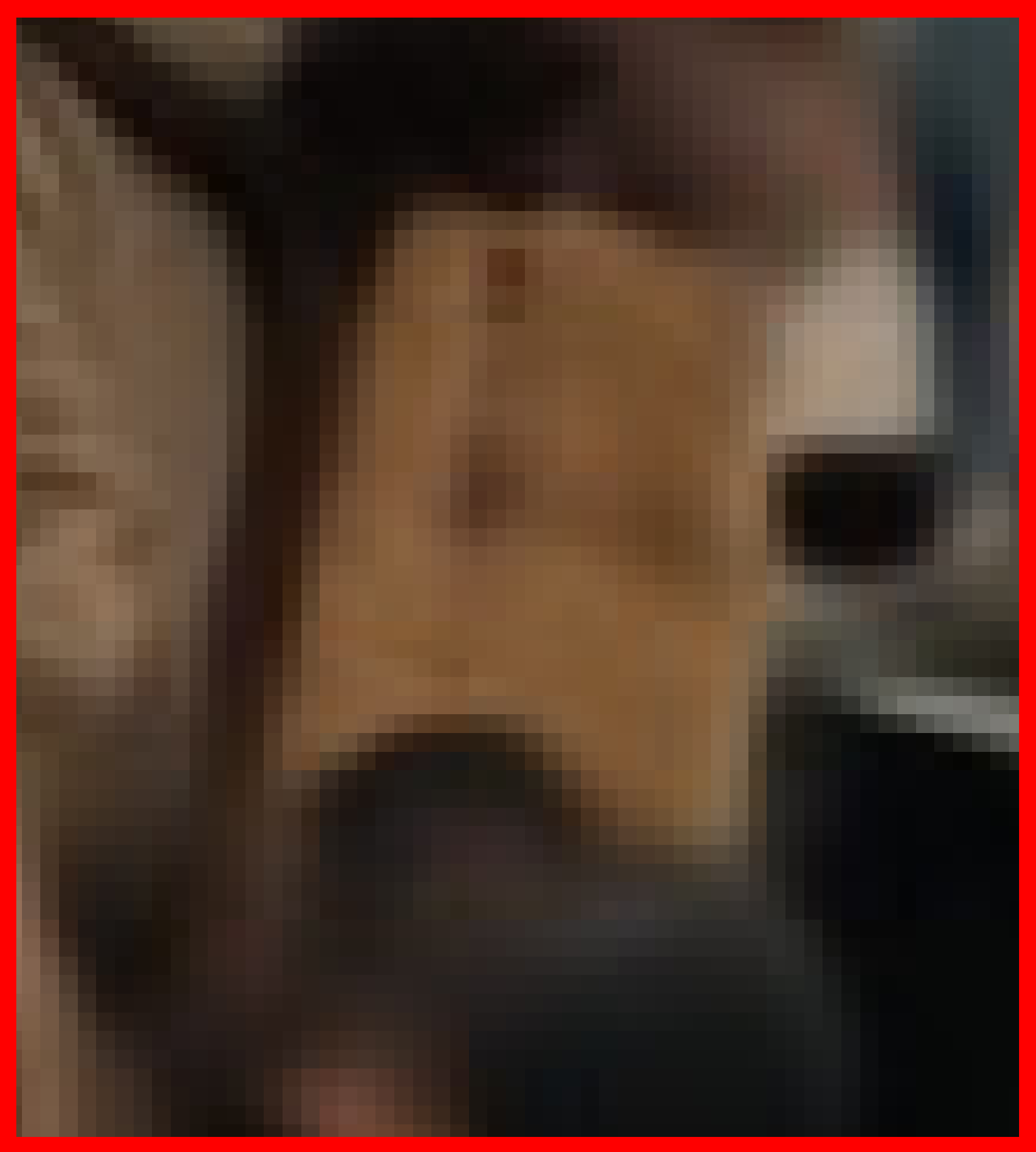} & 
	\includegraphics[width=0.08\textwidth]{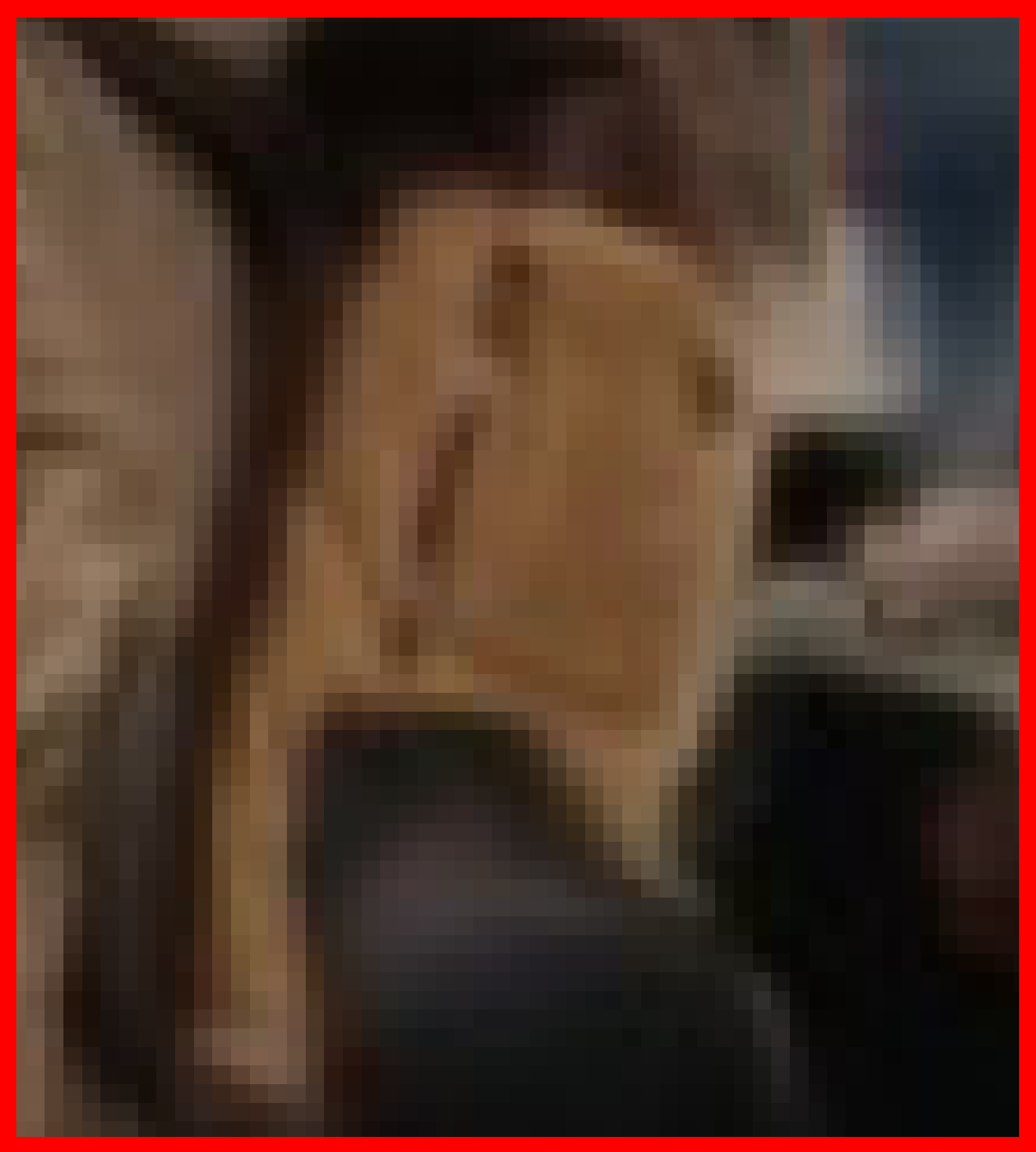} &
	\includegraphics[width=0.08\textwidth]{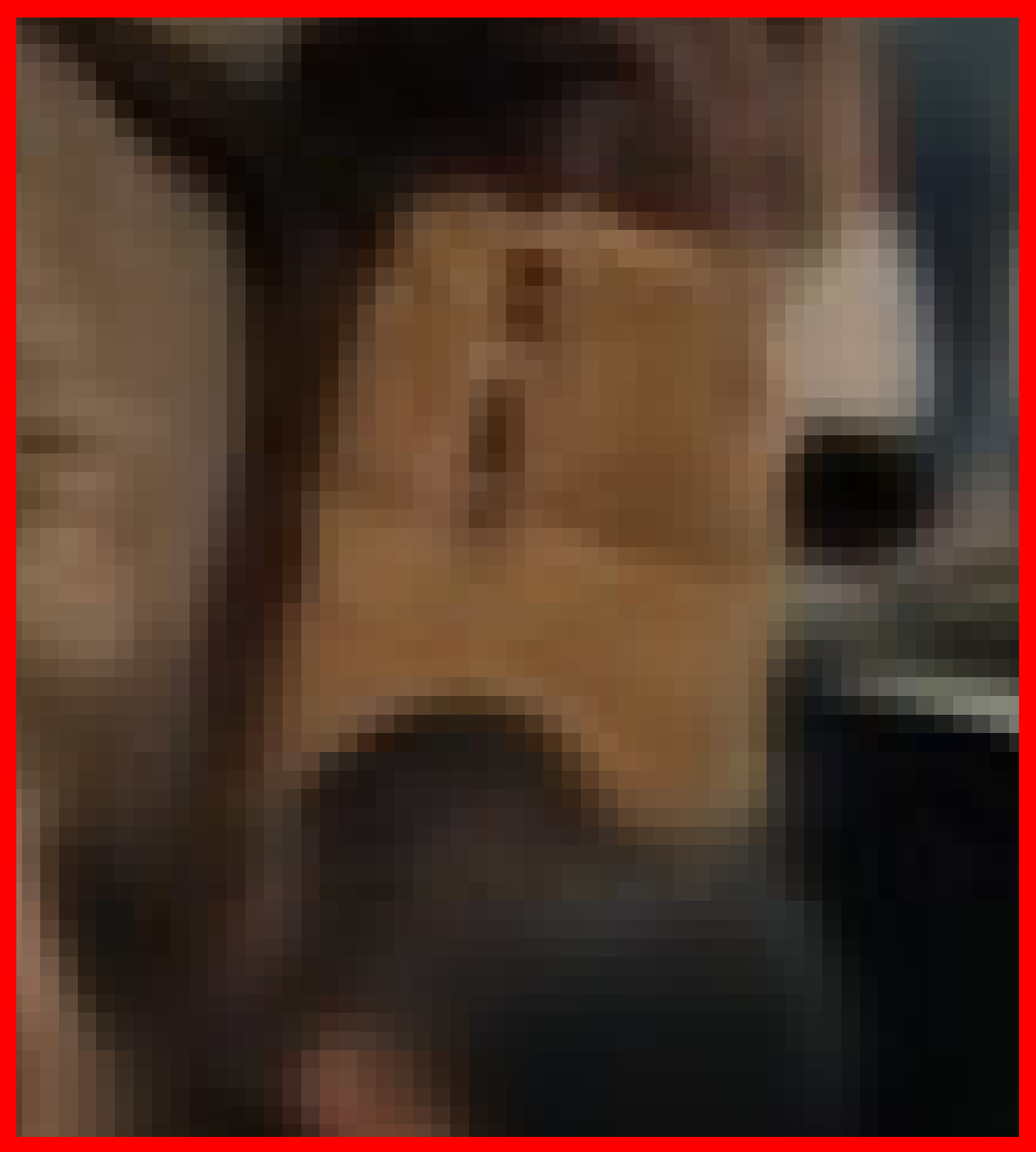} \\
	 Stereoscopic & GT & Warped & Output & Warped & Output \\
	\multicolumn{2}{c}{(a) Stereoscopic \& GT} & \multicolumn{2}{c}{(b) AWnet~\cite{cheng2020dual}} & \multicolumn{2}{c}{(c) Ours} \\
	\end{tabular}
    \vspace{-10pt}
	\caption{\textbf{Comparisons in small disparity scene.} Stereoscopic denotes the ground truth frame with stereoscopic effect. 
	GT denotes the ground truth frame in LSR-HFR view. The warped frame for AWnet denotes the warped frame based on flow. The warped frame for ours denotes the warped frame with our proposed disparity-guided flow-based warping method.}
	\label{fig:smalldisp}
	\vspace{-15pt}
\end{figure}

\begin{table}[t]
\caption{\textbf{Analysis of the complementary warping for hsr-lfr view.}}
\vspace{-5pt}
\label{table:right_ablation}
\centering
\footnotesize
\renewcommand{\tabcolsep}{5pt} % adjust horizontal space
\renewcommand{\arraystretch}{0.9}
% \vspace{-5pt}

\begin{tabular}{ l c }
\toprule
Methods & PSNR [dB] \\
\midrule
Ours & \textbf{39.19} \\
Ours w/o motion information in HSR-LFR view & \underline{38.84} \\
Ours w/o motion information from LSR-HFR view & 37.25 \\
Ours w/o disparities & 36.61 \\
\bottomrule
\end{tabular}
\vspace{-10pt}
\end{table}

    If the disparity is small, direct flow estimation between $R^0, R^T$ and $\hat{L}^t$ will not be challenging.
  	In addition, the warping artifacts introduced by view occlusion is slight.
  	Even so, our method can estimate the displacement more accurately, especially in areas with significant disparity changes around small objects.
  	As shown in Figure~\ref{fig:smalldisp}, the background of the khaki bag is a far-away building. Thus, the disparity change around the khaki bag is significant. 
  	Thus, it cannot be well aligned by flow-based warping method of AWnet~\cite{cheng2020dual}.
  	In contrast, our disparity-guided flow-based method can well align the khaki bag and restore the fine-grained textures.

    High-resolution videos have been widely used in smartphones and high-resolution display devices. Large disparity has been an unavoidable problem in hybrid camera systems.
    Therefore, the disparity network is necessary.
    
\begin{table}[t]
\caption{\textbf{Analysis of the feature-based multi-scale fusion for hsr-lfr view in psnr $\textup{[dB]}$.}}
\vspace{-5pt}
\label{table:gridnet}
\centering
\footnotesize
\renewcommand{\tabcolsep}{6pt} % adjust horizontal space
\renewcommand{\arraystretch}{0.9}
% \vspace{-5pt}

\begin{tabular}{ c c c c c }
\toprule
\textit{org} Grid & Pixel-\textit{based} & \multicolumn{3}{c}{Ours}  \\
\midrule
3-scale & 1-scale & 1-scale & 2-scale & 3-scale \\
\midrule
38.03 & 36.84 & 37.78 & \underline{38.49} & \textbf{39.19} \\
\bottomrule
\end{tabular}
\vspace{-15pt}
\end{table}
\begin{figure}[t]
% \vspace{-20pt}
	\footnotesize
% 	\tiny 
% 	\scriptsize        
	\centering
	\newcommand{\quantTit}[1]{\multicolumn{3}{c}{\scriptsize #1}}
    \newcommand{\quantSec}[1]{\scriptsize #1}
    \newcommand{\quantInd}[1]{\scriptsize #1}
    \newcommand{\quantVal}[1]{\scalebox{0.83}[1.0]{$ #1 $}}
    \newcommand{\quantBes}[1]{\scalebox{0.83}[1.0]{$\uline{ #1 }$}}
	\renewcommand{\tabcolsep}{0.5pt} % adjust horizontal space
	\renewcommand{\arraystretch}{1.0} % adjust vertical space
	\begin{tabular}{cccccc}
	\includegraphics[width=0.165\linewidth]{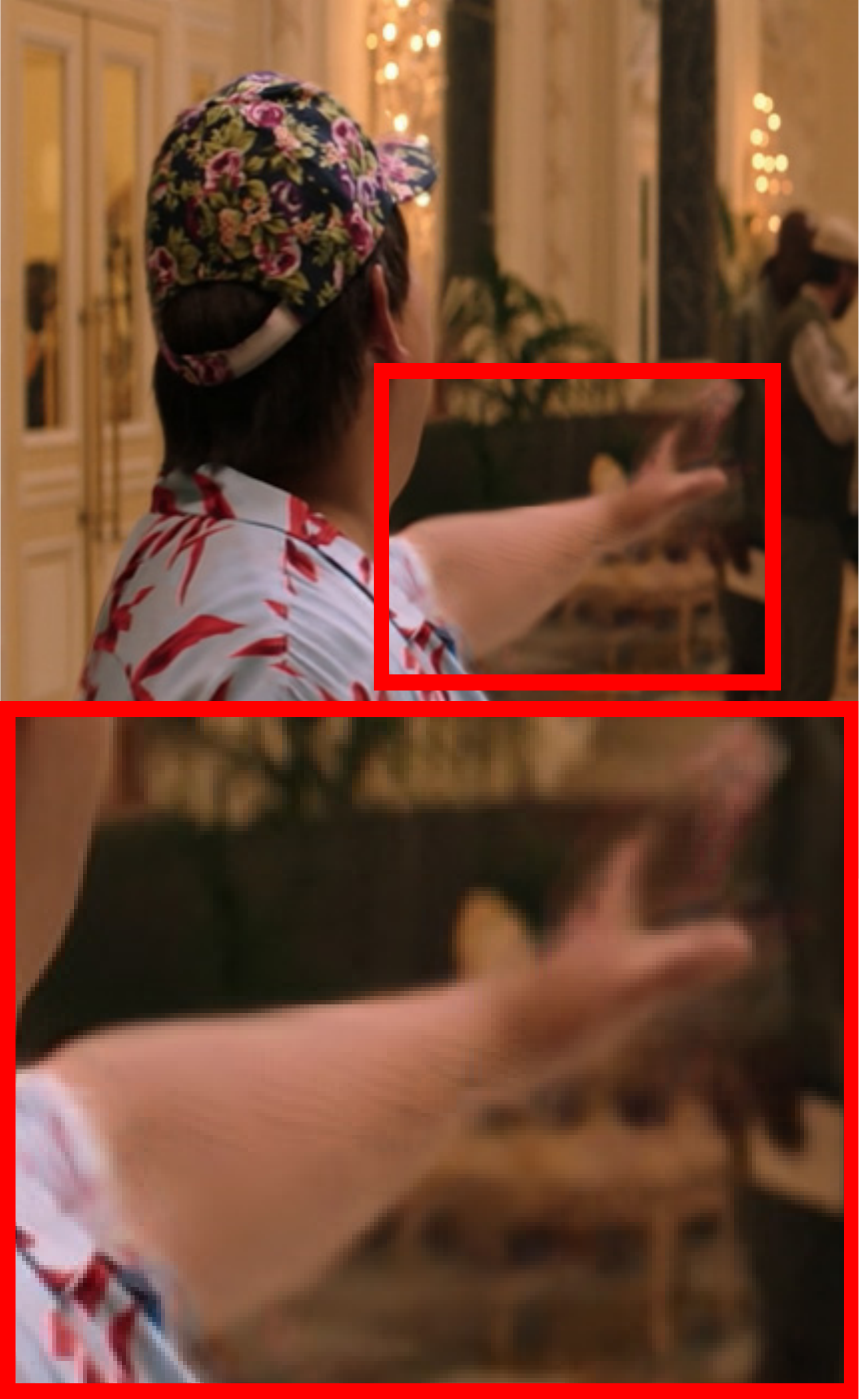} &
	\includegraphics[width=0.165\linewidth]{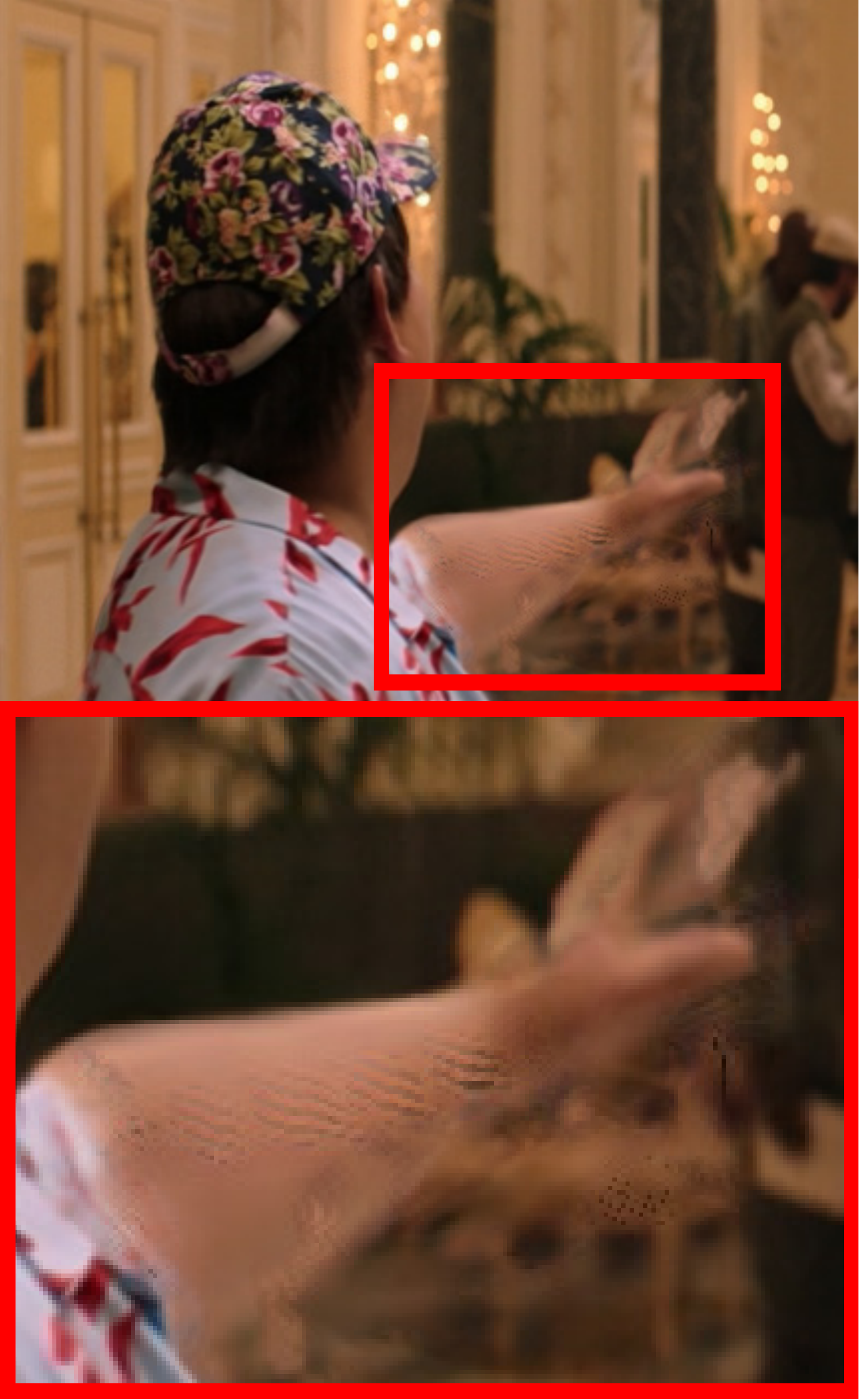} &
	\includegraphics[width=0.165\linewidth]{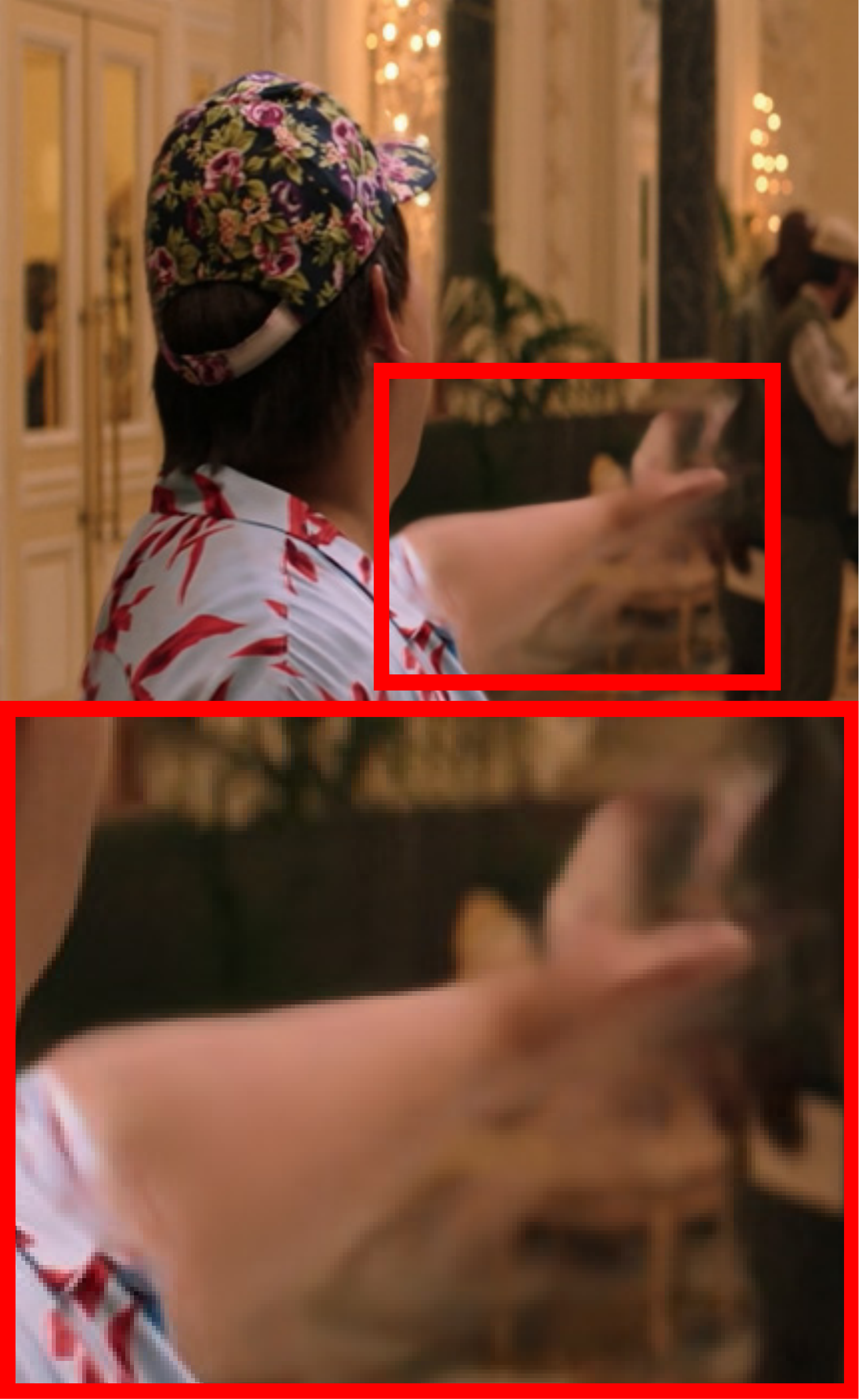} &
	\includegraphics[width=0.165\linewidth]{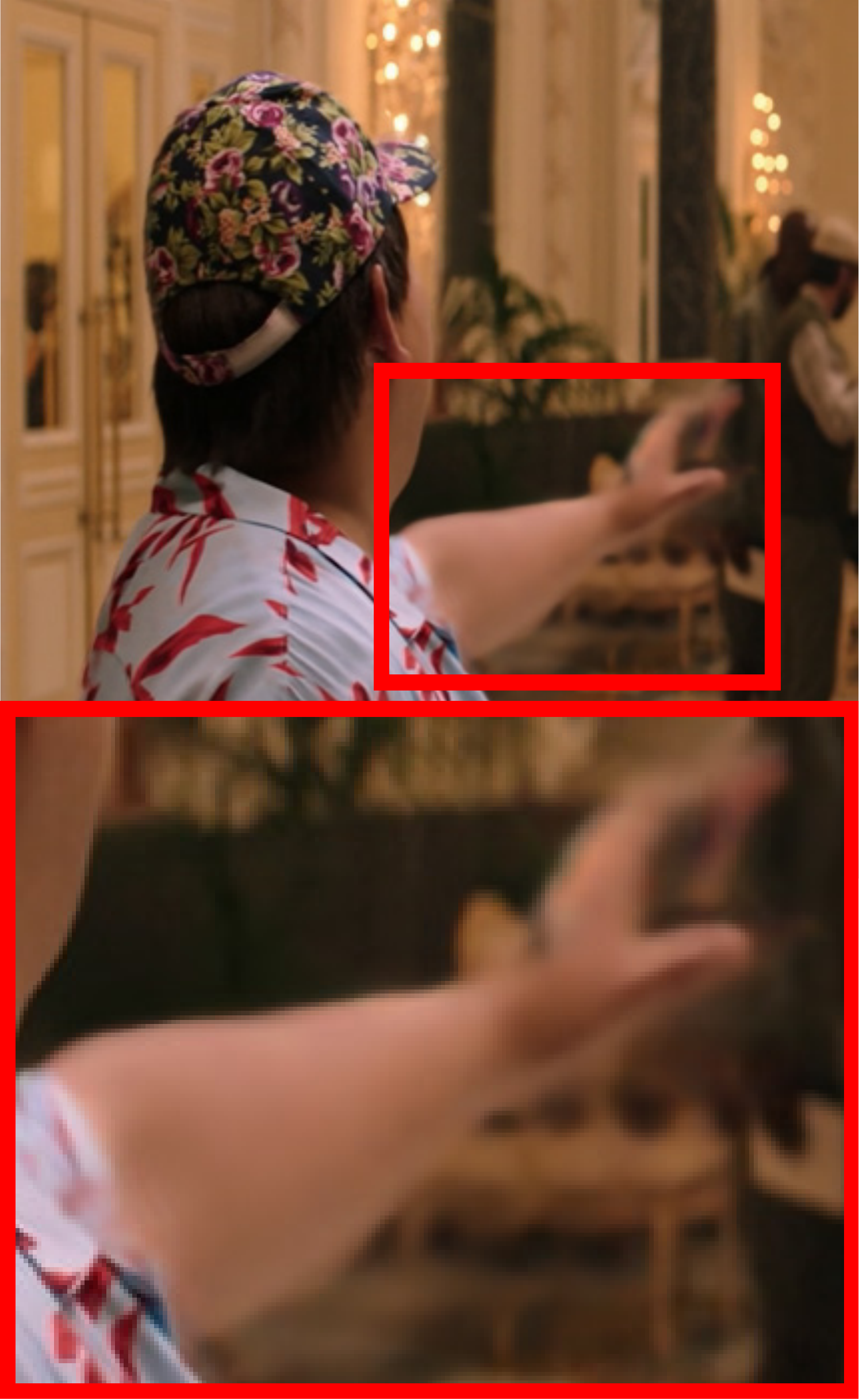} &
	\includegraphics[width=0.165\linewidth]{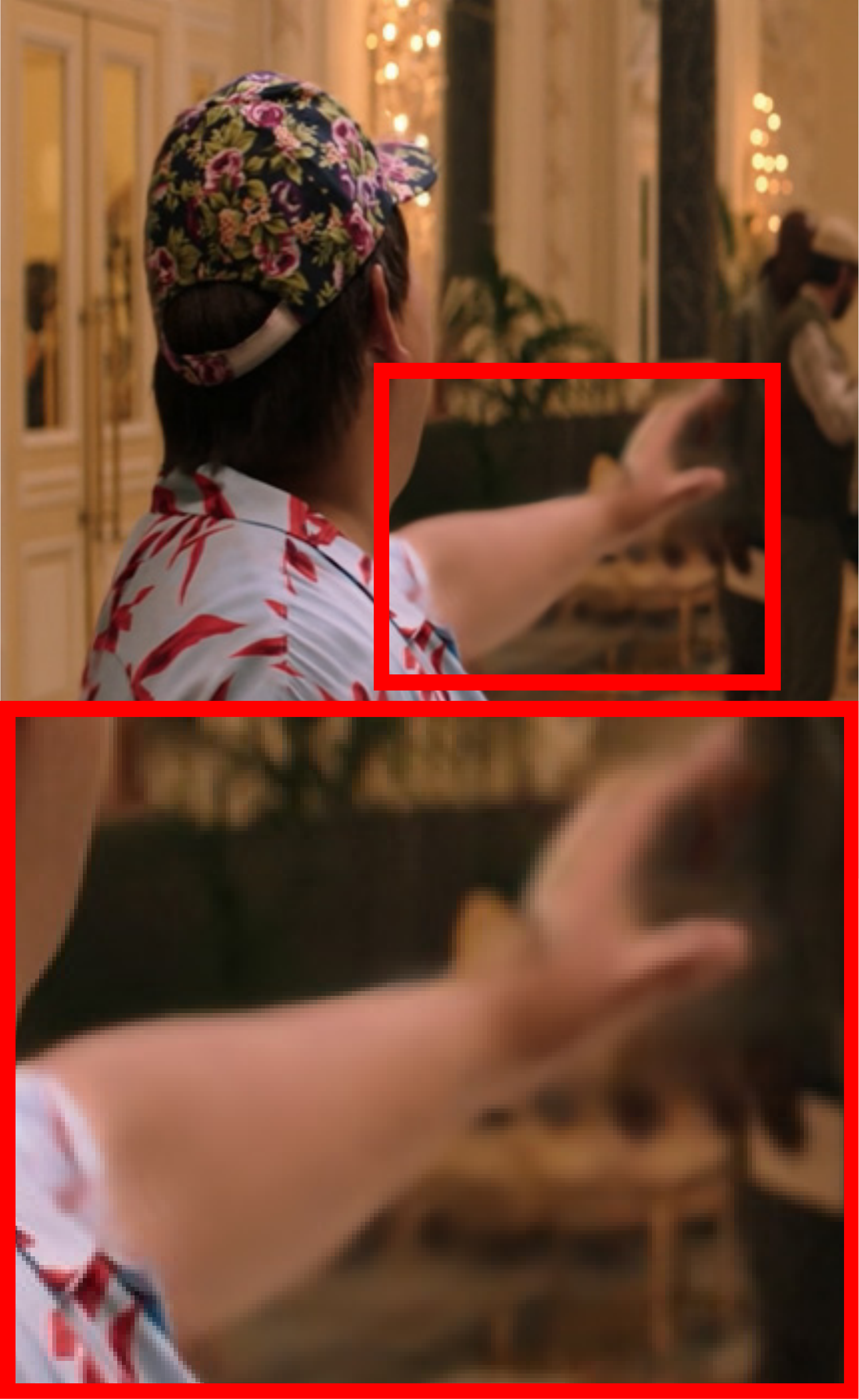} &
	\includegraphics[width=0.165\linewidth]{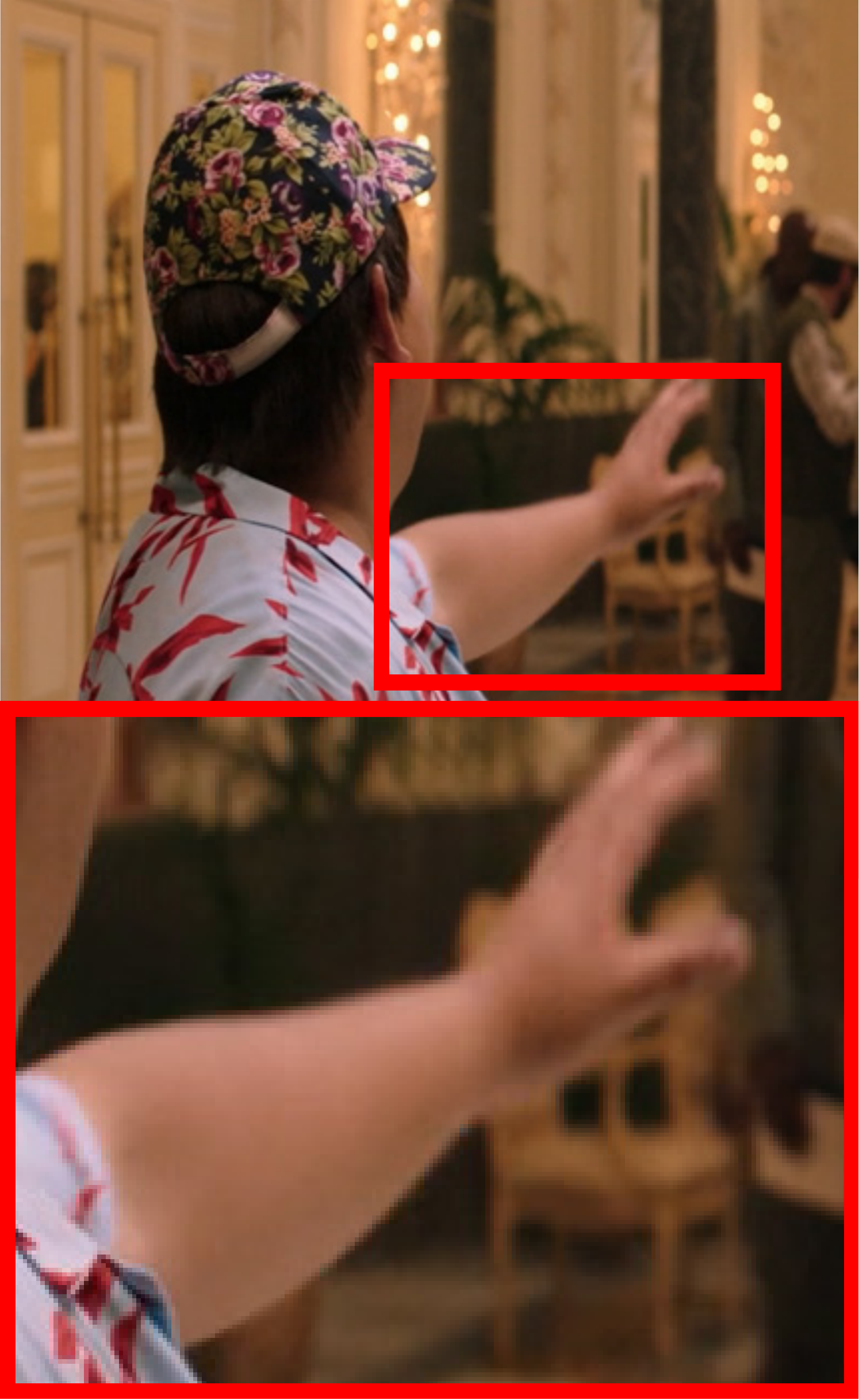} \\
	(a) \textit{org} Grid & \scriptsize{(b) pixel \textit{based}} & (c) Ours$_{1s}$ & (d) Ours$_{2s}$ & (e) Ours$_{3s}$ & (f) GT \\
	\end{tabular}
    % \vspace{-10pt}
	\caption{\textbf{Ablation on feature-based multi-scale fusion.} (a) Result of original GridNet~\cite{fourure2017residual} architecture with three scales of features as inputs. (b) Result of fusion with one scale of pixel. (c), (d) and (e) are the results of fusing one, two and three scales of features using our proposed architecture. (f) Ground truth frame in HSR-LFR view. 
	}
	\label{fig:fuse_abla}
	\vspace{-15pt}
\end{figure}
\begin{table}[t]
% \vspace{-5pt}
\caption{\textbf{Effect of spatiotemporal resolutions.}
    The numbers are the psnr $\textup{[dB]}$ of the results in lsr-hfr/hsr-lfr views.
    }
\label{table:spatiotemporal}
\vspace{-5pt}
\centering
\footnotesize
\newcommand{\quantTit}[1]{\multicolumn{3}{c}{\scriptsize #1}}
    \newcommand{\quantSec}[1]{\scriptsize #1}
    \newcommand{\quantInd}[1]{\scalebox{0.95}[1.0]{\scriptsize #1}}
    \newcommand{\quantVal}[1]{\scalebox{0.83}[1.0]{$ #1 $}}
    \newcommand{\quantBes}[1]{\scalebox{0.83}[1.0]{$\uline{ #1 }$}}
\renewcommand{\tabcolsep}{6.0pt} % adjust horizontal space
\renewcommand{\arraystretch}{0.9} % adjust vertical space
% \vspace{-5pt}
\begin{tabular}{ c  c  c  c }
\toprule
\diagbox{Temporal}{Spatial} & $ 4\times \downarrow$ & $ 6\times \downarrow $ & $ 8\times \downarrow$ \\ 
\midrule
$ 2\times \downarrow$ &  40.59/40.78 & 38.64/40.37 & 36.76/39.59 \\
$ 4\times \downarrow$ & 40.22/39.19 & 38.15/38.63 & 36.15/37.68 \\
$ 6\times \downarrow$   & 39.94/38.15 & 37.80/37.55 & 35.77/36.58 \\
\bottomrule
\end{tabular}
\vspace{-15pt}
\end{table} 
\begin{figure}[t]
\begin{center}
\includegraphics[width=0.9\linewidth]{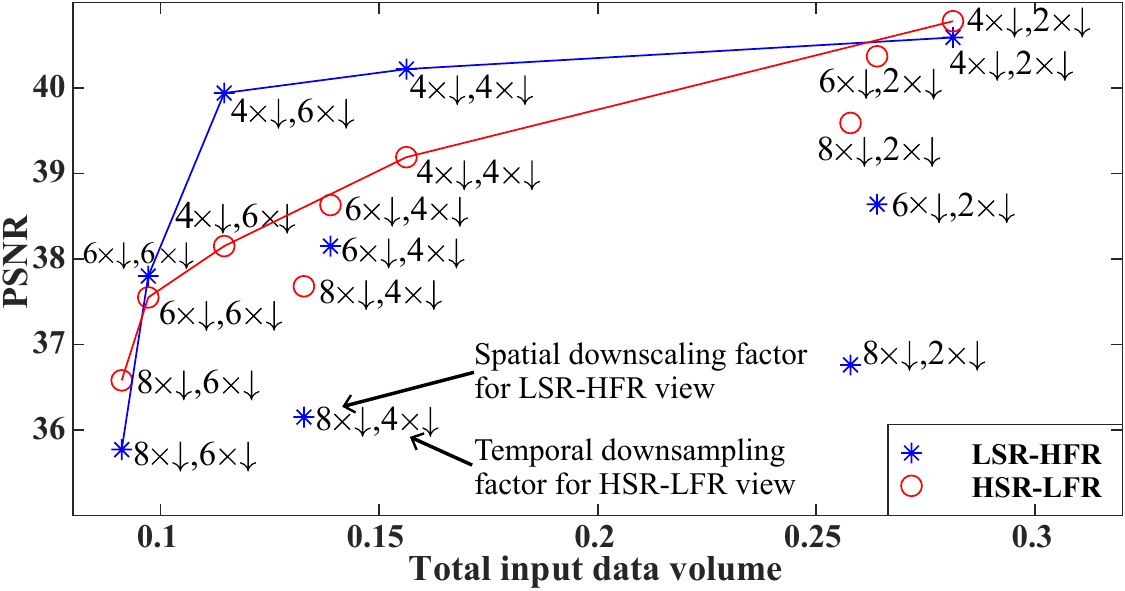}
\end{center}
  \vspace{-14pt}
  \caption{\textbf{Spatiotemporal resolution combinations.} The X-axis represents the total input data volume, which is the sum of the data volumes of both input views. The Y-axis represents the PSNR of the reconstructed frame. {\color{blue}blue *} denotes the results in LSR-HFR view and {\color{red}red O} denotes the results in HSR-LFR view. The texts beside the points, such as $6\times\!\downarrow,4\times\!\downarrow$, are spatial downscaling factors for LSR-HFR view and temporal downsampling factors for HSR-LFR view, respectively.}
\label{fig:optima}
 \vspace{-10pt}
\end{figure}   
% \begin{table}[htbp]
% \caption{Ablation study on the baseline.}
% \label{table:baseline}
% \centering
% \renewcommand
% \arraystretch{1.05}
% \begin{tabular}{ c c c c c }
% % \specialrule{0em}{1pt}{1pt}
% \toprule
% & (0,1) & (0,2) & (0,3) & (0,4)\\
% \midrule
% Boxer & 36.53/37.40 & 35.20/37.40 & 34.17/37.14 & 33.51/36.68       \\
% chess & 37.13/37.76 & 35.87/37.71 & 34.91/37.39 & 34.24/36.98       \\
% ChessPieces & 35.69/34.99 & 35.41/34.66 & 35.15/34.31 & 34.87/34.05 \\
% \bottomrule
% \end{tabular}
% \end{table}

\begin{table}[t]
% \vspace{-5pt}
\caption{\textbf{Effect of camera baseline in psnr $\textup{[dB]}$.}}
\vspace{-5pt}
\label{table:baseline}
\centering
\footnotesize
\newcommand{\quantInd}[1]{\scalebox{0.8}[1.0]{\scriptsize #1}}
\renewcommand{\tabcolsep}{10pt} % adjust horizontal space
\renewcommand{\arraystretch}{0.9} % adjust vertical space
% \vspace{-5pt}
\begin{tabular}{ccccc}
\toprule
HSR-LFR view =& (1,0) & (2,0) & (3,0) & (4,0)\\
\midrule
LSR-HFR (0,0) & 36.45 & 35.49 & 34.74 & 34.21 \\
HSR-LFR & 36.72 & 36.59 & 36.28 & 35.90 \\
\bottomrule
\end{tabular}
\vspace{-10pt}
\end{table}  
\begin{table}[h!]
\caption{\textbf{Effect of temporal desynchronization between hsr-lfr and lsr-hfr videos in psnr $\textup{[dB]}$.}}
\label{table:desyn}
\centering
\footnotesize
\renewcommand{\tabcolsep}{10pt} % adjust horizontal space
\renewcommand{\arraystretch}{0.9}
\vspace{-5pt}

\begin{tabular}{ l c c c c }
\toprule
\#frames & 0 & 1 & 2 & 3 \\
\midrule
LSR-HFR (0,0) & 36.45 & 35.86 & 35.50 & 35.42 \\
HSR-LFR (1,0) & 36.72 & 34.51 & 31.36 & 29.03 \\
\bottomrule
\end{tabular}
\vspace{-15pt}
\end{table}

    \subsection{Complementary Warping for HSR-LFR View}
    We evaluate the performance gain of each warping component for HSR-LFR view by removing the corresponding warping frames from the fusion inputs.
    Here, the Stereo Video dataset is used for evaluation.
    As shown in Table~\ref{table:right_ablation}, all the three warping components are critical for the synthesis of the missing frame in HSR-LFR view.
    The warped frames based on disparities and motion information from LSR-HFR view significantly improve the reconstruction performance, 2.58dB and 1.94dB, which are more significant than the motion information in HSR-LFR view.
    We present visual demonstrations of the complementary warped frames in Figure~\ref{fig:output}(g)(h)(i). The warped frame using the motion information in HSR-LFR view retains the high-quality textures in static and small motion regions but fails in fast-moving regions. The warped frame using the motion information from LSR-HFR view restores the accurate location of the fast-moving objects but with slight distortions in motion occlusion regions. The warped frame based on disparity retains the fine-grained shape of the fast-moving objects. However, it introduces warping holes in view occlusion regions.
    The subjective performances of the warped frames in Figure~\ref{fig:output} are consistent with the objective results in Table~\ref{table:right_ablation}.
    All of the warped frames complement each other, resulting in superior performance, as shown in Table~\ref{table:right_ablation}.

    \subsection{Feature-based Multi-scale Fusion for HSR-LFR View}
    We present the objective and subjective ablation experiments on the feature-based multi-scale fusion network in Table~\ref{table:gridnet} and Figure~\ref{fig:fuse_abla}.
    For comparison, we retrain the original GridNet~\cite{fourure2017residual} architecture with three scales of features as inputs for image fusion (denoted by \textit{org} Grid).
    According to Figure~\ref{fig:fuse_abla}, our variant architecture with bilinear upscaling and PReLu activation layer can restore much more accurate appearances and smoother textures. From the quantitative results in  Table~\ref{table:gridnet}, our method achieves 1.16dB performance gain against the original GridNet~\cite{fourure2017residual} architecture.
    The reason is that the bilinear upscaling layer avoids checkerboard artifacts due to the deconvolution layer and results in more smooth textures.
    In addition, the PReLU activation layer avoids the ``dead ReLU'' problem. Moreover, its slope in the negative part can be learned to be a more specialized activation. Thus, the PReLU activation layer outperforms ReLU.
    To understand the effect of the fusion in feature domain, we trained a fusion network that only receives one-scale warped pixels as input (denoted by pixel-\textit{based}). From Table~\ref{table:gridnet}, adding one scale of features into the fusion network achieves 0.94dB gain than fusing in pixel domain. From Figure~\ref{fig:fuse_abla}(b)(c), the smooth textures of the arm are preserved by feature-based fusion.
    {It demonstrates that warping and fusing in feature domain retain more context information of the warped frames and restore smoother textures than in pixel domain.}
    We also analyze the effect of the number of feature scales by inputting different scales of warped features into the fusion network.
    Our model with three feature scales has a 1.41dB performance gain against one scale, verifying the effectiveness of our multi-scale strategy.
    From Figure~\ref{fig:fuse_abla}(c)-(e), fusing with three feature scales restores complete appearances of the fingers.
    {It demonstrates that multi-scale warping method retains more context information, and our multi-scale fusion network effectively fuses the multi-scale features.}

    \subsection{Effect of Spatiotemporal Resolutions}
    We evaluate the effect of different spatiotemporal resolutions combinations with Stereo Video dataset.
    We downscale the spatial resolution of the left view by $4\times$, $6\times$ or $8\times$ as LSR-HFR input, and reduce the temporal resolution of the right view by $2\times$, $4\times$ or $6\times$ as HSR-LFR input.
    As shown in Table~\ref{table:spatiotemporal}, the results improve as the spatiotemporal resolutions increase.
    Even with an extremely low-frame-rate ($\frac{1}{6}$ frame rate) or low-resolution ($\frac{1}{8}\times \frac{1}{8}$ resolution) video for compensation, our method still outperforms existing super-resolution method EDVR~\cite{wang2019edvr} (37.56dB on Stereo Video dataset in Table~\ref{table:stereoresults}) and frame interpolation method DAIN~\cite{bao2019depth} (33.29dB on Stereo Video dataset in Table~\ref{table:stereoresults}) with significant gains of 2.38dB and 4.39dB, respectively.
    In addition, the results also demonstrate the robustness of our method to the variation of spatial and temporal resolutions.

    The optimal spatiotemporal combination of LSR-HFR and HSR-LFR videos is a critical problem in practice.
    We try to find the optimal spatiotemporal combinations experimentally.
    Based on the results in Table~\ref{table:spatiotemporal}, We plot the total input data volumes and corresponding PSNRs of both views in Figure~\ref{fig:optima}, where {\color{red}red O} denotes the result in HSR-LFR view and {\color{blue}blue *} denotes the result in LSR-HFR view.
    We assume that the data volume for each view at the original spatiotemporal resolution is 0.5, then 1 for two views.
    The total input data volume is the sum of the data volumes of both views. 
    We plot an optima curve for each view to maximal the output PSNR at each total input data volume, which is the top half of the convex hull of the result points calculated with \textit{convexHull} function in Matlab.
  	The code for the optimal curve is presented in the supplementary material.
    We use \textit{convexHull} function to calculate the smallest polygon that contains all the results in this polygon and remove the bottom half of the polygon.
    Then, all the result points will be under the curve.
    Thus, the point on the optima curves should be the theoretical optimal PSNR at each data volume.
    The best spatiotemporal combination can be interpolated by the two adjacent result points on the curve.
    For example, the best spatiotemporal combination at the data volume of 0.2 is $4\times$ spatially downscaling for LSR-HFR view, and $3\times$ frame rate reduction for HSR-LFR view.

    \subsection{Effect of Camera Baseline}
    We use Light Field Video dataset~\cite{guillo2018light} to analyze the effect of camera baseline.
    From Table~\ref{table:baseline}, the results degrade as the baseline increases.
    {
    The reason is that disparity estimation accuracy decreases as the camera baseline increases.
    In addition,  the overlap of the field of view of the two views becomes smaller as the camera baseline increases. At the same time, the view occlusion becomes larger.
    Then, the amount of the shared spatiotemporal information between the two views decreases with the increase of the camera baseline.
    }
    Nevertheless, our LIFnet produces better results than super-resolution and frame-interpolation methods even with large baseline, which demonstrates the effectiveness of our method in transferring spatiotemporal information across views.

  \begin{table}[t]
% \vspace{-5pt}
\caption{\textbf{Robustness to noise and blur in psnr {\upshape [dB]}.}}
\label{table:longshort}
\centering
\renewcommand{\tabcolsep}{4pt} 
\vspace{-5pt}
\begin{tabular}{c | c c c c c}
\toprule
\diagbox[width=5em,trim=l]{blur}{$\sigma$} & 5 & 10 & 15 & 20 \\
\midrule
3 & 37.79/38.65 & 36.87/37.93 & 36.03/37.16 & 32.25/35.32 \\
5 & 37.63/38.48 & 36.72/37.76 & 35.89/37.02 & 32.26/35.22\\
7 & 37.36/37.65 & 36.48/37.03 & 35.67/36.39 & 32.14/34.73\\
\bottomrule
\end{tabular}
\vspace{-10pt}
\end{table}
    \begin{figure}[t]
% \vspace{-20pt}
	\footnotesize
% 	\tiny 
% 	\scriptsize        
	\centering
	\renewcommand{\tabcolsep}{1.1pt} % adjust horizontal space
	\renewcommand{\arraystretch}{1} % adjust vertical space
      \newcommand{\quantTit}[1]{\multicolumn{3}{c}{\scriptsize #1}}
    \newcommand{\quantSec}[1]{\scriptsize #1}
    \newcommand{\quantInd}[1]{\scriptsize #1}
    \newcommand{\quantVal}[1]{\scalebox{0.83}[1.0]{$ #1 $}}
    \newcommand{\quantBes}[1]{\scalebox{0.83}[1.0]{$\uline{ #1 }$}}
	\begin{tabular}{cc}
	\includegraphics[width=0.15\textwidth]{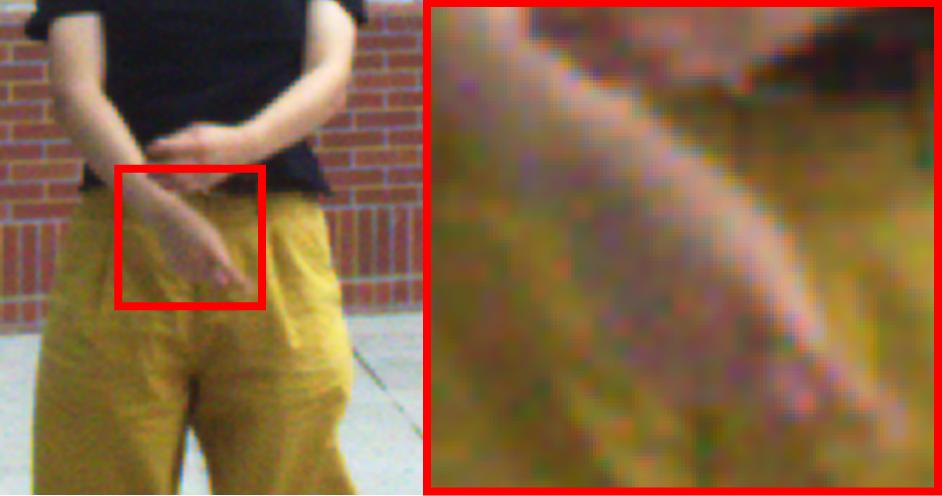} &
	\includegraphics[width=0.15\textwidth]{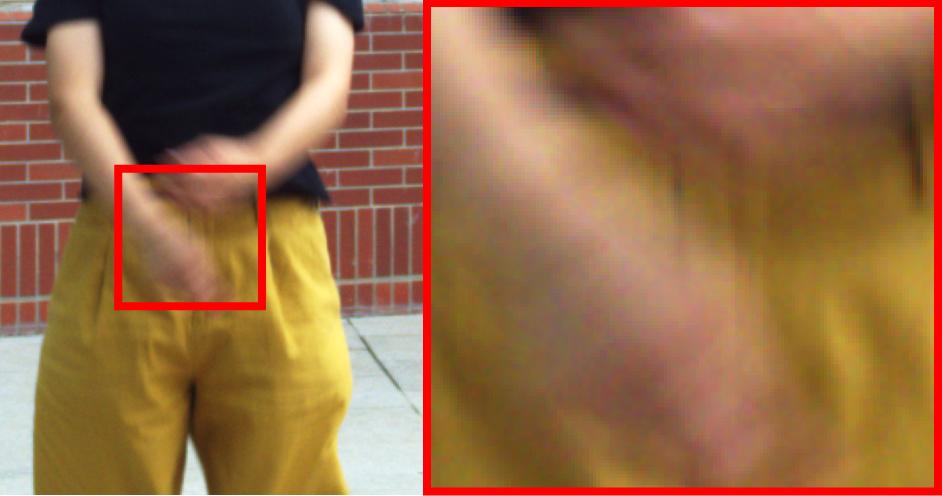} \\
	(a) LSR-HFR input & (b) HSR-LFR input \\
	\includegraphics[width=0.15\textwidth]{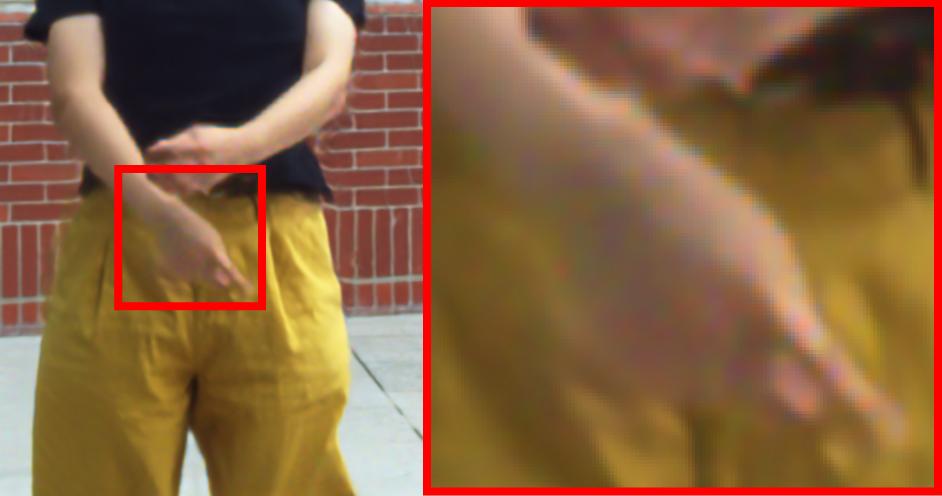} &
    \includegraphics[width=0.15\textwidth]{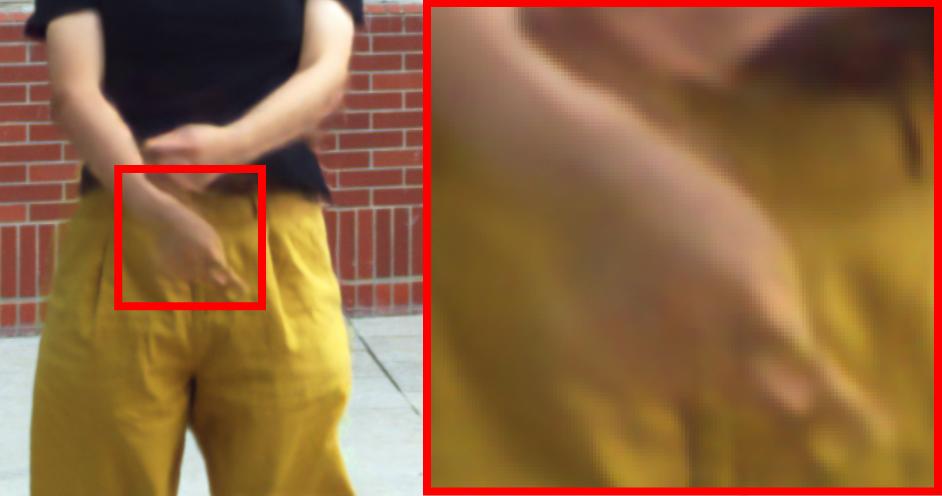} \\
	(c) Ours (LSR-HFR) & (d) Ours (HSR-LFR) \\
    \end{tabular}
    % \vspace{-5pt}
	\caption{\textbf{Robustness to noise and blur on real data.} 
	(a) is the input LSR-HFR frame with camera noise.
	(b) is the input HSR-LFR frame with motion blur.
	(c) is our reconstructed frame in LSR-HFR view.
	(d) is our reconstructed frame in HSR-LFR view.
	}
	\label{fig:dual_focal_2_blur}
	\vspace{-15pt}
\end{figure}  
    
    \subsection{Camera Desynchronization}

    We analyze the effect of camera desynchronization by deliberately desynchronizing the LSR-HFR video from 0 to 3 frames.
    The testing HSR-LFR (960p) and LSR-HFR (240p) videos have a $10\times$ frame rate gap. 
    As shown in Table~\ref{table:desyn}, the performances of both views degrade as the increase of desynchronization frame, especially for HSR-LFR view.
    The reason is that the dynamic filter in the adaptive weighting fusion network can slightly rectify the misalignment for LSR-HFR view introduced by desynchronization.
    However, the desynchronization directly leads to the misalignment of $R^t_3, R^t_4, R^t_5, R^t_6$ without any rectification.
    From Table~\ref{table:right_ablation}, disparities and flows from LSR-HFR view are critical for the reconstruction of HSR-LFR view. Thus, disparity error introduced by desynchronization will cause serious degradation on HSR-LSR view.
    Nevertheless, this is not a major challenge as the two cameras can be synchronized to trigger simultaneously on typical dual-camera devices.
    
    \subsection{Long/short Exposures}
    Our H$^2$-Stereo system can not only share the spatiotemporal information across views, but also solve the problems of noise and motion blur in real scenes through long/short exposures.
    The high frame rate of LSR-HFR camera suppresses its exposure time, which will lead to noisy frames, especially under low light conditions.
    We can choose an appropriate exposure time for HSR-LFR camera based on auto-exposure algorithms for better image quality.
    The longer exposure time of HSR-LFR camera may lead to motion-blurred frames.
    Motion blurs and noises deteriorate disparity and flow estimations and the qualities of the outputs.
    Thus, we pre-remove noises and motion blurs using video denoising and deblurring methods, fastDVD~\cite{Tassano_2020_CVPR} and CDVD~\cite{Pan_2020_CVPR}.
    Then, we input the pre-processed frames into LIFnet for H$^2$-Stereo video reconstruction.
    Here, we analyze the robustness of our LIFnet to noises and blurs and our performance in challenging motion blur scenes.

    \subsubsection{Robustness to noises and blurs}
    In order to verify the robustness of our method to noises and blurs, we use H$^2$-Stereo dataset to evaluate the effect of the noise and motion blur.
    We add Gaussian noise with different standard deviations ($\sigma=$ 5, 10, 15 or 20) on the input LSR-HFR frames as camera noise.
    We simulate motion blur by averaging $\tau$ (set $\tau$ as 3, 5 or 7) high-frame-rate consecutive frames on HSR-LFR view and sample one frame from each eight frames as input HSR-LFR frame.
    Before input the noised and blurred frames into LIFnet, we use FastDVD~\cite{Tassano_2020_CVPR} and CDVD~\cite{Pan_2020_CVPR} for pre-denoising and pre-deblurring, which significantly improve the reconstruction of both views.
    We fine-tune LIFnet with H$^2$-Stereo dataset with different noise and blur levels.
    As shown in Table~\ref{table:longshort}, the added noises and blurs degrade the performance of both views.
    However, comparing with the result of frame interpolation method DAIN~\cite{bao2019depth} with sharp HSR-LFR input (36.33dB in Table~\ref{table:stereoresults}), our method still achieves 1.46dB gain with $\sigma=5, \tau=3$ and 0.54dB gain with $\sigma=10, \tau=3$.
    However, when the noise on LSR-HFR frames is too strong ($\sigma>=15$), LSR-HFR frames will not help or even damage the reconstruction of HSR-LFR view due to the disparity estimation is failed.
    We also test our model on real long-/short- exposed HSR-LFR/LSR-HFR frames, as shown in Figure~\ref{fig:dual_focal_2_blur}.
    The input LSR-HFR frame is noisy due to the short exposure time of LSR-HFR camera. The input HSR-LFR frame is heavily blurry due to the long exposure time of HSR-LFR camera and the fast moving of the hands. Our LIFnet can remove the heavy motion blur with the help of the blur-free LSR-HFR frame. At the same time, the noises on the input LSR-HFR frame are removed on the results with the help of the HSR-LFR frame.
    Although the noise and blur model in real data is complex, our model trained with Gaussian noise and averaging-based motion blur has visually pleasant results.

    \subsubsection{Challenging motion blur scenes}
    We analyze the high-effectiveness of our method in challenging motion blur scenes, such as severe camera shake and fast-moving of objects.

    We evaluate the joint deblurring and frame interpolation performances of HSR-LFR view with Stereo Blur dataset~\textit{\etal}~\cite{zhou2019davanet}, which dominant motion is camera shake.
    We use blurred right view videos with $\frac{1}{2}$ frame rate as HSR-LFR inputs and down-scaled the sharp left view videos by $4\times$ as LSR-HFR frames.
    We pre-deblur the blurred right view with CDVD~\cite{Pan_2020_CVPR} and DAVA~\cite{zhou2019davanet}, as the input of the following frame interpolation procedure.
    As shown in Table~\ref{table:blur_FI}, our method brings 5.62dB performance gain against frame interpolation methods.
    As shown in Figure~\ref{fig:blur_FI}, although the state-of-the-art deblurring methods can not remove the heavy blur of the logo "Kempinski", our LIFnet can reconstruct the sharp "Kempinski" with the help of the sharp LSR-HFR frame.

    \begin{table}[t]
% \vspace{-10pt}
\caption{\textbf{Quantitative evaluations on stereo blur dataset~\cite{zhou2019davanet}} 
}
\label{table:blur_FI}
\centering
\footnotesize
      \newcommand{\quantTit}[1]{\multicolumn{3}{c}{\scriptsize #1}}
    \newcommand{\quantSec}[1]{\scriptsize #1}
    \newcommand{\quantInd}[1]{\scriptsize #1}
    \newcommand{\quantVal}[1]{\scalebox{0.83}[1.0]{$ #1 $}}
    \newcommand{\quantBes}[1]{\scalebox{0.83}[1.0]{$\uline{ #1 }$}}
\renewcommand{\tabcolsep}{10pt} % adjust horizontal space
\renewcommand{\arraystretch}{0.9} % adjust vertical space
\vspace{-5pt}
\begin{tabular}{ l c c}
\toprule
\multirow{2}{*}{Methods} & \multicolumn{2}{c}{HSR-LFR} \\
\cmidrule{2-3}
& PSNR [dB] & SSIM\\
\midrule
    CDVD~\cite{Pan_2020_CVPR} + DAIN~\cite{bao2019depth}
        & 22.31
        & 0.7281
        \\
    CDVD~\cite{Pan_2020_CVPR} + BMBC~\cite{park2020bmbc}
        & 20.85
        & 0.6990
        \\
    \midrule
    DAVA~\cite{zhou2019davanet} + DAIN~\cite{bao2019depth}
        & \underline{22.36}
        & \underline{0.7304}
        \\
    DAVA~\cite{zhou2019davanet} + BMBC~\cite{park2020bmbc}
        & 20.87
        & 0.7003
        \\
    \midrule
    CDVD~\cite{Pan_2020_CVPR} + LIFnet (Ours)
        & \textbf{27.98}
        & \textbf{0.8941}
        \\
\bottomrule
\end{tabular}
\vspace{-10pt}
\end{table}
    \begin{figure}[t]
% \vspace{-20pt}
	\footnotesize
% 	\tiny 
% 	\scriptsize        
	\centering
	\renewcommand{\tabcolsep}{1pt} % adjust horizontal space
	\renewcommand{\arraystretch}{1} % adjust vertical space
      \newcommand{\quantTit}[1]{\multicolumn{3}{c}{\scriptsize #1}}
    \newcommand{\quantSec}[1]{\scriptsize #1}
    \newcommand{\quantInd}[1]{\scriptsize #1}
    \newcommand{\quantVal}[1]{\scalebox{0.83}[1.0]{$ #1 $}}
    \newcommand{\quantBes}[1]{\scalebox{0.83}[1.0]{$\uline{ #1 }$}}
	\begin{tabular}{ccc}
    \includegraphics[width=0.14\textwidth]{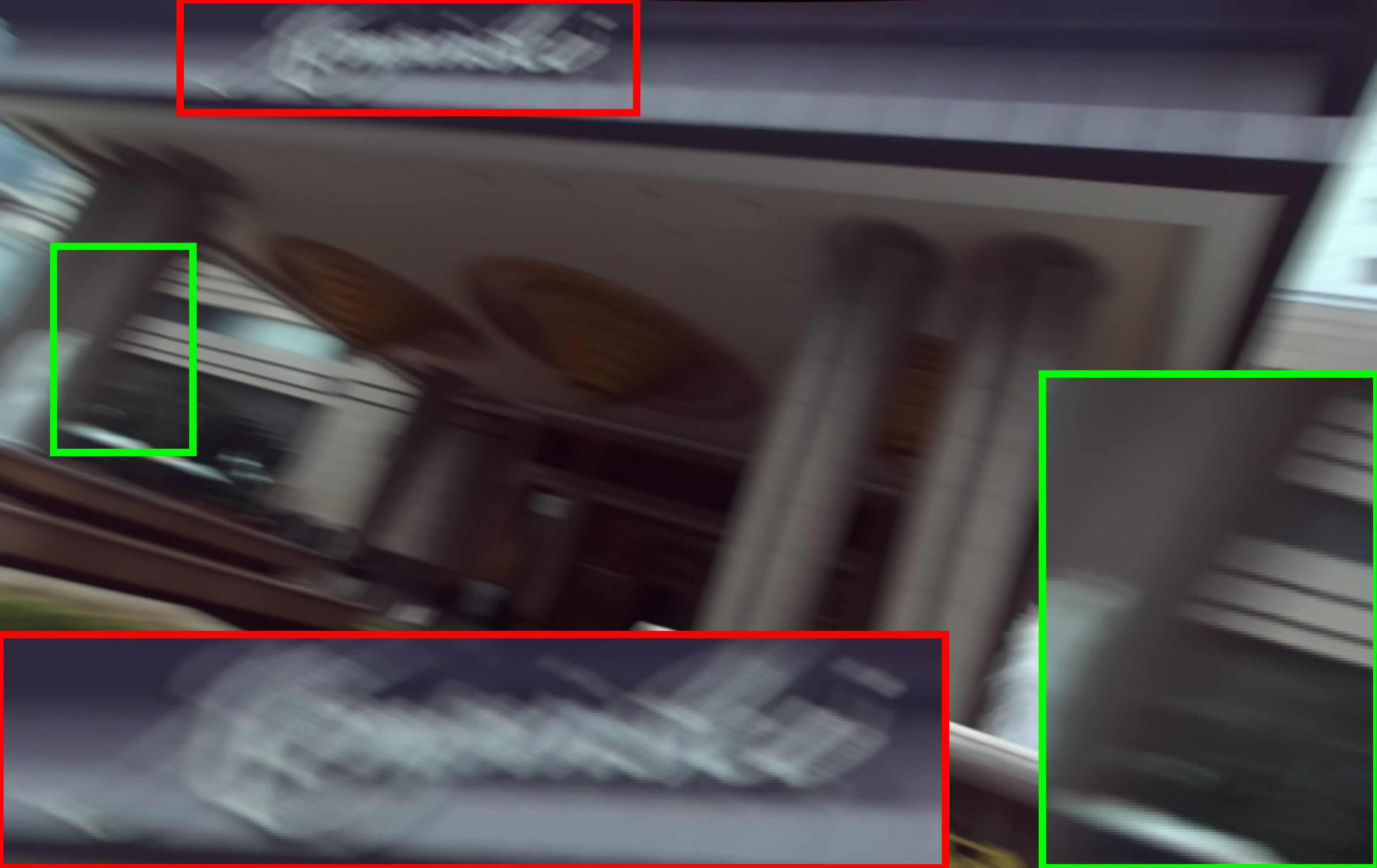} & \includegraphics[width=0.14\textwidth]{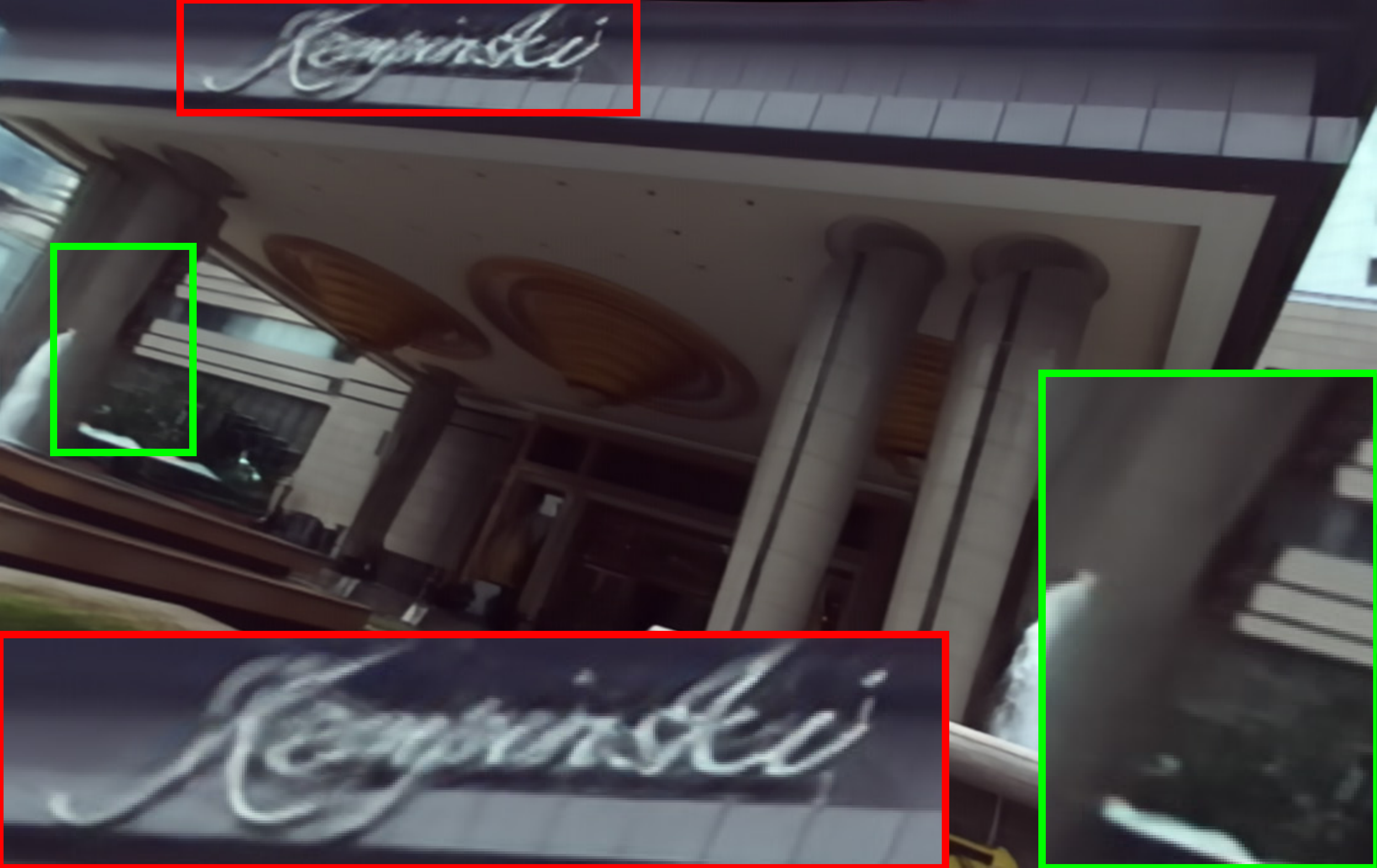} &
    \includegraphics[width=0.14\textwidth]{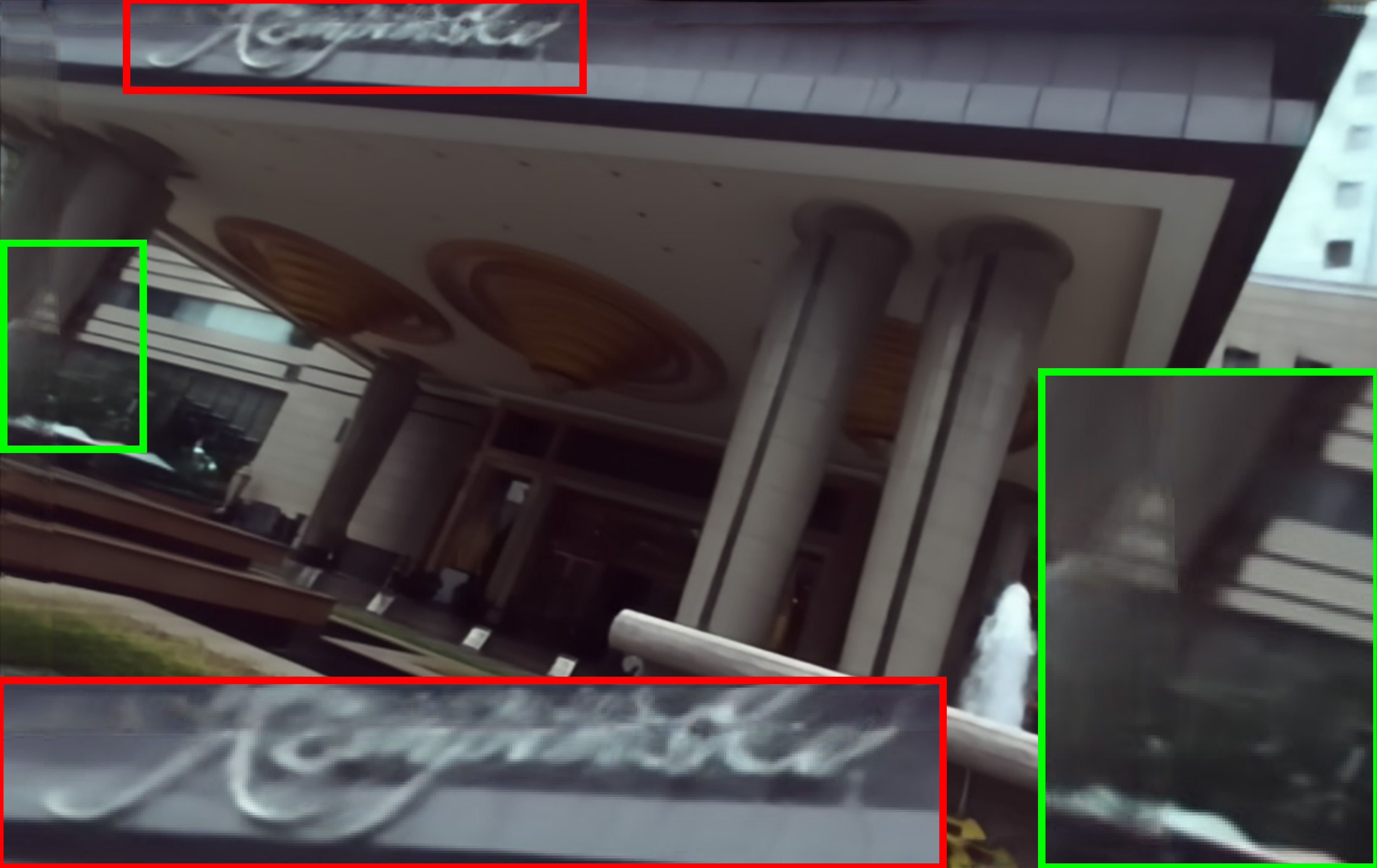} \\
    (a) Blurred HSR-LFR & (b) DAVA~\cite{zhou2019davanet} & (c) \scriptsize{DAVA~\cite{zhou2019davanet}+DAIN~\cite{bao2019depth}} \\
    \includegraphics[width=0.14\textwidth]{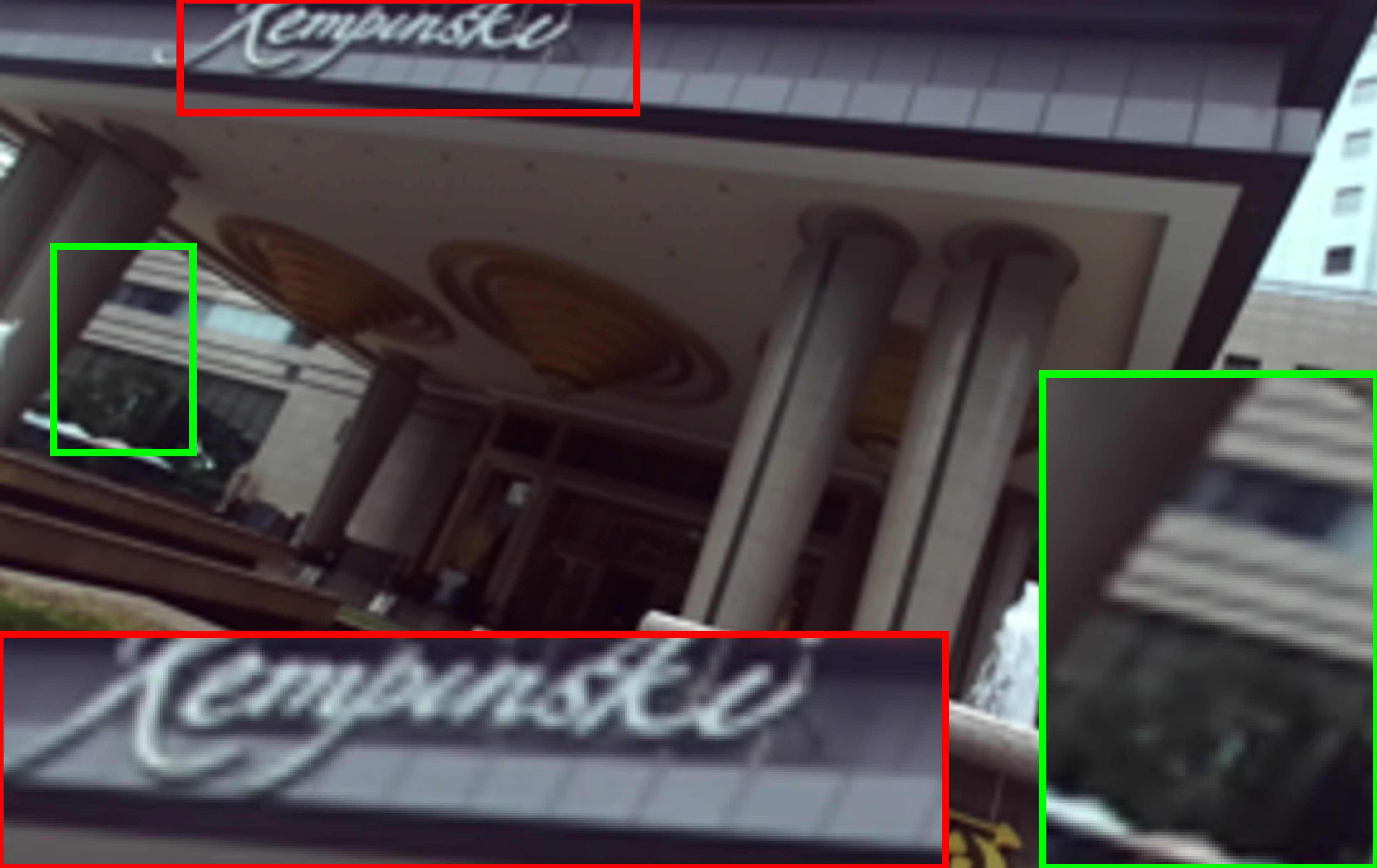} &
    \includegraphics[width=0.14\textwidth]{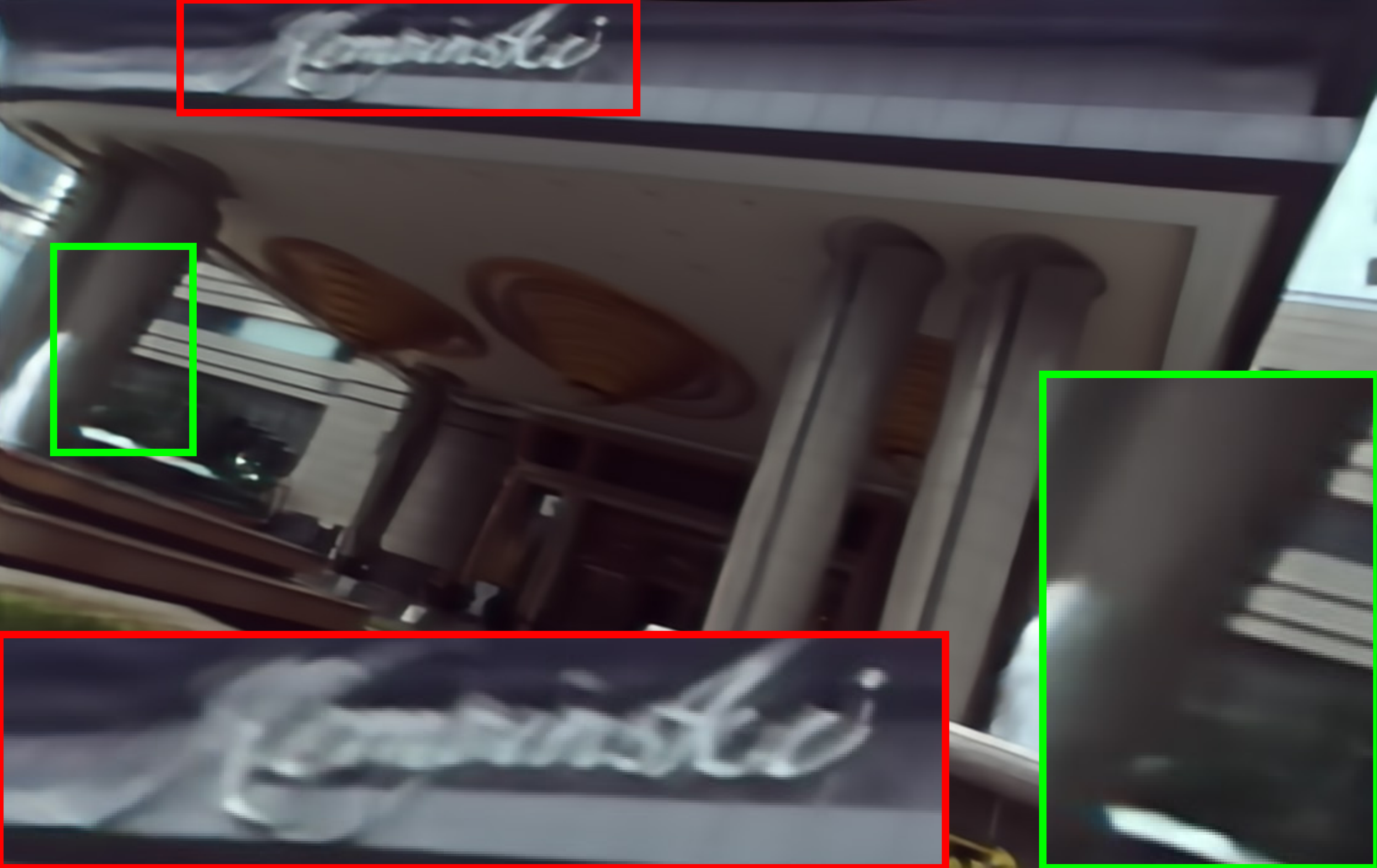} &
    \includegraphics[width=0.14\textwidth]{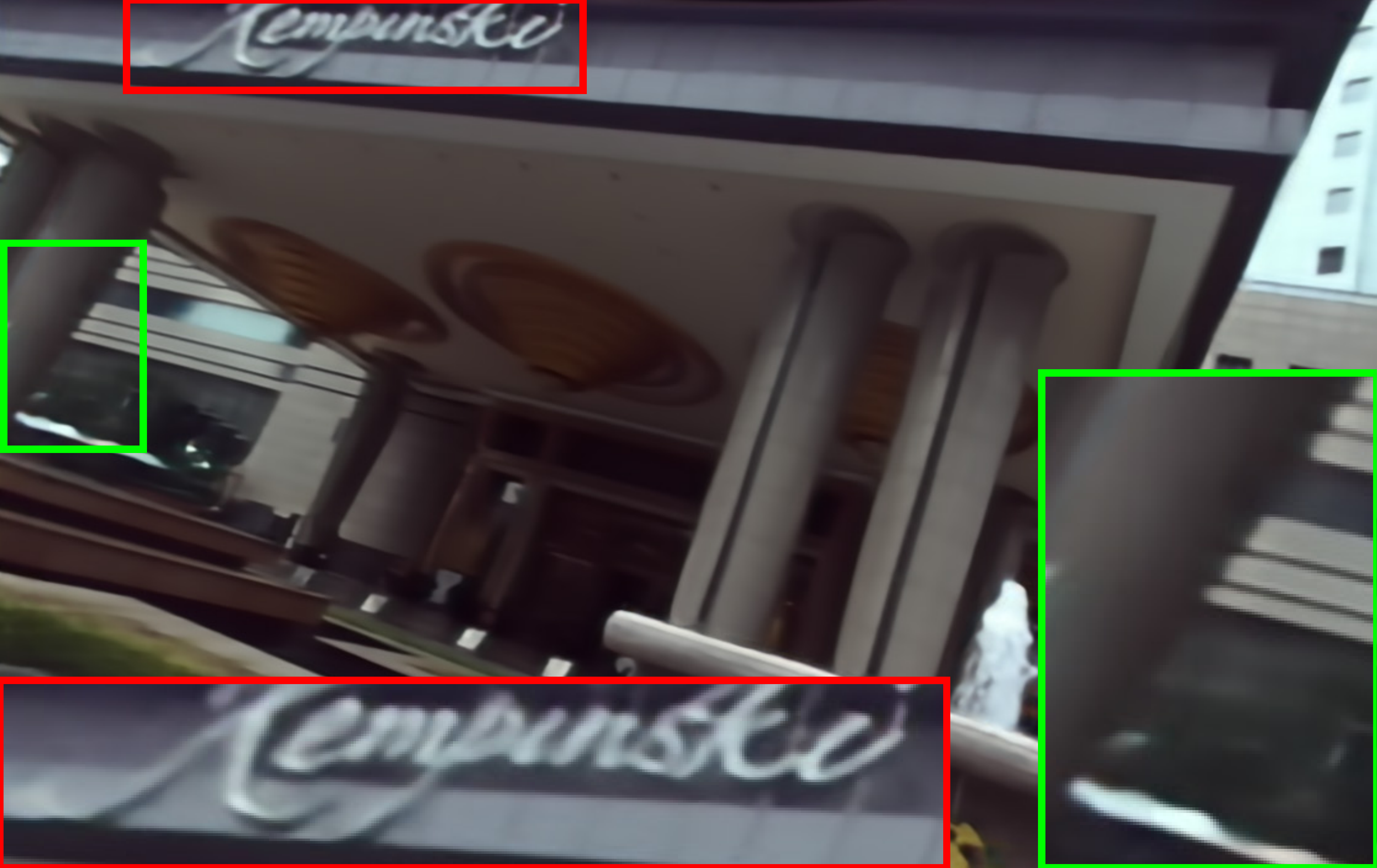} \\
    (d) LSR-HFR & (e) CDVD~\cite{Pan_2020_CVPR} & (f) \scriptsize{CDVD~\cite{Pan_2020_CVPR}+LIFnet}\\
	\end{tabular}
    % \vspace{-5pt}
	\caption{\textbf{Visual comparisons with blurred HSR-LFR inputs.} 
	}
	\label{fig:blur_FI}
	\vspace{-15pt}
\end{figure}
    \begin{figure}[t]
% \vspace{-20pt}
	\footnotesize
% 	\tiny 
% 	\scriptsize        
	\centering
	\renewcommand{\tabcolsep}{0.8pt} % adjust horizontal space
	\renewcommand{\arraystretch}{1} % adjust vertical space
      \newcommand{\quantTit}[1]{\multicolumn{3}{c}{\scriptsize #1}}
    \newcommand{\quantSec}[1]{\scriptsize #1}
    \newcommand{\quantInd}[1]{\scriptsize #1}
    \newcommand{\quantVal}[1]{\scalebox{0.83}[1.0]{$ #1 $}}
    \newcommand{\quantBes}[1]{\scalebox{0.83}[1.0]{$\uline{ #1 }$}}
	\begin{tabular}{cccccc}
	\includegraphics[width=0.079\textwidth]{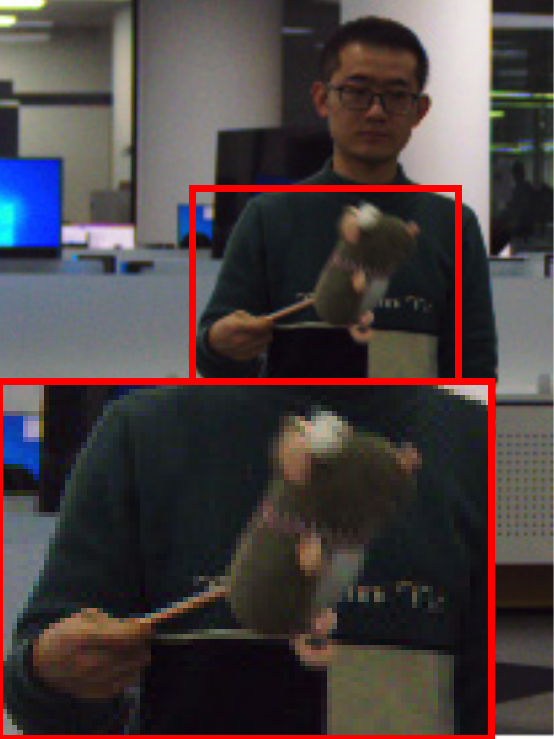} &
	\includegraphics[width=0.079\textwidth]{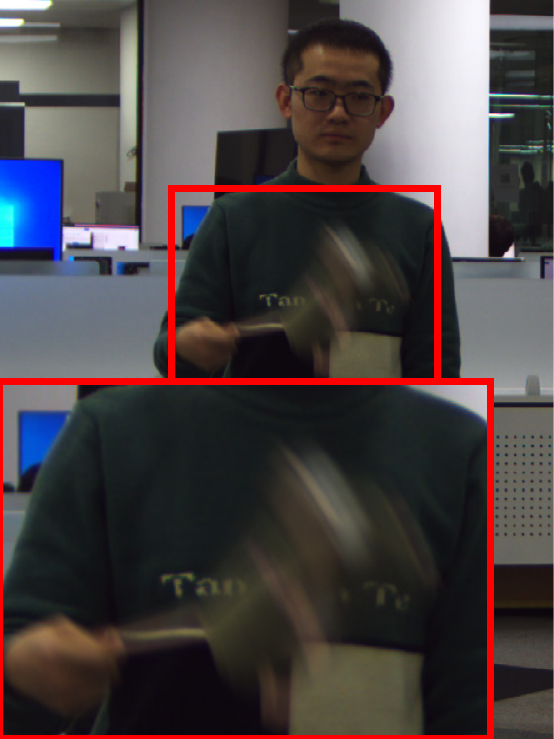} &
    \includegraphics[width=0.079\textwidth]{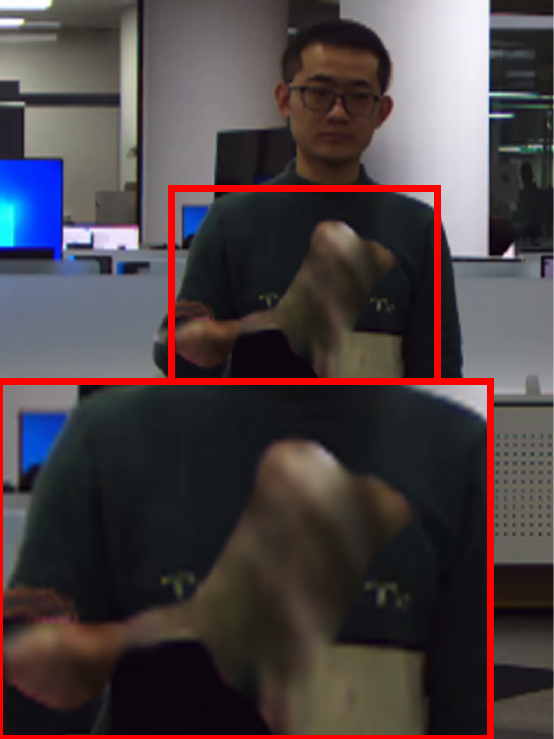} &
    \includegraphics[width=0.079\textwidth]{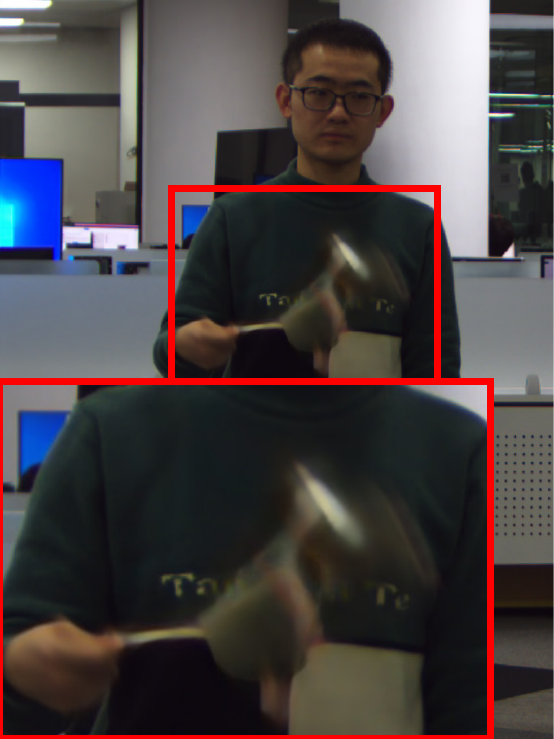} &
	\includegraphics[width=0.079\textwidth]{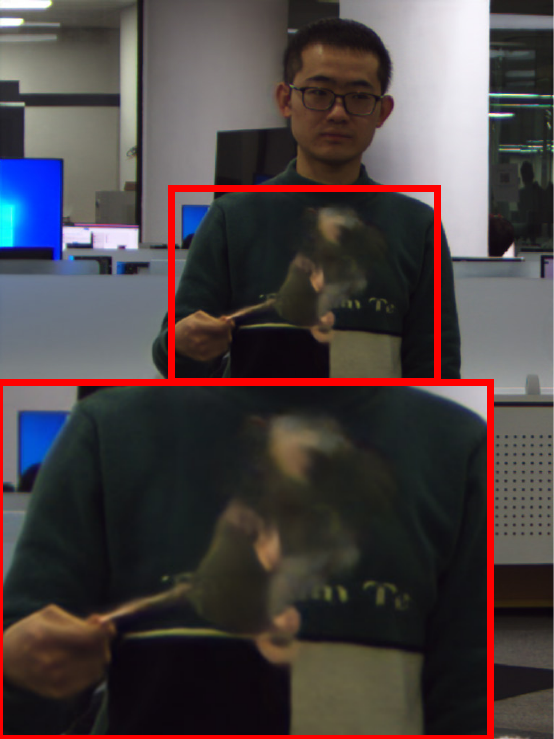} &
	\includegraphics[width=0.079\textwidth]{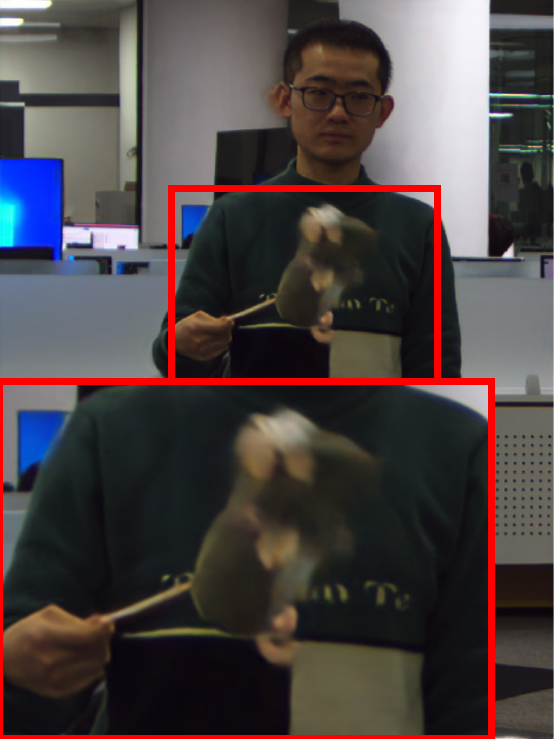} \\
    (a)  & (b) & (c) & (d) & (e) & (f) \\
	\end{tabular}
    \vspace{-8pt}
	\caption{\textbf{Visual comparisons on blurred fast-moving object.} (a) is the short-exposed LSR-HFR input. (b) is the long-exposed HSR-LFR input. (c) is the deblurred moving object with DeFMO~\cite{rozumnyi2020defmo}. (d) is the deblurred HSR-LFR frame with CDVD~\cite{Pan_2020_CVPR}. (e) and (f) are the reconstructed HSR-LFR frames without (\textit{-w/o}) and with (\textit{-w}) pre-deblurring method CDVD~\cite{Pan_2020_CVPR}, respectively. 
	}
	\label{fig:blurobject}
	\vspace{-10pt}
\end{figure}
    \begin{figure}[h!]
% \vspace{-10pt}
	\footnotesize
% 	\tiny 
% 	\scriptsize        
	\centering
% 	\vspace{-10pt}
	\renewcommand{\tabcolsep}{1.0pt} % adjust horizontal space
	\renewcommand{\arraystretch}{1.0} % adjust vertical space
    
	\begin{tabular}{cccc}
	\multicolumn{2}{c}{LSR-HFR} & \multicolumn{2}{c}{HSR-LFR} \\
	\includegraphics[width=0.22\linewidth]{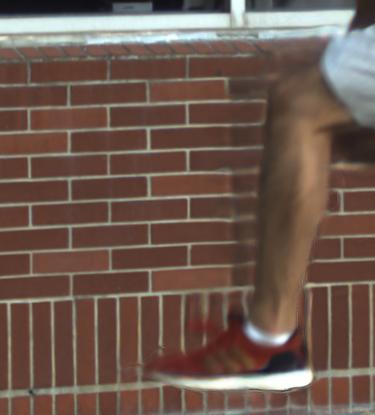} &
	\includegraphics[width=0.22\linewidth]{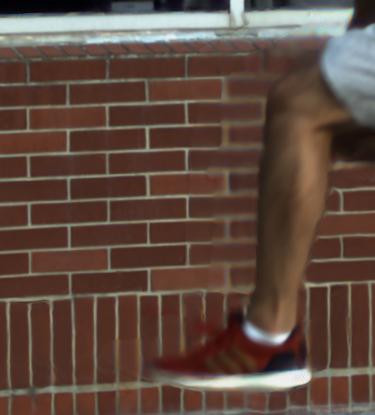} &
	\includegraphics[width=0.22\linewidth]{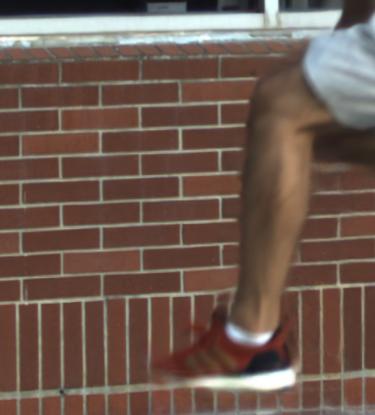} &
	\includegraphics[width=0.22\linewidth]{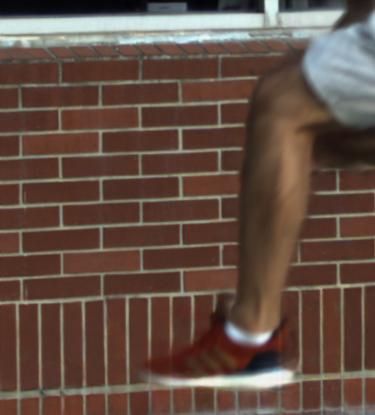}\\
	(a) LIFnet-HD$^3$S & (b) LIFnet-LEAS & (c) LIFnet-HD$^3$S & (d) LIFnet-LEAS \\
	\end{tabular}
	\caption{\textbf{Different disparity networks.} (a) and (c) are the results using HD$^3$S~\cite{yin2019hierarchical} for disparity estimation. (b) and (d) are using LEAStereo~\cite{cheng2020hierarchical}. 
    }
	\label{fig:lea}
	\vspace{-15pt}
\end{figure}

    Here, we analyze our effectiveness on deblurring of fast-moving objects, which is a challenging problem.
    Most of the existing image/video deblurring methods consider only low to medium blur, emerging from global camera blur due to camera motion or objects moving at moderate speed.
    They always fail to reconstruct sharp textures of fast-moving objects, as shown in Figure~\ref{fig:blurobject}(e).
    In recent years, some researchers try to reconstruct sharp fast-moving object~\cite{rozumnyi2020defmo} by tracking and appearance estimation.
    However, they will fail when the shape of the object is complex, and they can not preserve the fine-grained textures of the object, as shown in Figure~\ref{fig:blurobject}(c).
    With the sharp LSR-HFR frame as reference (Figure~\ref{fig:blurobject}(a)), our method can reconstruct the good appearance and accurate motion of the object as shown in Figure~\ref{fig:blurobject}(e).
    From Figure~\ref{fig:blurobject}(f), the pre-deblurring process can further improve the reconstruction of the fast-moving object. 
    Because the pre-deblurring process can slightly sharpen the blurred object, which alleviates the difficulties of disparity and flow estimations.

    \begin{table}[t]
% \vspace{-5pt}
\caption{\textbf{Running times on image of size $640\times 512$.}}
\label{table:hd3slea}
\centering
\footnotesize
\newcommand{\quantVal}[1]{\scalebox{1.0}[1.0]{#1}}
\newcommand{\quantInd}[1]{\scalebox{1.0}[1.0]{\scriptsize #1}}
\renewcommand{\tabcolsep}{5pt} % adjust horizontal space
\renewcommand{\arraystretch}{0.9} % adjust vertical space
\vspace{-5pt}
\begin{tabular}{ccc}
\toprule
{Methods} & Running time (ms) & Inference memory (MB) \\
\midrule
HD$^3$-Stereo~\cite{yin2019hierarchical}  & 82.8  & 1833 \\
LEAStereo~\cite{cheng2020hierarchical}  & 271.2  & 4784 \\
\bottomrule
\end{tabular}
\vspace{-10pt}
\end{table} 
    \begin{figure}[t]
	\footnotesize
% 	\tiny 
% 	\scriptsize        
	\centering
	\renewcommand{\tabcolsep}{1pt} % adjust horizontal space
	\renewcommand{\arraystretch}{1} % adjust vertical space
      \newcommand{\quantTit}[1]{\multicolumn{3}{c}{\scriptsize #1}}
    \newcommand{\quantSec}[1]{\scriptsize #1}
    \newcommand{\quantInd}[1]{\scriptsize #1}
    \newcommand{\quantVal}[1]{\scalebox{0.83}[1.0]{$ #1 $}}
    \newcommand{\quantBes}[1]{\scalebox{0.83}[1.0]{$\uline{ #1 }$}}
	\begin{tabular}{cc}
	\includegraphics[width=0.15\textwidth]{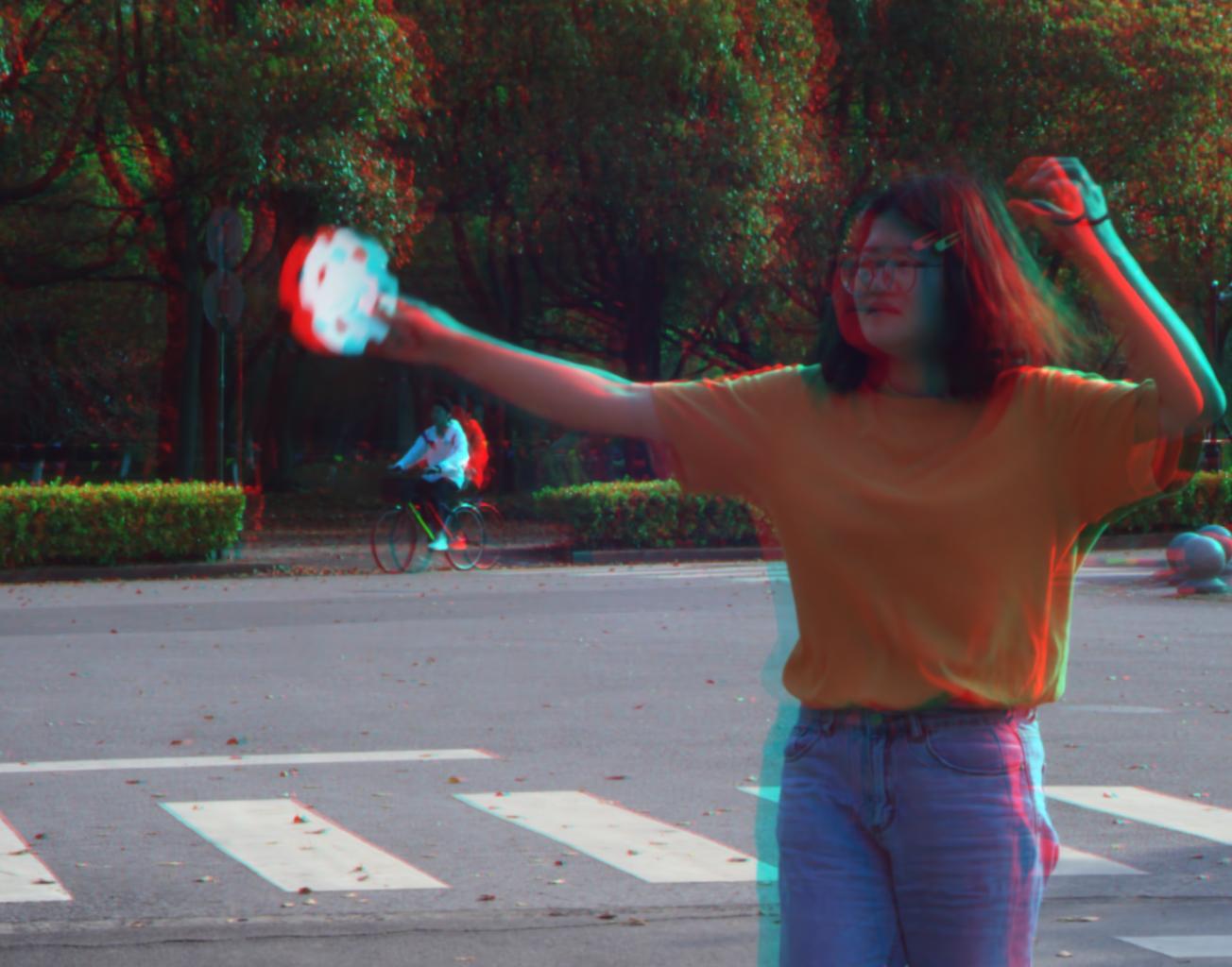} &
	\includegraphics[width=0.15\textwidth]{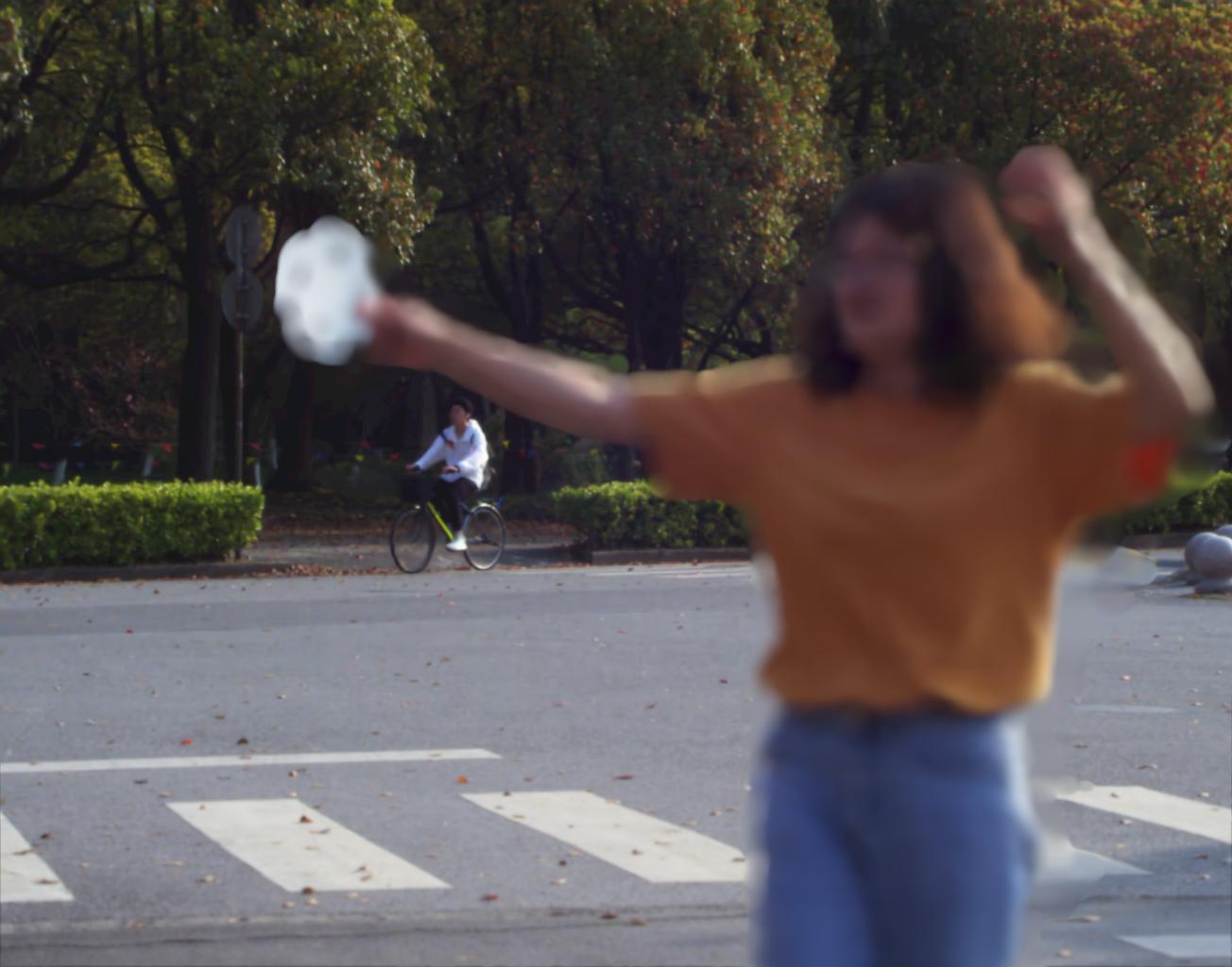} \\
	(a) stereoscopic frame & (b) Background Focus \\
	\end{tabular}
    % \vspace{-5pt}
	\caption{\textbf{Applications.} (a) Reconstructed stereoscopic frame. You can see the stereoscopic effect with a pair of anaglyph 3D glasses (red and cyan). (b) Refocused frame focused on the background.}
	\label{fig:stereoeffect}
	\vspace{-10pt}
\end{figure}
  	\begin{table}[t]
% \vspace{-10pt}
\caption{\textbf{Quantitative evaluations on disparity estimation.}}
\label{table:disps}
\centering
\renewcommand{\tabcolsep}{10pt} 
\renewcommand{\arraystretch}{0.9} % adjust vertical space
\vspace{-5pt}

\begin{tabular}{l | c c c c c}
\toprule
methods & PSNR [dB]$\uparrow$ & NEPE$\downarrow$ \\
\midrule
DAIN~\cite{bao2019depth} & 33.08 & 31.77 \\
RBPN~\cite{haris2019recurrent} & 34.60 & 33.21 \\
PASSRnet~\cite{wang2019learning} & \underline{35.47} & \underline{29.31} \\
LIFnet(Ours) & \textbf{36.55} & \textbf{20.88} \\
\bottomrule
\end{tabular}
\vspace{-10pt}
\end{table}
    \begin{figure}[h!]
% \vspace{-20pt}
	\footnotesize
% 	\tiny 
% 	\scriptsize        
	\centering
	\renewcommand{\tabcolsep}{1pt} % adjust horizontal space
	\renewcommand{\arraystretch}{1} % adjust vertical space
      \newcommand{\quantTit}[1]{\multicolumn{3}{c}{\scriptsize #1}}
    \newcommand{\quantSec}[1]{\scriptsize #1}
    \newcommand{\quantInd}[1]{\scriptsize #1}
    \newcommand{\quantVal}[1]{\scalebox{0.83}[1.0]{$ #1 $}}
    \newcommand{\quantBes}[1]{\scalebox{0.83}[1.0]{$\uline{ #1 }$}}
	\begin{tabular}{cc}
	\includegraphics[width=0.19\textwidth]{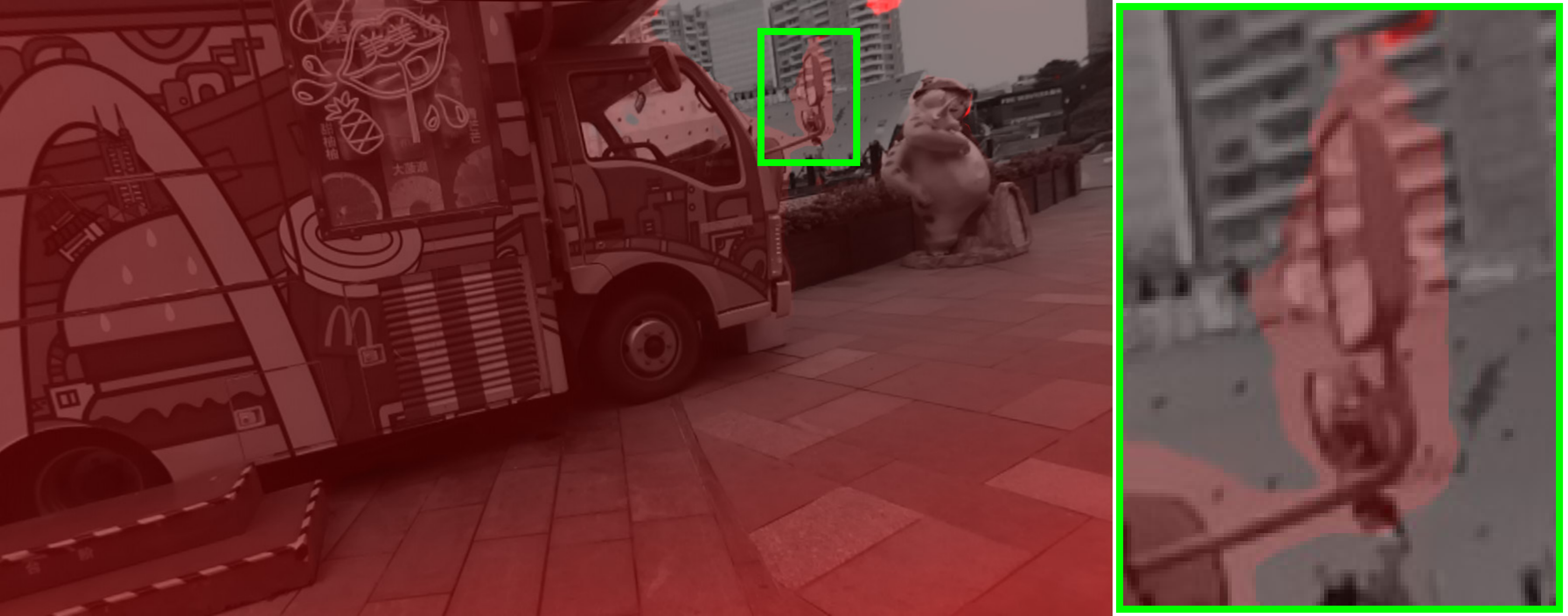} &
	\includegraphics[width=0.19\textwidth]{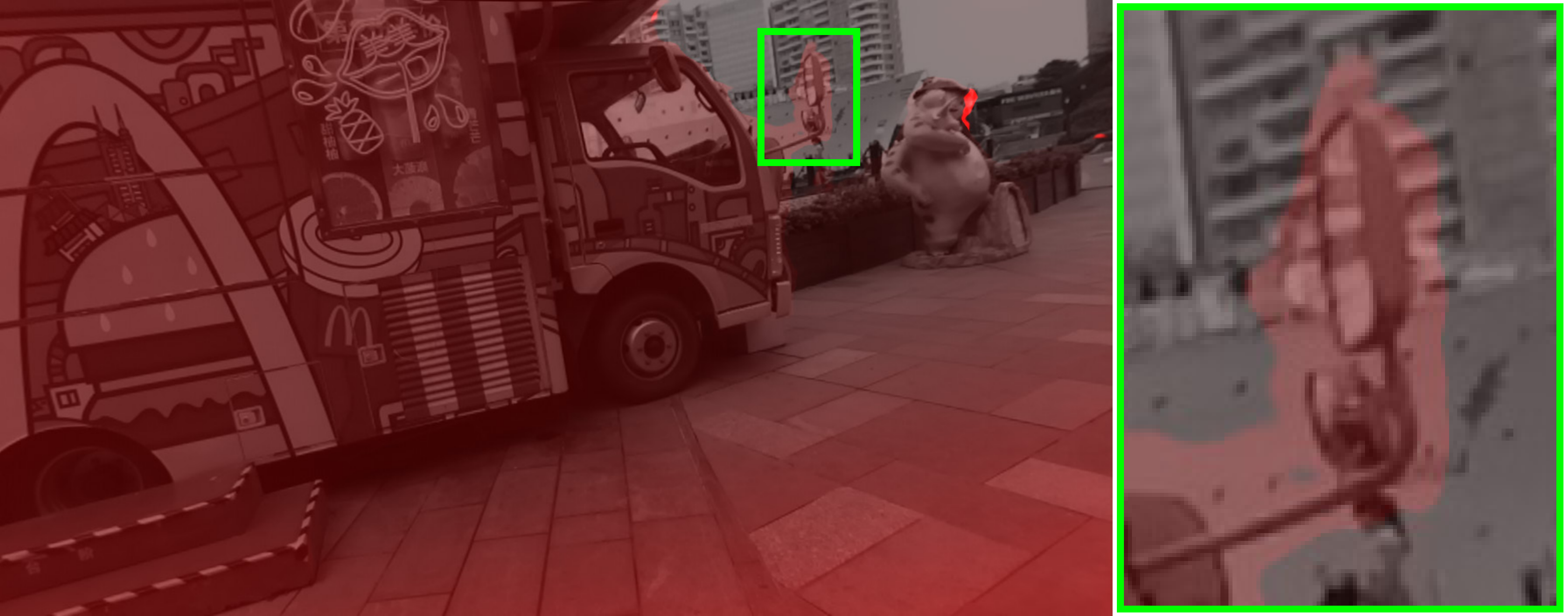} \\
	(a) PASSRnet~\cite{wang2019learning} (38.10 dB) & (b) RBPN~\cite{haris2019recurrent} (38.67 dB)\\
	\includegraphics[width=0.19\textwidth]{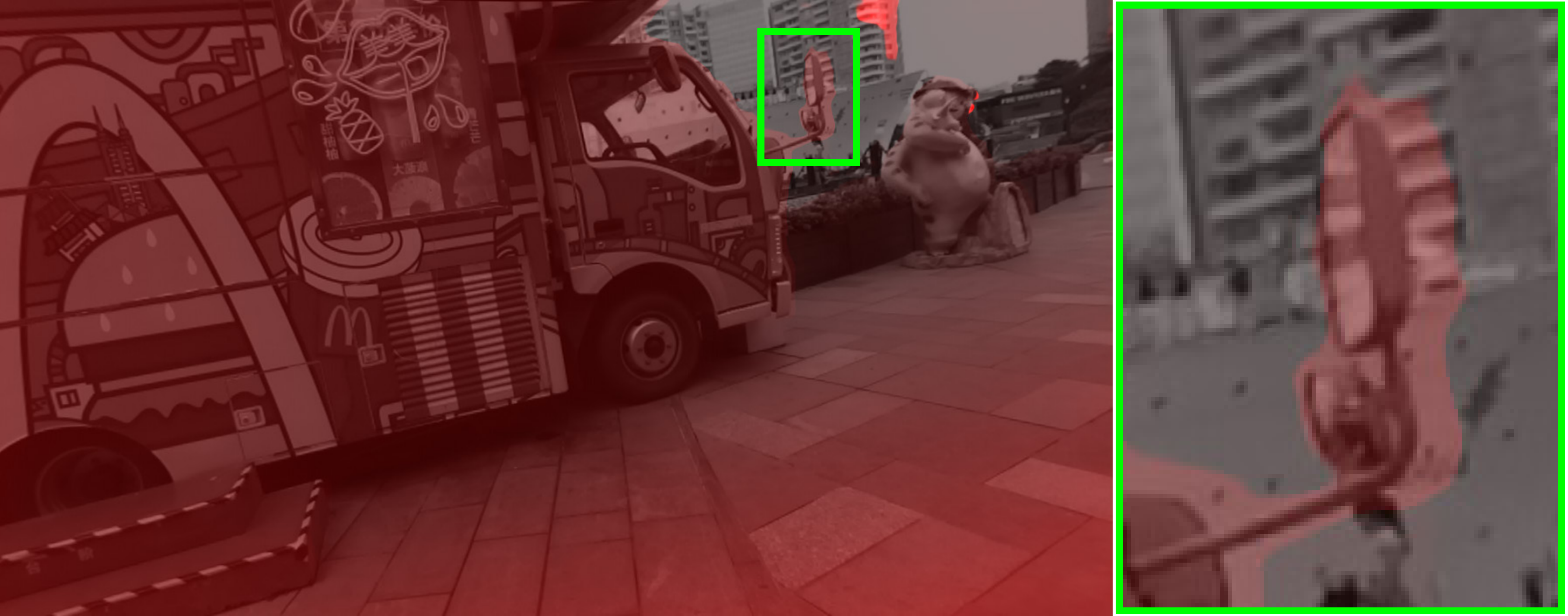} & 
	\includegraphics[width=0.19\textwidth]{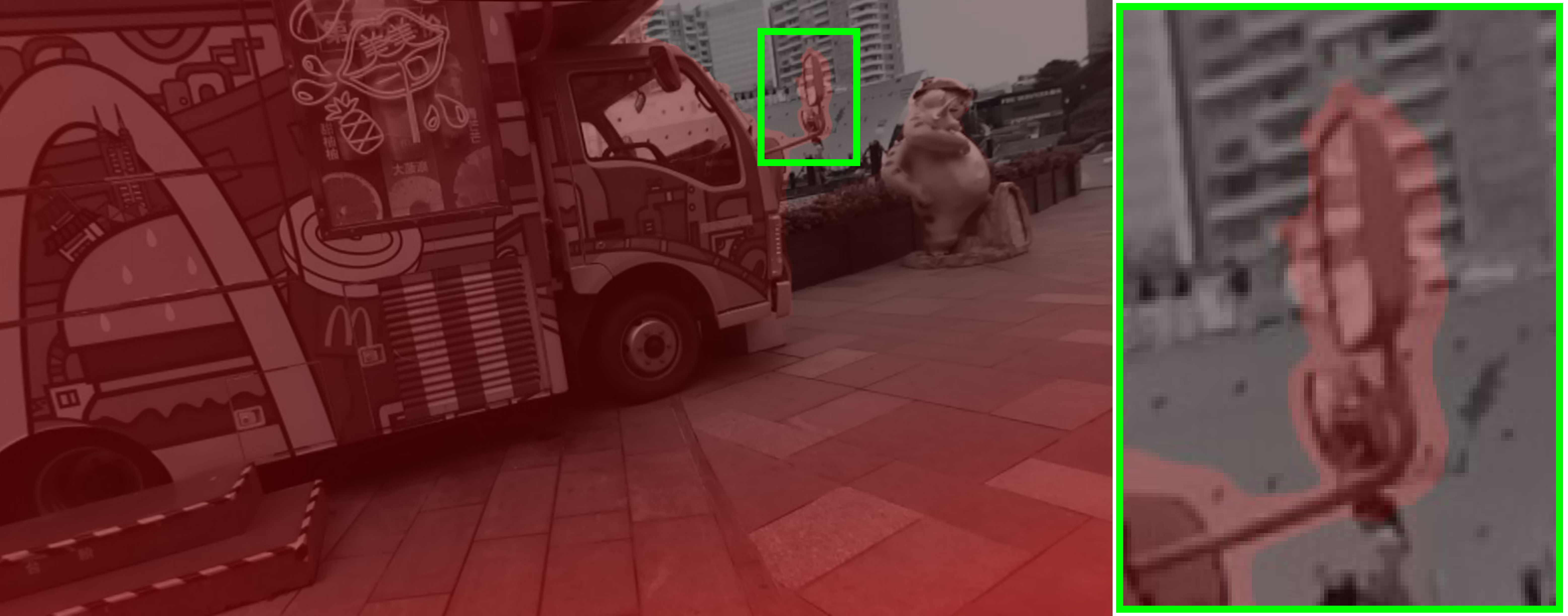} \\
	(c) DAIN~\cite{bao2019depth} (36.65 dB)& (d) Ours (47.57 dB)\\
	\end{tabular}
    \vspace{-5pt}
	\caption{\textbf{Visual comparisons on disparity.} 
	We overlap the disparities and the LSR-HFR frame for visually clearer comparison. The numbers in the brackets are the PSNR of the disparities compared with the ground truth disparity. 
	}
	\label{fig:disp2}
	\vspace{-15pt}
\end{figure}

    \subsection{Effect of Different Disparity Network}
    
    We replace the disparity network HD$^3$S~\cite{yin2019hierarchical} in our LIFnet (denoted as LIFnet-HD$^3$S) with the state-of-the-art disparity network LEAStereo~\cite{cheng2020hierarchical} (denoted as LIFnet-LEAS) for comparison.
    We implement the same training process for LIFnet-HD$^3$S and LIFnet-LEAS.
    We evaluate the performance of our LIFnet-HD$^3$S and LIFnet-LEAS in our captured challenging real-world scene with very large disparity.
  	From the results in Figure~\ref{fig:lea}, in most areas, LIFnet-HD$^3$S has the similar performance as LIFnet-LEAS.
  	However, the tiptoe of the athlete is blurred with LIFnet-HD$^3$S but is completely reconstructed by LIFnet-LEAS.
  	Thus, a larger and better disparity network can further improve the performance of our method.
  	We present the running times and inference memory consumptions using one RTX3090 GPU card of LEAStereo~\cite{cheng2020hierarchical} and HD$^3$S~\cite{yin2019hierarchical} in Table~\ref{table:hd3slea}. The input image size is $640\times 512$.
  	The running time of LEAStereo is 3.28 times of HD$^3$Stereo, and the inference memory consumption of LEAStereo is 2.61 times of HD$^3$Stereo.
  	In this paper, we choose HD$^3$S~\cite{yin2019hierarchical} to balance the accuracy and the computational complexity.
  	We can use LEAStereo in the cases where high reconstruction quality is required regardless of computational cost.

    \subsection{Applications}
    In addition to the stereoscopic effect (Figure~\ref{fig:stereoeffect}(a)), we can also estimate high spatiotemporal resolution disparities based on our reconstructed high-speed, high-resolution stereoscopic videos.
    We use GANet~\cite{Zhang2019GANet}, which is the state-of-the-art disparity estimation network, to estimate disparities on the reconstructed H$^2$-Stereo videos.
    The ground truth disparities are generated based on the H$^2$-Stereo ground truth video.
    We use two metrics to evaluate the disparity accuracy, PSNR and normalized end-point-error (NEPE).
    The definition of NEPE is as $NEPE = |d'-d_{gt}|/|d_{gt}+\xi|$,
    where $d'$ is the estimated disparity on the reconstructed stereoscopic video, $\xi$ is a small constant to prevent the denominator from being $0$ and $d_{gt}$ is the disparity of the ground truth video.
    The subjective and objective comparisons are shown in Table~\ref{table:disps} and Figure~\ref{fig:disp2}.
    The estimated disparity on our reconstructed frames outperforms all the other methods.
    See the rearview mirror of the truck, our LIFnet can reconstruct more precise disparity than PASSRnet~\cite{wang2019learning} in Figure~\ref{fig:disp2}(a) and RBPN~\cite{haris2019recurrent} in Figure~\ref{fig:disp2}(b). Due to the position offset of the interpolated frames by DAIN~\cite{bao2019depth}, the reconstructed disparity in Figure~\ref{fig:disp2}(c) is also misaligned with the ground truth LSR-HFR frame. However, our disparity is well aligned with the LSR-HFR frame.
    The estimated high-quality disparity is useful for downstream video applications, such as refocusing.
    We show the background refocused frame in Figure~\ref{fig:stereoeffect}(b).
  	
  	 \begin{figure}[t]
	\footnotesize
% 	\tiny 
% 	\scriptsize        
	\centering
	\renewcommand{\tabcolsep}{0.8pt} % adjust horizontal space
	\renewcommand{\arraystretch}{0.5} % adjust vertical space
      \newcommand{\quantTit}[1]{\multicolumn{3}{c}{\scriptsize #1}}
    \newcommand{\quantSec}[1]{\scriptsize #1}
    \newcommand{\quantInd}[1]{\scriptsize #1}
    \newcommand{\quantVal}[1]{\scalebox{0.83}[1.0]{$ #1 $}}
    \newcommand{\quantBes}[1]{\scalebox{0.83}[1.0]{$\uline{ #1 }$}}
	\begin{tabular}{cccc}
	 30 & 75 & 30 & 75\\
	\includegraphics[width=0.11\textwidth]{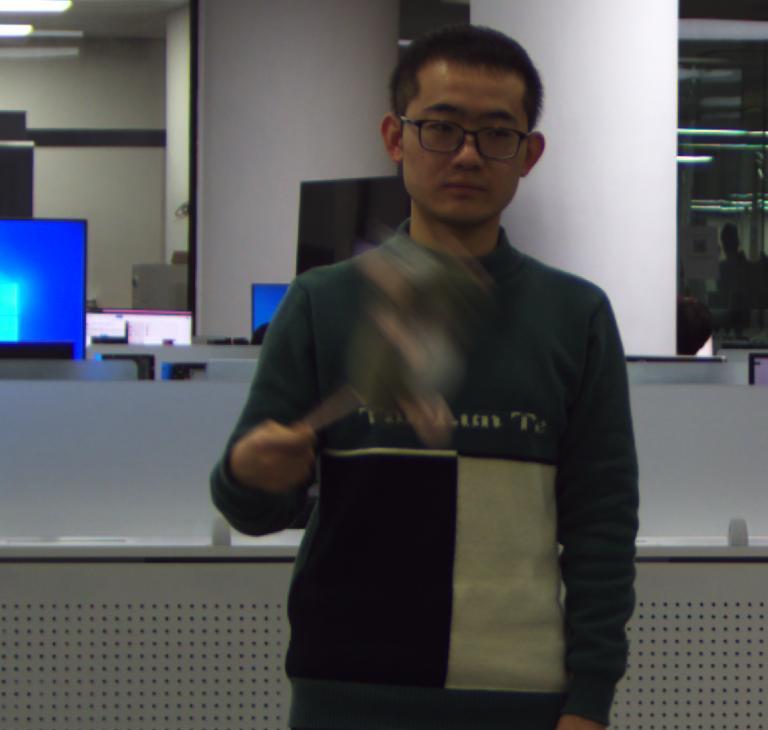} &
	\includegraphics[width=0.11\textwidth]{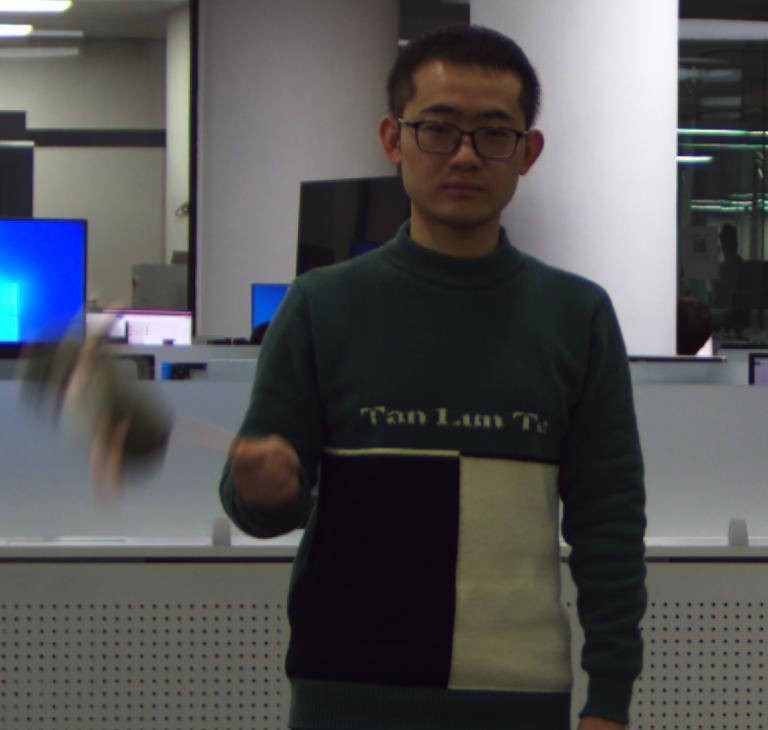} &
    \includegraphics[width=0.11\textwidth]{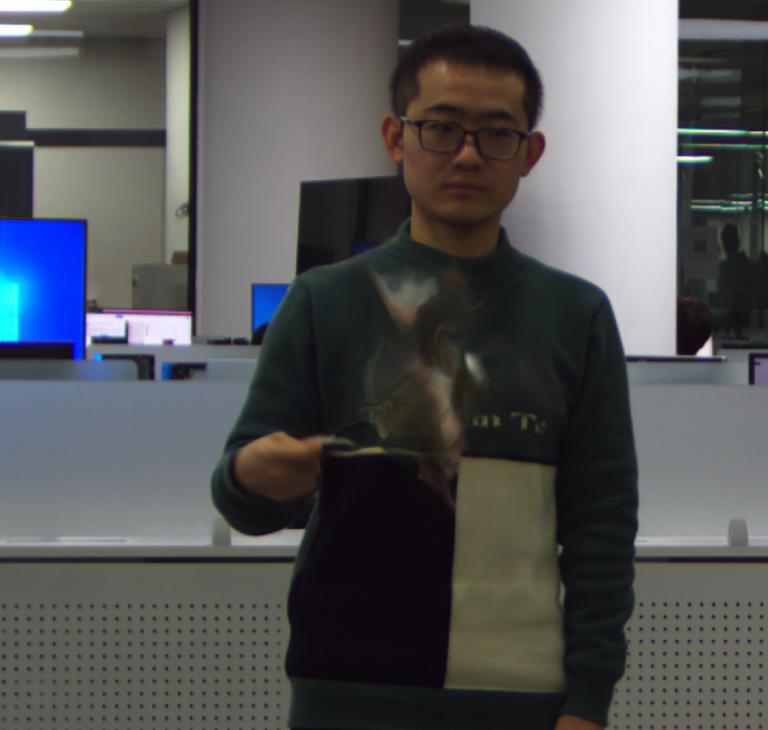} &
    \includegraphics[width=0.11\textwidth]{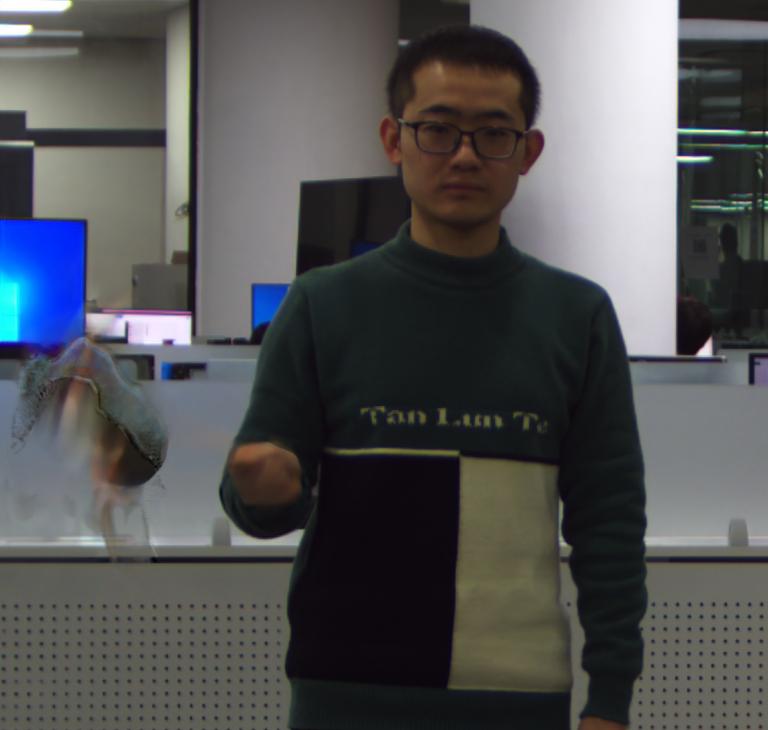} \\
    \multicolumn{2}{c}{{(a) HSR-LFR (Blurred input)}} & \multicolumn{2}{c}{{(b) HSR-LFR(CDVD+DAIN)}} \\
	\includegraphics[width=0.11\textwidth]{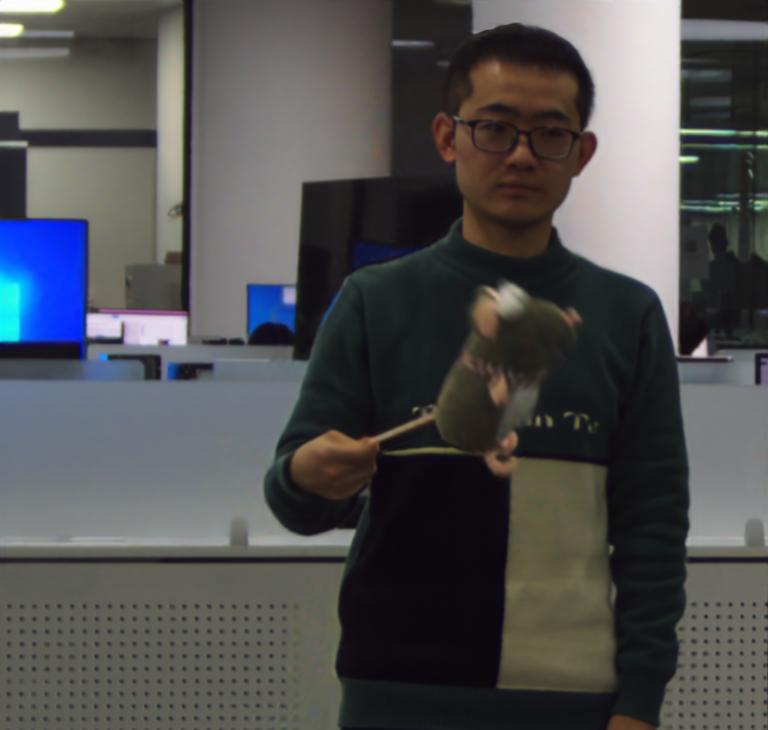} &
	\includegraphics[width=0.11\textwidth]{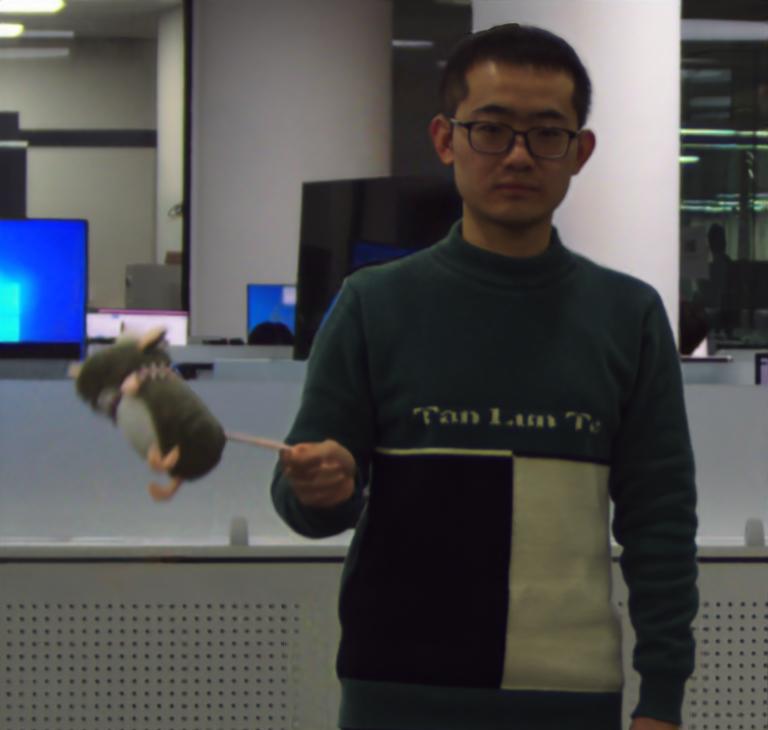} &
	\includegraphics[width=0.11\textwidth]{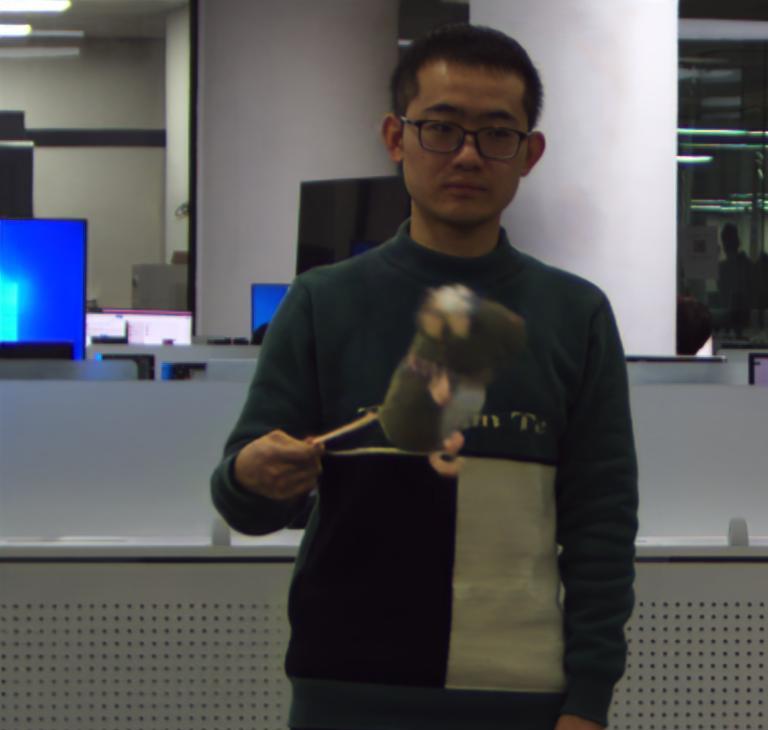} &
	\includegraphics[width=0.11\textwidth]{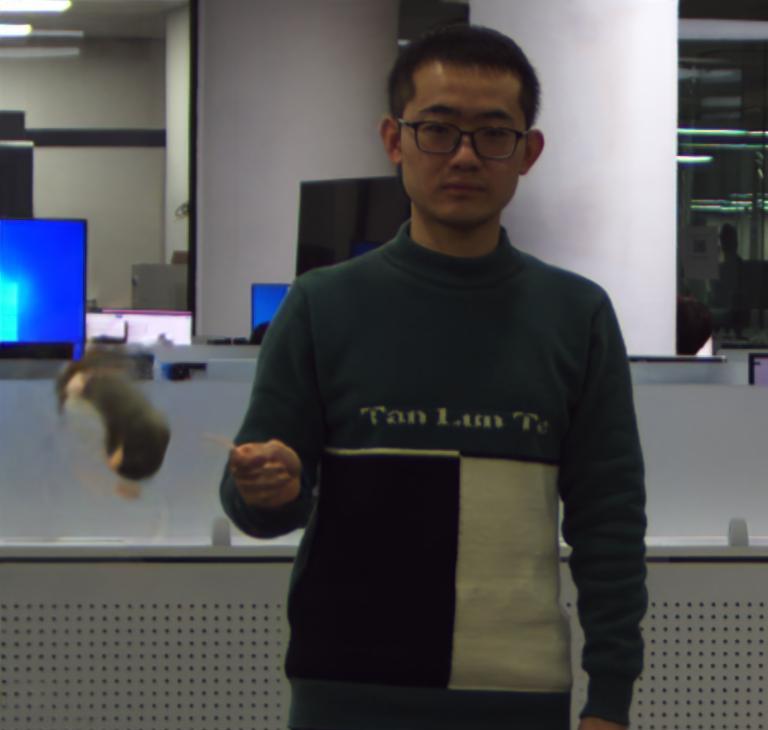} \\
    \multicolumn{2}{c}{{(c) LSR-HFR (Ours)}} & 
    \multicolumn{2}{c}{{(d) HSR-LFR (Ours)}} \\
	\end{tabular}
    % \vspace{-5pt}
	\caption{{\textbf{Failure case.} 30 and 75 are the frame numbers. CDVD~\cite{Pan_2020_CVPR}+DAIN~\cite{bao2019depth} denotes the joint video deblurring and frame interpolation method.}}
	\label{fig:failed}
	\vspace{-15pt}
\end{figure}
  	
  	\subsection{Failure Case}
    A challenging scene for our method is large motion and heavy motion blur of thin structures, and the disparity difference between the moving object and its background is large.
    As shown in frame 30 in Figure~\ref{fig:failed}, the depth difference between the mouse doll and its background (the man) is small. Then, the mouse doll can be preserved even with heavily blurred HSR-LFR inputs. However, the mouse doll is damaged in frame 75 because the background of the mouse doll becomes the far-away wall. The estimation of large disparity is challenging with heavily blurred input.
    In this case, the fast-moving object will be deformed in HSR-LFR view but preserved in LSR-HFR view.
    The straightforward solution to avoid these failure cases is to increase the frame rate or reduce the exposure time of the HSR-LFR view.
    In algorithm design, we could refine the disparity at each time $t$ with the help of the blur-free LSR-HFR frame. Then, the accurate appearance of the fast-moving object can be transferred from LSR-HFR view to HSR-LFR view.
    Since the input HSR-LFR frames are seriously blurred, it is challenging to warp the blurred HSR-LFR frames to intermediate time to reconstruct a sharp and well-aligned intermediate frame. 
    As shown in Figure~\ref{fig:failed}(b), the mouse completely disappears in the results of the joint video deblurring and frame interpolation method (CDVD + DAIN).
    Note that the LSR-HFR frame is sharp and in good shape.
    Thus, the most promising solution is to warp the blur-free LSR-HFR frame to HSR-LFR view to assist the frame reconstruction. According to the analyses in Section~\ref{sec:comwarp}, the warped frame based on disparity is critical for reconstructing fast-moving objects. 
    Then, the accuracy of the disparity network is critical in this challenging scene.
    However,  it will be difficult to estimate an accurate disparity between a blurred HSR-LFR frame and a sharp LSR-HFR frame, especially when the foreground blurred object is far away from the background.
    Notice that the disparity maps in our method are aligned with the LSR frames.
    Inspired by monocular image depth estimation and image-based depth completion methods, we can refine the disparity $d^t$ with the blur-free LSR-HFR frame $\hat{L}^t$ using a disparity refinement network.
    Then, we can restore the good shape of the fast-moving blurred objects in HSR-LFR view based on the accurate disparity.
    Due to the space limitation, the disparity refinement will be our future work.

\section{Conclusion}
    We propose a hybrid camera system for high-speed, high-resolution stereoscopic (H$^2$-Stereo) video synthesis, where one camera captures high spatial-resolution low-frame-rate video and the other captures low-spatial-resolution high-frame-rate video.
    We propose a learned information fusion network (LIFnet), which introduces a disparity network to transfer spatiotemporal information across views even in large disparity scenes, and a featured-based multi-scale fusion network to minimize occlusion-induced warping ghosts and holes in HSR-LFR view.
    We create a Stereo Video dataset and an H$^2$-Stereo dataset for training and evaluation.
    Objective and subjective results and extensive analyses, including spatiotemporal resolution, camera baseline, camera desynchronization, long/short exposures and applications, demonstrate the robustness and effectiveness of our dual camera system.

\ifCLASSOPTIONcaptionsoff
  \newpage
\fi

\normalem
\bibliographystyle{IEEEtran}
\bibliography{egbib}

\vspace{-25pt}
\begin{IEEEbiography}[{\includegraphics[width=1in,height=1.25in,clip,keepaspectratio]{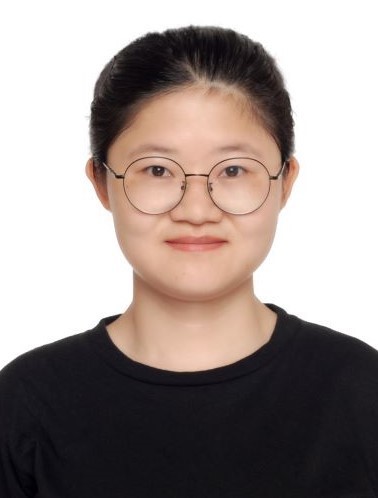}}]{Ming Cheng}
is a Ph.D. candidate of Electrical Engineering with the Institute of Image Communication and Network Engineering, Shanghai Jiao Tong University, Shanghai, China.
She received the B.S. degree in the University of Electronic Science and Technology of China, China, in 2016.
Her research interests include computer vision, video processing and multi-cameras.
\vspace{-20pt}
\end{IEEEbiography}

\begin{IEEEbiography}[{\includegraphics[width=1in,height=1.25in,clip,keepaspectratio]{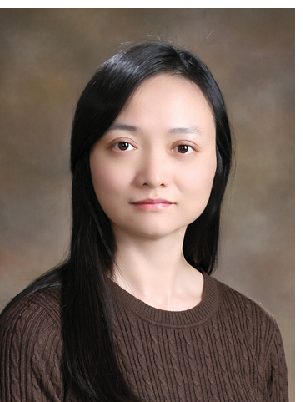}}]{Yiling Xu} is a full researcher of School of Electronic Information and Electronic Engineering, Shanghai Jiao Tong University, Shanghai, 200145, China.
She received the B.S., M.S. and Ph.D. from the University of Electronic Science and Technology of China, China, in 1999, 2001 and 2004 respectively. From 2004 to 2013, she was with Multimedia Communication Research Institute of Samsung Electronics Inc, Korea.
Her main research interests include architecture design for next generation multimedia systems, dynamic data encapsulation, adaptive cross layer design, dynamic adaption for heterogenous networks and N-screen content presentation.
\vspace{-20pt}
\end{IEEEbiography}

\begin{IEEEbiography}[{\includegraphics[width=1in,height=1.25in,clip]{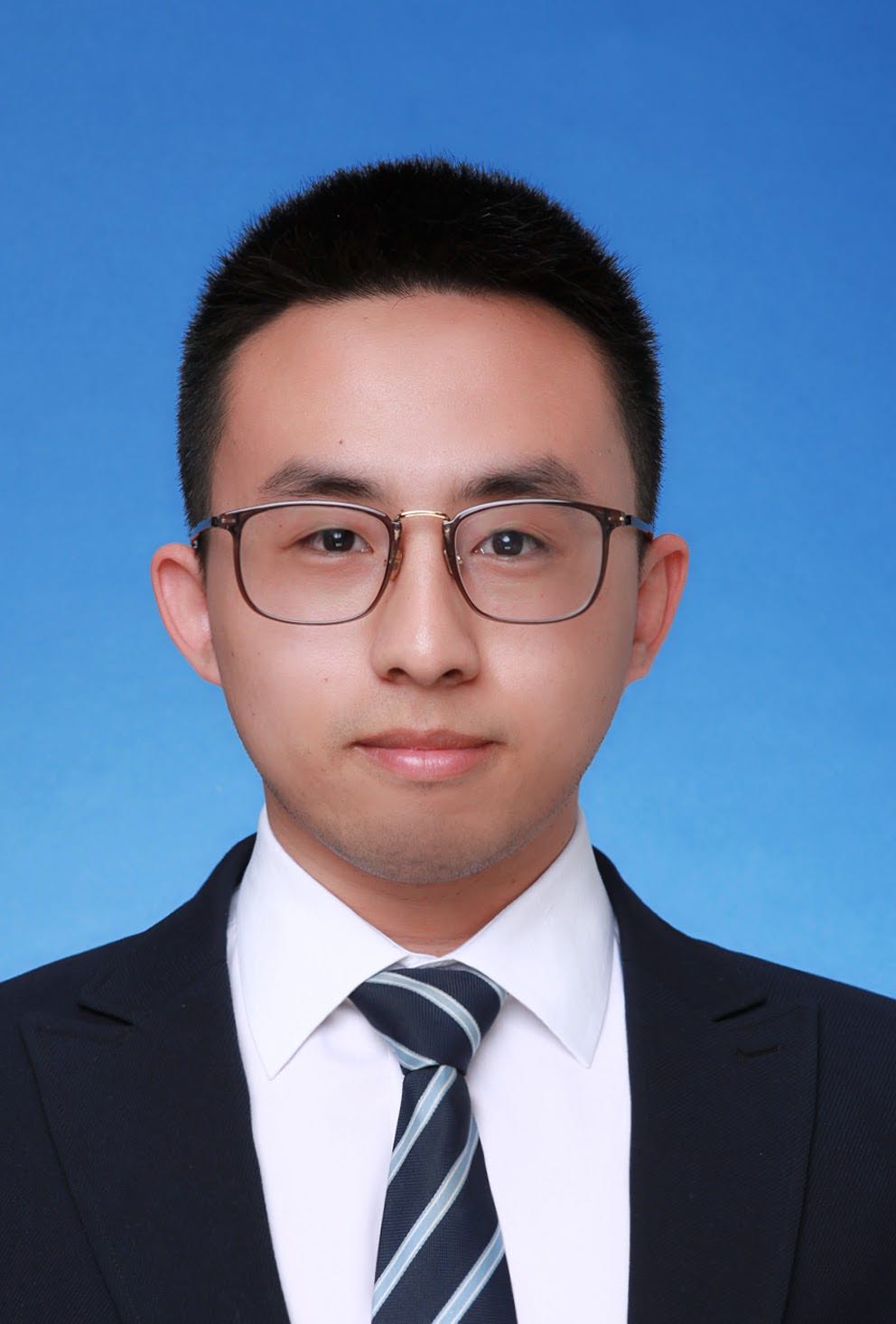}}]{Wang Shen}
recieved the B.S. degree in electronic information engineering from the University of Electronic Science and Technology of China, Chengdu, China, in 2017. He is currently pursuing the Ph.D. degree in electrical engineering with the Institute of Image Communication and Network Engineering, Shanghai Jiao Tong University, Shanghai, China. His current research interests include computer vision, machine learning, and video processing.
\vspace{-5 mm}
\end{IEEEbiography}

\begin{IEEEbiography}[{\includegraphics[width=1in,height=1.25in,clip,keepaspectratio]{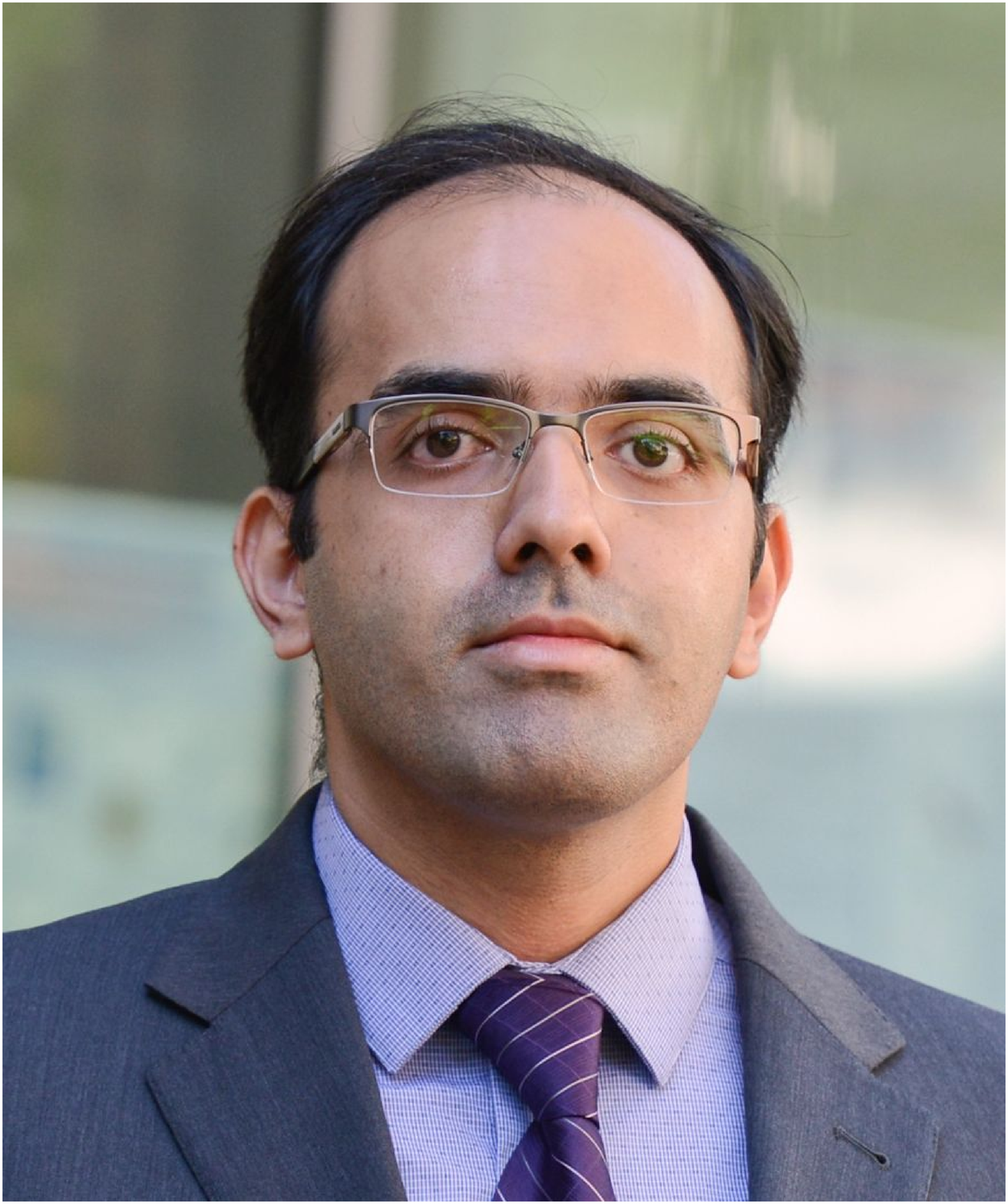}}] {M. Salman Asif} is currently an Assistant Professor in the Department of Electrical and Computer Engineering at the University of California, Riverside.
Prior to joining UC Riverside, he was a postdoctoral research associate in the DSP group at Rice University. Before that he briefly worked as a Research Engineer at Samsung Research America, Dallas. He received the Ph.D. at the Georgia Institute of Technology under the supervision of Justin Romberg.
His research interests broadly lie in the areas of information processing and computational sensing with applications in signal processing, machine learning, and computational imaging.
\vspace{-20pt}
\end{IEEEbiography}

\begin{IEEEbiography}[{\includegraphics[width=1in,height=1.25in,clip,keepaspectratio]{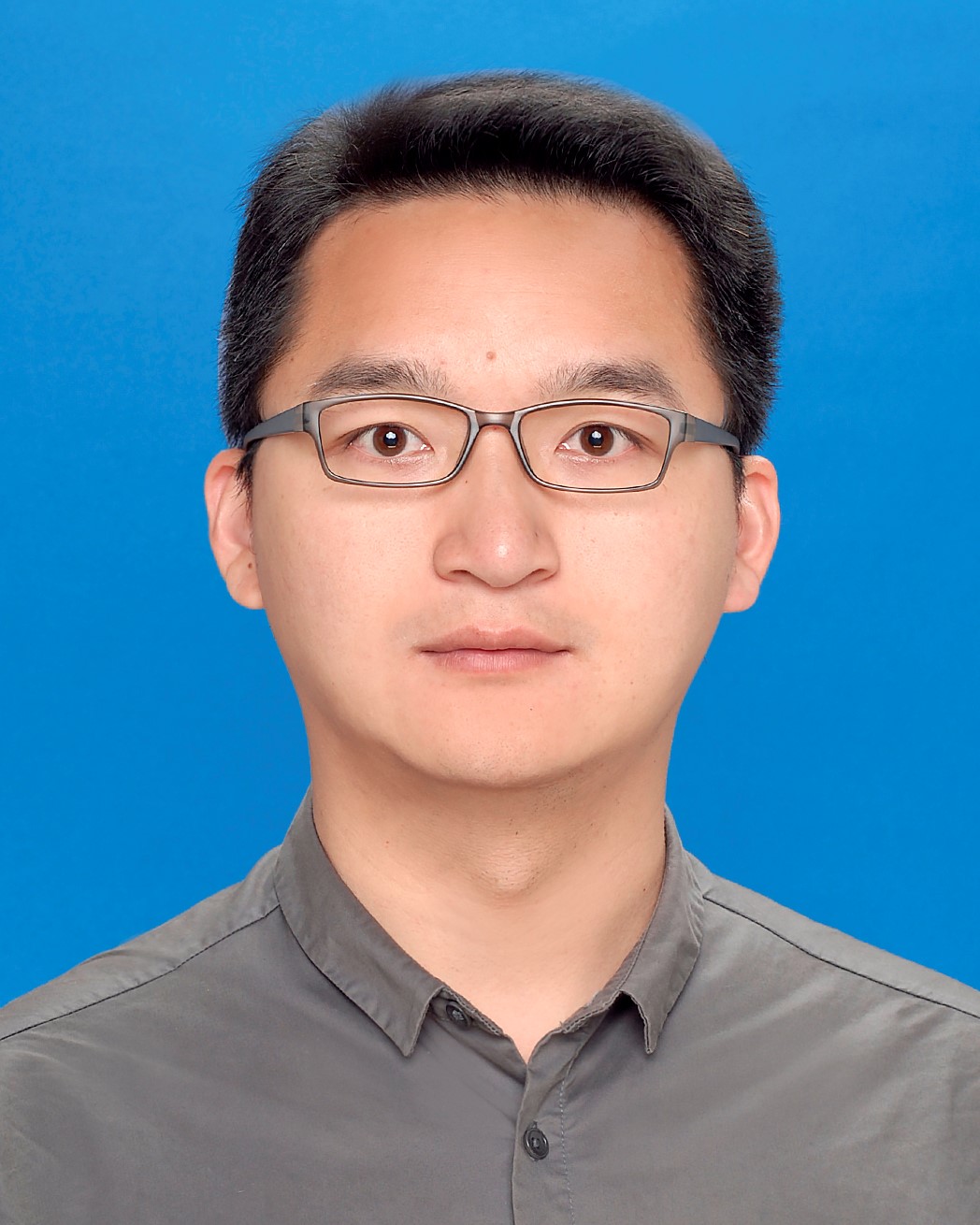}}]{Chao Ma} (Member IEEE) received the Ph.D. degree from Shanghai Jiao Tong University in 2016. He was a Senior Research Associate with the Australian Centre of Robotic Vision, The University of Adelaide, from 2016 to 2018. He has been an Assistant Professor with Shanghai Jiao Tong University since 2019. He was sponsored by the China Scholarship Council as a visiting Ph.D. Student at the University of California at Merced from 2013 to 2015. His research interests include computer vision and machine learning.
\vspace{-20pt}
\end{IEEEbiography}

\begin{IEEEbiography}[{\includegraphics[width=1in,height=1.25in,clip,keepaspectratio]{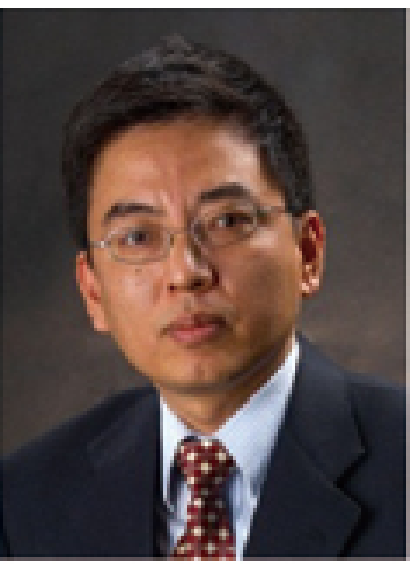}}]{Jun Sun} is currently a professor and Ph.D. advisor of Shanghai Jiao Tong University.
He received his B.S. in 1989 from University of Electronic Sciences and technology of China, Chengdu, China, and a Ph.D. degree in 1995 from Shanghai Jiao Tong University, all in electrical engineering.
In 1996, he was elected as the member of HDTV Technical Executive Experts Group (TEEG) of China. Since then, he has been acting as one of the main technical experts for the Chinese government in the field of digital television and multimedia communications. In the past five years, he has been responsible for several national projects in DTV and IPTV fields. He has published over 50 technical papers in the area of digital television and multimedia communications and received 2nd Prize of National Sci. \& Tech. Development Award in 2003, 2008.
His research interests include digital television, image communication, and video encoding.
\vspace{-20pt}
\end{IEEEbiography}
\begin{IEEEbiography}[{\includegraphics[width=1in,height=1.25in,clip,keepaspectratio]{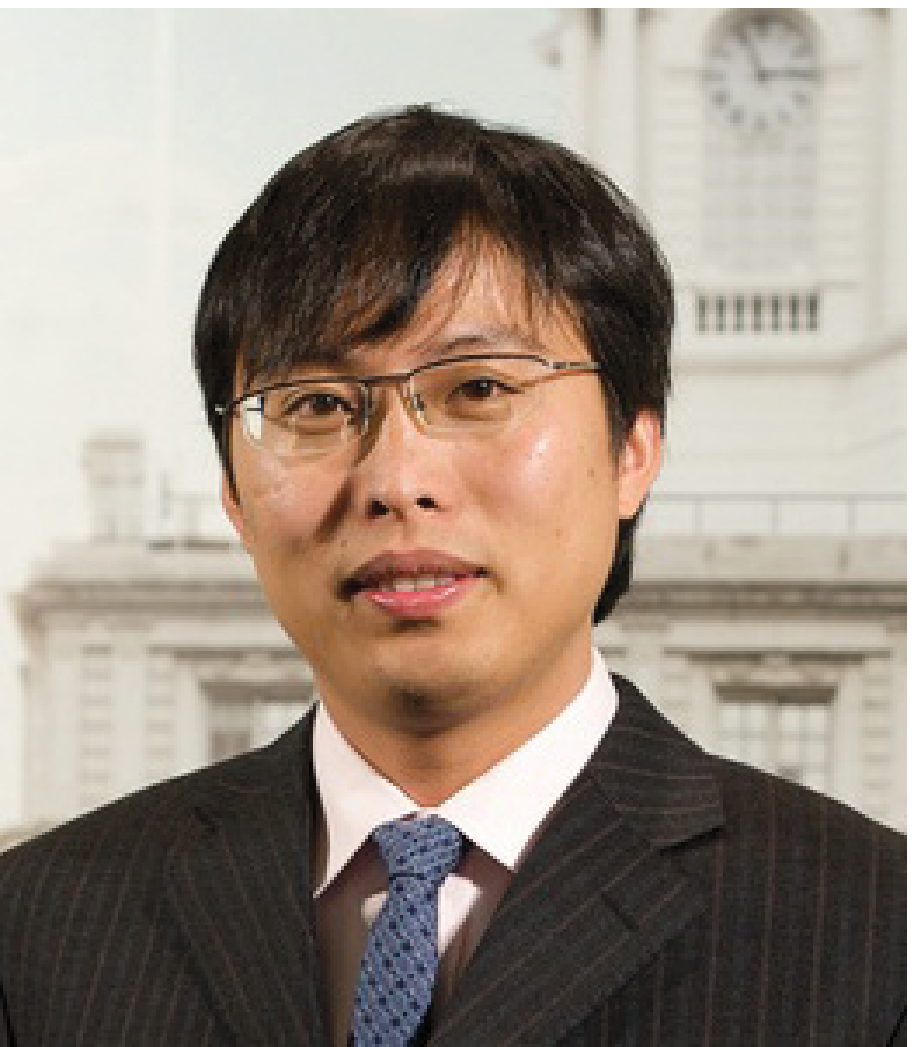}}]{Zhan Ma} (SM'19) is now on the faculty of Electronic Science and Engineering School, Nanjing University, Jiangsu, 210093, China.
He received the B.S. and M.S. from Huazhong University of Science and Technology (HUST), Wuhan, China, in 2004 and 2006 respectively, and the Ph.D. degree from the New York University, New York, in 2011.
From 2011 to 2014, he has been with Samsung Research America, Dallas TX, and  Futurewei Technologies, Inc., Santa Clara, CA, respectively. His current research focuses on the next-generation video coding, energy-efficient communication, gigapixel streaming and deep learning. He is a co-recipient of 2018 ACM SIGCOMM Student Research Competition Finalist, 2018 PCM Best Paper Finalist, and 2019 IEEE Broadcast Technology Society Best Paper Award.
\vspace{-20pt}
\end{IEEEbiography}

\end{document}